\documentclass[10pt,twocolumn,letterpaper]{article}
\pdfoutput=1 
\usepackage{cvpr}
\usepackage{times}
\usepackage{epsfig}
\usepackage{graphicx}
\usepackage{amsmath}
\usepackage{amssymb}
\usepackage{booktabs} 
\usepackage{makecell} 
\usepackage{caption} 
\usepackage{graphbox} 
\usepackage{tabularx} 
\usepackage{flushend}
\usepackage{placeins}
\usepackage{subcaption} 
\usepackage{float} 
\usepackage{multirow}
\graphicspath{{./supplement/}} 

\usepackage[font=small]{caption}

\usepackage[final]{changes}
\definecolor{burntorange}{rgb}{0.8, 0.33, 0.0}
\colorlet{Changes@Color}{burntorange}

\newcolumntype{Y}{>{\centering\arraybackslash}X}
\newcolumntype{C}[1]{>{\centering\let\newline\\\arraybackslash\hspace{0pt}}m{#1}}

\usepackage{enumitem}
\makeatletter
\renewcommand{\paragraph}{%
  \@startsection{paragraph}{4}%
  {\z@}{0.8ex \@plus 0.4ex \@minus .2ex}{-1em}
  {\normalfont\normalsize\bfseries}%
}
\makeatother
\newcommand{\bx}{\mathbf{x}}

\newcommand{\bz}{\mathbf{z}}

\newcommand{\bw}{\mathbf{w}}

\newcommand{\bA}{\mathbf{A}}

\newcommand{\E}{\mathbb{E}}

\newcommand{\bc}{\mathbf{c}}
\newcommand{\bct}{\bar{\bc}}

\newcommand{\bxt}{\bar{\bx}}

\newcommand{\animatetwo}[2]{#1}

\usepackage{xcolor}

\cvprfinalcopy 

\def\cvprPaperID{1051} 

\begin{document}

\title{The GAN that Warped: Semantic Attribute Editing with Unpaired Data}

\author{Garoe Dorta\,$^{1,2}$ \quad Sara Vicente\,$^{2}$ \quad Neill D.\,F. Campbell\,$^{1}$ \quad Ivor J.\,A. Simpson\,$^{2,3}$\\
$^{1}$University of Bath \qquad $^{2}$Anthropics Technology Ltd. \qquad $^{3}$University of Sussex \\
{\tt\small \{g.dorta.perez,n.campbell\}@bath.ac.uk \quad sara@anthropics.com \quad i.simpson@sussex.ac.uk}
}

\twocolumn[{
\renewcommand\twocolumn[1][]{#1}
\maketitle
\def\plotw{0.32\linewidth}

\def\imga{3} 

\begin{center}
	\centering
	\setlength{\tabcolsep}{1pt} 
	\footnotesize{
	\begin{tabular}{ccc}
	\includegraphics[trim={325px 0cm 200px 150px},clip,width=\plotw]{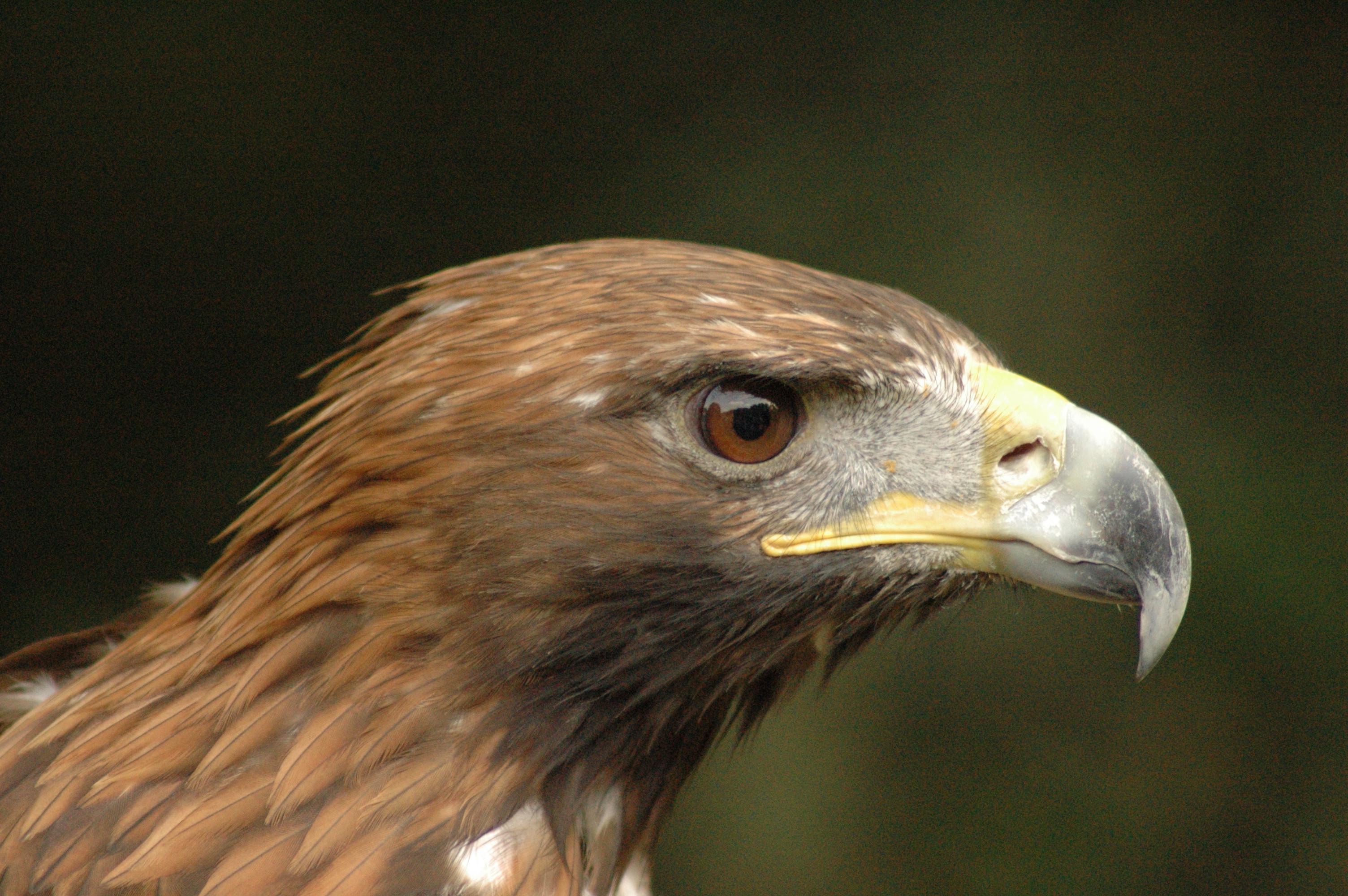} &
	\includegraphics[trim={325px 0cm 200px 150px},clip,width=\plotw]{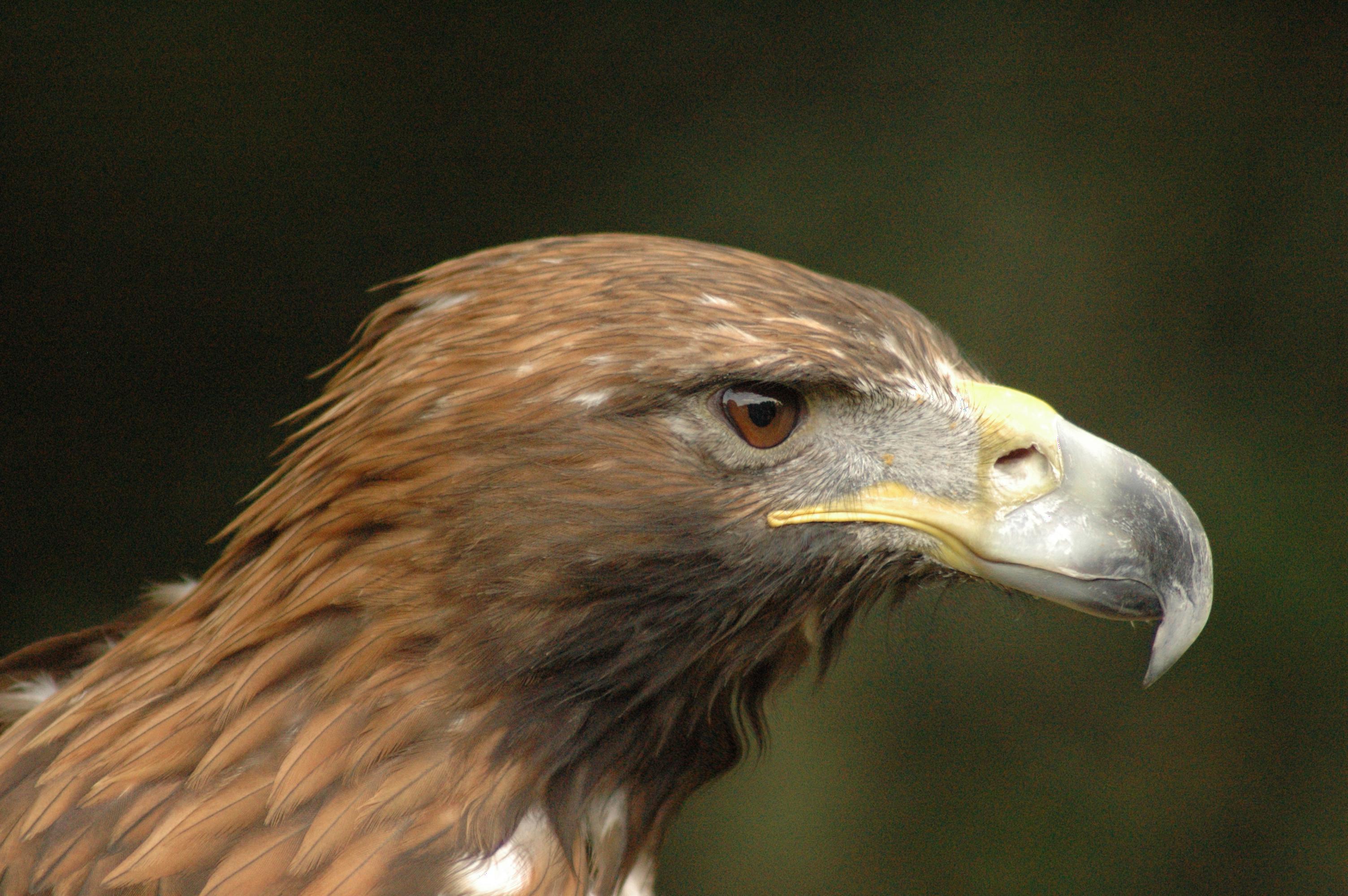} &
	\includegraphics[trim={325px 0cm 200px 150px},clip,width=\plotw]{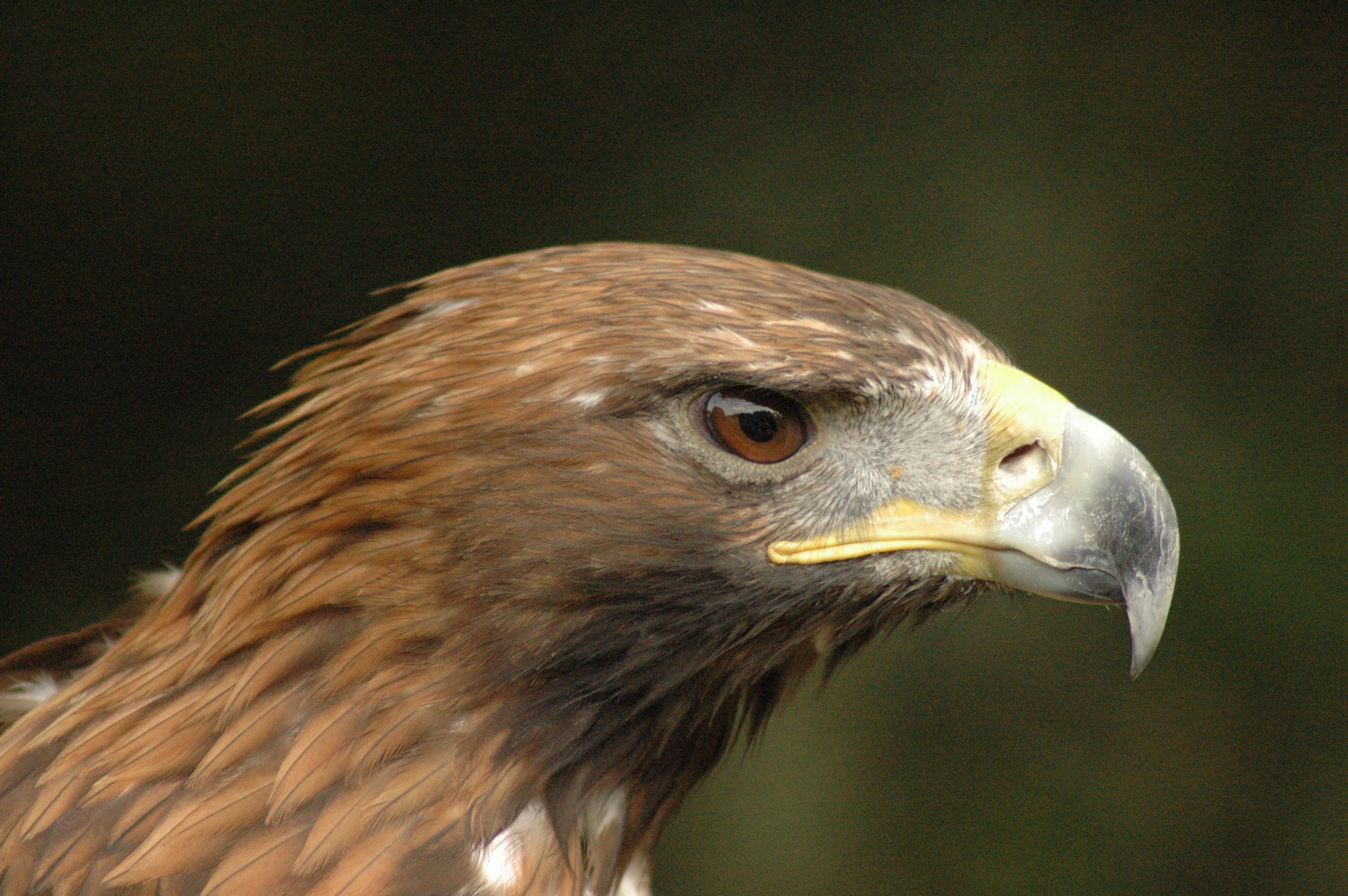}
	\\
	(a) Input image (previously unseen) & (b)~User requested edit: ``beak larger than head''  & (c)~User requested edit: ``beak smaller than head''
	\end{tabular}
	}
	\captionof{figure}{Semantic image editing at high resolution (2480$\times$1850).
	The user requests a change in a semantic attribute and the input image (a) is automatically transformed by our method into, \eg, an image with ``beak larger than head''~(b) or ``beak smaller than head''~(c).
	The content of the original input, including fine details, is preserved.
	Our focus is on face editing, as previous work, yet the method is general enough to be applied to other datasets.
	Please see the supplemental material for videos of these and other edits. (Zoom in for details)\protect\footnotemark{\label{ft:animations}}}
	\label{fig:teaser}
\end{center}%

\thispagestyle{empty}
}]

\def\animatedfootnote{}

\footnotetext{Image courtesy of Flickr user Christoph Landers.}


\begin{abstract}\vspace{-10pt}
Deep neural networks have recently been used to edit images with great success, in particular for faces.
However, they are often limited to only being able to work at a restricted range of resolutions. 
Many methods are so flexible that face edits can often result in an unwanted loss of identity.
This work proposes to learn how to perform semantic image edits through the application of smooth warp fields.
Previous approaches that attempted to use warping for semantic edits required paired data, \ie example images of the same subject with different semantic attributes.
In contrast, we employ recent advances in Generative Adversarial Networks that allow our model to be trained with unpaired data. 
We demonstrate face editing at very high resolutions (4k images) with a single forward pass of a deep network at a lower resolution.
We also show that our edits are substantially better at preserving the subject's identity.
The robustness of our approach is demonstrated by showing plausible image editing results on the Cub200~\cite{Wah2011CUB_200} birds dataset.
To our knowledge this has not been previously accomplished, due the challenging nature of the dataset.
\end{abstract}


\section{Introduction}
\label{sec:introduction}
\vspace{-2pt}

Face editing has a long history in computer vision~\cite{Liu_RatioImage2001, Mohammed_visio2009, Turk1991EigenFaces} and has been made increasingly relevant with the rise in the number of pictures people take of themselves or others.
The type of edits that are performed usually manipulate a semantic attribute, such as removing a moustache or changing the subject's expression from a frown to a smile.

In the last few years, deep learning approaches have become the standard in most editing tasks, including inpainting~\cite{Pathak2016ContextEncoders}\replaced{ and}{,} super-resolution~\cite{Ledig2017SRGAN}\deleted{, and face editing~\cite{Portenier_FaceShop2018}}.
Particularly, image-to-image translation methods~\cite{Isola_pix2pix2017} have been proposed, which learn how to transform an image from a source domain to a target domain.
\deleted{The }Cycle-GAN\deleted{ approach}~\cite{Zhu_CycleGAN2017} allows learning such translations from unpaired data, \ie~for each source image in the dataset a corresponding target image is not required.

We are interested in photo-realistic image editing, which is a subset of image-to-image translation.
We also focus on methods that provide a simple interface for users to edit images, \ie a single control per semantic attribute~\cite{Choi_StarGAN2018,Pumarola_Ganimation2018}, as this makes editing easier for novice users.

A disadvantage of current methods for editing~\cite{Isola_pix2pix2017, Choi_StarGAN2018, Yeh_FlowVAE2016} is that they focus on binary attribute changes.
In order to allow partial edits an extensive collection of soft attribute data is usually required, which is labor intensive.
Also, at inference each intermediate value requires a forward pass of the network, creating increased computational expense~\cite{Pumarola_Ganimation2018}.

Most deep learning methods for image editing predict the pixel values of the resulting image directly~\cite{Choi_StarGAN2018, Dekel2018GrandietContours, Portenier_FaceShop2018, Pumarola_Ganimation2018}.
As a consequence these methods are only effective on images that have a similar resolution to the training data.

Recently, some interesting approaches that do allow edits at higher resolutions have been proposed.
They proceed by estimating the edits at a fixed resolution and then applying them to images at a higher resolution.
The types of possible edits are restricted to either warping~\cite{Yeh_FlowVAE2016} or local linear color transforms~\cite{Gharbi_Bilateral2017}.
However, these approaches are limited by requiring paired data, \ie~for each source image in the dataset, they need the corresponding edited image. 

Inspired by these high resolution methods, we introduce an approach to learn smooth warp fields for semantic image editing without the requirement of paired training data samples.
This is achieved by exploiting the recent advances in learning edits from unpaired data with cycle-consistency checks, which derive from the Cycle-GAN~\cite{Zhu_CycleGAN2017} method.
Our proposed model uses a similar framework to StarGAN~\cite{Choi_StarGAN2018} (an extension of Cycle-GAN) to predict warp fields that transform the image to achieve the requested edits.
As the predicted warp fields are smooth, they can be trivially upsampled and applied at high resolutions.

A potential criticism is that there are clear limitations to the types of edits possible through warping.
We argue that, for the changes that \emph{can} be described in this way, there are several distinct benefits.
The advantages of using warping with respect to pixel based models can be summarized as:
\begin{enumerate}[topsep=1pt,itemsep=0pt,partopsep=0pt,label=\roman*.]
	\item Smooth warp fields can be upsampled and applied to higher resolution images with a minimal loss of fidelity.
	This is opposed to upsampling images, which commonly results in unrealistic high frequency details.
	We show plausible edits using warp fields upscaled by up to 30$\times$ the resolution they were estimated at.
	\item Geometric transformations are a subset of image transformation models.
	These models make it easier to add priors to regularize against unrealistic edits.
	We demonstrate that editing by warping leads to a model that is better at preserving a subject's identity.
	\item Warp fields are more interpretable than pixelwise differences. 
	We illustrate this with maps showing the degree of local stretching or squashing.
	\item Warp fields are much more suited to allow partial edits than pixel based approaches.
	We demonstrate the simplest implementation of this by scaling the warp field to show interpolation and extrapolation, and qualitatively show edits that are plausible.
\end{enumerate}

An additional contribution of this work is to improve the specificity of editing attributes in StarGAN based models.
We have observed that when these models are trained with several binary labels, they can transform more than one attribute of the image, even if only a single attribute should be edited.
This is caused by the model having no indication of the attributes that should be edited, only of the final expected labels. 
For example, when enlarging the nose of a subject that has a slight smile, the model will not only make the nose bigger, but also make the smile more pronounced.
In order to overcome this limitation, we propose to transform the labels to inform the model of which attributes should be edited, and which should remain fixed.
This produces only the expected changes, and it does not require any extra label annotation.
Moreover, it removes the need to rely on a label classifier during inference.

We demonstrate the advantages of our contributions by providing quantitative and qualitative results by manipulating facial expressions and semantic attributes.


\vspace{-1pt}
\section{Previous work}
\vspace{-1pt}

This work builds upon recent work in image-to-image translation.
These models can be used to modify the semantic attributes of an image.
Our novelty is in describing these edits as smooth deformation fields, rather than producing an entirely new image.
Smooth warp fields can be upsampled and applied to higher resolution images with a minimal loss of fidelity.
Some previous works that allow high resolution editing rely upon paired data examples or require costly optimization, rather than a single forward pass of a network; neither of which is required for the proposed approach. 
An overview of the characteristics of our work compared to previous methods is shown in Table~\ref{tb:method_characteristics}.

\begin{table}[tbp!]
	\scalebox{0.95}{%
		\begin{minipage}{\columnwidth}
			\scriptsize{ 
				\begin{tabular}{lccc}
					\textbf{\footnotesize{Method}} & \textbf{\footnotesize{Unpaired data}} & \textbf{\footnotesize{High resolution}} & \textbf{\footnotesize{Forward pass}} \\ \midrule
					StarGAN~\cite{Choi_StarGAN2018}         & \checkmark &             & \checkmark \\ \midrule
					FaceShop~\cite{Portenier_FaceShop2018}  & \checkmark &             & \checkmark \\ \midrule
					WG-GAN~\cite{Geng2018WarpGANAnimation}  &            &            & \checkmark \\ \midrule
					FlowVAE~\cite{Yeh_FlowVAE2016}          &            & \checkmark  & \checkmark \\ \midrule
					CWF~\cite{Ganin2016Gaze}                &            & $\sim$      & \checkmark \\ \midrule
					DBL~\cite{Gharbi_Bilateral2017}         &            & \checkmark  & \checkmark \\ \midrule
					iGAN~\cite{Zhu2016GanShoes}             & \checkmark & $\sim$      &            \\ \midrule
					DFI~\cite{Upchurch2017DFI}              & \checkmark & $\sim$      &            \\ \midrule
					\added{RelGAN~\cite{Wu2019RelAttrGAN}}   & \checkmark &             & \checkmark \\ \midrule
					\added{SPM+R~\cite{Wu2019WarpInpaintGAN}}& \checkmark & $\sim$      & \checkmark \\ \midrule
					\textbf{Ours}                           & \checkmark & \checkmark  & \checkmark \\ \bottomrule
				\end{tabular}
			}
		\end{minipage}%
	}
	\caption{Compared to previous work on image-to-image translation, our model is the only one that is able to edit high-resolution images in a single forward pass of the network, 
		without paired training data.
		Partial fulfillment of the criterion is denoted by $\sim$.}
	\label{tb:method_characteristics}
\end{table}

\subsection{Image-to-Image translation}

The Pix2Pix~\cite{Isola_pix2pix2017} model learns to transform an image from a source domain to a target domain using an adversarial loss~\cite{Goodfellow2014GAN}.
This approach requires paired training data; \ie each image in the source domain must have a corresponding image in the target domain.
Given this restriction, the method is often applied to problems where collecting paired data is easier, such as colorization. 

Several extensions have been proposed to perform image-to-image translation without requiring paired data. 
In Cycle-GANs~\cite{Zhu_CycleGAN2017}, two generators are trained, from the source to the target domain and vice versa, with a cycle-consistency loss on the generation process.
However, this does not scale well with an increase in the number of domains, since 2 generators and 2 discriminators are needed for each domain pair.
StarGAN~\cite{Choi_StarGAN2018} addresses this issue by conditioning the generator on a domain vector, and adding a domain classification output layer to the discriminator.

These models can find undesired correlations within a domain, which lead to changes in unexpected parts of the image.
At least two techniques have been explored in order to encourage localized edits.
Editing with residual images~\cite{Shen2017ResidualGAN}, and restricting the changes to a region given by a mask~\cite{Mejjati2018AttCycleGAN,Pumarola_Ganimation2018}.
The first is an overcomplicated representation for edits describing shape changes.
It has to model the content in the region, subtract it, and add it in a second region.
The second complicates the model significantly by adding an unsupervised mask prediction network.

\added{Preceding this publication, two relevant extensions to StarGAN have been proposed: RelGAN~\cite{Wu2019RelAttrGAN} and SPM+R~\cite{Wu2019WarpInpaintGAN}.
RelGAN proposes a binary label transformation approach similar to ours.
However, their method is trained using a conditional adversarial loss that takes triplets composed of two images and a vector of changed attributes.
In contrast, our approach uses a simpler classification loss, where only modified attributes count.
RelGAN also enables partial editing, however it requires a forward pass of the network for each edit strength.
In contrast, our approach trivially enables partial editing as a consequence of the edit being performed through warping.
Similarly to our work, SPM+R suggests using a warping function to edit images; yet, this is followed by inpainting, which is not resolution agnostic.
They do not demonstrate their approach for editing high resolution images ($>$$512$$\times$$512$), or on a more complex dataset such as Cub200.
A further distinction is that instead of using a simple smoothness loss, as we propose, they use a warp field discriminator.
Their resulting warp fields appear substantially less smooth and less sparse than the ones obtained by our approach.}

\subsection{Editing of high resolution images}
Methods for editing images at high resolution can be divided into two categories:
(i) those that use intermediate representations designed to upsample well, and (ii) those that directly predict pixel values at high resolutions.

\paragraph{Methods designed for upsampling}

These approaches are based on predicting constrained intermediate representations that are relatively agnostic to image resolution; \eg warp fields or local color affine transformations. 

Warp fields, if sufficiently smooth, can be predicted at a lower resolution, upsampled and applied at high resolution with minimal loss of accuracy.
Previous work has applied them to: redirecting eye gaze~\cite{Ganin2016Gaze}, editing emotional expressions~\cite{Yeh_FlowVAE2016} and synthesizing objects in novel views~\cite{Zhou2016ViewFlow}.
However, these methods require paired training data.

Spatial Transformer GANs~\cite{Lin2018STGan} predict a global affine deformation for image compositing.
Although the deformation can be applied at arbitrary resolutions, face editing by compositing is limiting, as it requires an infeasibly large dataset of suitable face parts to use as foreground images.


Local affine color transformations~\cite{Chen2016BGU,Gharbi_Bilateral2017} have been predicted from low resolution images and applied at the original resolution. 
However, these methods require paired data and have limited capacity for making semantic changes.

%

Blendshapes have been used as an intermediate representation to edit expressions in the context of video reenactment~\cite{Thies2016Face2Face,Ma2018BlenshapeCycleGAN}.
Similar to our approach, the blendshape weights are resolution independent.
However, several input video frames are required for the blendshape face model.

Rather than predict intermediate representations, iGAN~\cite{Zhu2016GanShoes} trains a low-resolution GAN and then fit a dense warp field and local affine color transformation to a pair of input-output images. 
The GAN generator is unaware of these restricted transformations, so it may learn edits that are not representable by such transformations.

\paragraph{Direct prediction at high resolution}

Several techniques have been proposed to scale deep image synthesis methods to larger resolutions.
These include: synthesizing images in a pyramid of increasing resolutions~\cite{Denton2015LapGAN}, employing fully convolutional networks trained on patches~\cite{Ledig2017SRGAN}, and directly in full resolution~\cite{Brock2018LargeGAN, Karras2018ProgressiveGAN}. 
These methods have been successfully applied for image enhancement~\cite{Ignatov2017Wespe} and face editing~\cite{Portenier_FaceShop2018,Geng2018WarpGANAnimation}.
A limitation for direct or pyramid based approaches is that they do not scale well with resolution, while training on patches assumes that global image information is not needed for the edit. 

A method that modifies an image by following the gradient directions of a pretrained classification network, until it is classified as having the target attributes was proposed in~\cite{Upchurch2017DFI}.
However, this approach fails when the input resolution differs significantly from the training data.

\replaced{In}{Particularly similar to our method is} WG-GAN~\cite{Geng2018WarpGANAnimation}\deleted{, where} the input image is warped \added{based on a target image} and \replaced{two}{a} GAN generator\added{s} \replaced{are}{is} used thereafter to synthesize new content.
Contrary to our method, WG-GAN requires paired data during training, cannot be applied at arbitrary resolutions, \added{does not provide semantic controls} and does not support partial edits.

\section{Background}

\begin{figure*}[btp]
	\centering
	\includegraphics[width=0.9\linewidth]{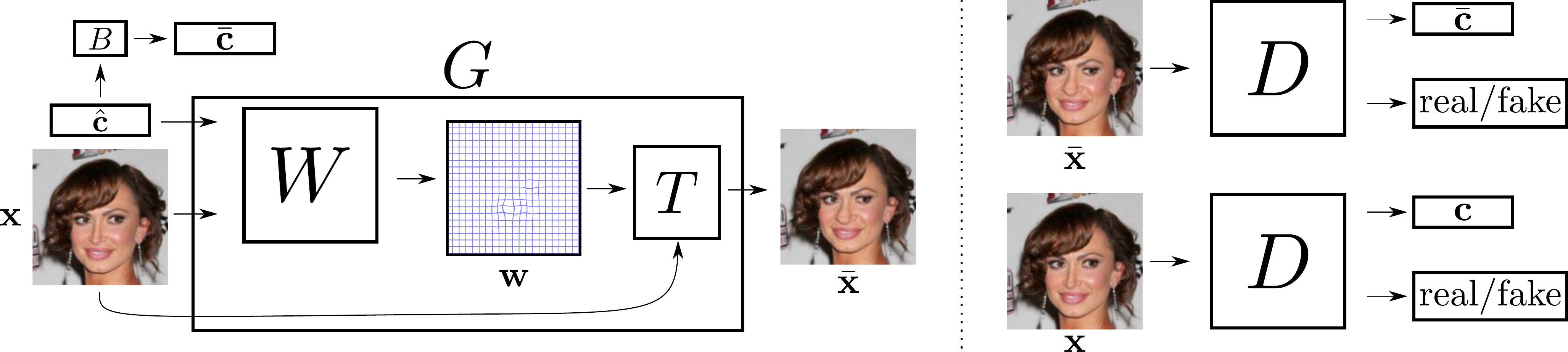}
	\caption{Overview of our model, which consists of a generator, $G$, and a discriminator, $D$.
	The generator contains a warping network, $W$, and a warping operator, $T$.
	The inputs to $W$ are an RGB image, $\bx$, and a transformed label vector, $\hat{\bc}$.
	The output is a dense warp field, $\bw$, which can be used by $T$ to deform the input image and produce the output image, $\bxt$.
	A label operator, $B$, converts the transformed labels, $\hat{\bc}$, to binary labels, $\bct$.
	The discriminator evaluates both the input image, $\bx$, and the generated image, $\bxt$, for realism and the presence of attributes that agree with the labels.
	In this example, the only change between $\bct$ and $\bc$ is the label for the attribute ``big nose''.}
	\label{fig:warp_gen_arch}
\end{figure*}

We start by reviewing GAN~\cite{Goodfellow2014GAN}, Cycle-GAN~\cite{Zhu_CycleGAN2017} and StarGAN~\cite{Choi_StarGAN2018}, as the latter is the basis for our model.

Generative Adversarial Network (GAN)~\cite{Goodfellow2014GAN} models consist of two parts, a generator and a discriminator.
The generator produces samples that resemble the data distribution samples, and the discriminator classifies data samples as real or fake.
The discriminator is trained with the real examples drawn from a training set and the fake examples as the output of the generator.
The generator is trained to fool the discriminator into classifying generated samples as real.
Formally, GANs are defined by a minimax game objective:
\begin{equation}
\resizebox{0.9\hsize}{!}{$ 
\min_G\max_D \, \E_{\bx} \left[\, \log(D(\bx)) \,\right] + \E_{\bz} \left[\, \log( 1 - D(G(\bz))) \,\right],
$}
\end{equation}
where $\bx$ is a sample from the dataset empirical distribution $p(\bx)$, $\bz$ is a random variable drawn from an arbitrary distribution $p(\bz)$, $G$ is the generator and $D$ is the discriminator.

Given two data domains, $A$ and $B$, Cycle-GAN~\cite{Zhu_CycleGAN2017} learns a pair of transformations $G : A \rightarrow B$ and $H: B \rightarrow A$.
Unlike previous approaches,~\cite{Isola_pix2pix2017}, this does not require paired samples from $A$ and $B$, but instead utilizes a cycle consistency loss ($\| \bx_a - H(G(\bx_a)) \|_1$, where $\bx_a$ is a sample image from domain $A$) to learn coherent transformations that preserve a reasonable amount of image content.
An equivalent cycle loss is used for domain $B$.
Cycle-GAN models are limited in that they require 2 generators and 2 discriminators for each domain pair.

Cycle-GAN was generalized by StarGAN~\cite{Choi_StarGAN2018} to require only a single generator and discriminator to translate between multiple domains. 
Here, each image $\bx$ has a set of domains, represented as a binary vector $\bc$. 
We use $(\bx, \bc)$ to denote a pair sampled from the annotated data distribution.
The generator, $G(\bx,\bct)$, transforms $\bx$ to match the target domains indicated by $\bct \sim p(\bc)$, where $p(\bc)$ is the empirical domains distribution.
The model is trained with:
\begin{enumerate}[topsep=1pt,itemsep=0pt,partopsep=0pt,parsep=0pt,label=\roman*.]
	\item a Wasserstein GAN~\cite{Arjovsky2017WGAN} loss:
	\begin{equation}
	L_{adv} = \E_{\bx} \left[ D(\bx) \right] - \E_{\bx, \bct} \left[ D(G(\bx, \bct)) \right],
	\label{eq:adv_loss}
	\end{equation}
	\item a Wasserstein gradient penalty~\cite{Gulrajani2017WGANGP} term:
	\begin{equation}
	L_{gp} = \E_{\dot{\bx}} \left[ \left( \| \nabla_{\dot{\bx}}  D(\dot{\bx}) \|_2 - 1 \right)^2 \right],
	\label{eq:adv_gpw_loss}
	\end{equation}
	\item a cycle consistency loss:
	\begin{equation}
	L_c =  \E_{(\bx, \bc), \bct} \left[ \| \bx - G(G(\bx, \bct),\bc)\|_1 \right], \label{eq:star_cycle_loss}
	\end{equation}
	\item and domain classification losses:
	\begin{eqnarray}
	L_{cls}^{d} &=& \E_{(\bx, \bc)} \left[ -\log(C(\bx, \bc)) \right] \label{eq:C_d}\\
	L_{cls}^{g} &=& \E_{\bx, \bct} \left[ -\log(C(G(\bx, \bct), \bct)) \right], \label{eq:C_g}
	\end{eqnarray}
\end{enumerate}
where $C(\bx, \bc)$ is a classifier that outputs the probability that $\bx$ has the associated domains $\bc$, and $\dot{\bx}$ is sampled uniformly along a line between a real and fake image.
The classifier is trained on the training set (eq. \ref{eq:C_d}) and eq.~\ref{eq:C_g} ensures that the translated image matches the target domains.

\section{Methodology}
Our goal is to learn image transformations that can be applied at arbitrary scales without paired training data.
An overview of our system is shown in Figure~\ref{fig:warp_gen_arch}.
We employ the StarGAN framework as the basis for our model and use the notation introduced above.
As we focus on semantic face editing, we use indistinctly semantic attributes or binary labels to refer to the domains, $\bc$ and $\bct$.

\paragraph{Warp parametrization}
We modify the generator such that the set of transformations is restricted to non-linear warps of the input image:
\begin{equation}
G(\bx,\bct) = T(\bx, W(\bx,\bct)),
\end{equation}
where $W(\bx,\bct) = \bw$ is a function that generates the warp parameters.
$T$ is a predefined warping function that applies a warp to an image.
$W$ is chosen to be a neural network.
We employ a dense warp parametrization, where $\bw$ contains a displacement vector for each pixel in the input image.
At train time, $T$ warps the input according to the generated displacement field, $\bw$, using bilinear interpolation. 
To improve image quality at inference time we use a bicubic interpolant. 

\subsection{Learning}
We use the same adversarial losses (eq.~\ref{eq:adv_loss} and eq.~\ref{eq:adv_gpw_loss}) and domain classification loss (eq.~\ref{eq:C_d}) as StarGAN.

\paragraph{Warp cycle loss}
The cycle consistency loss (eq.~\ref{eq:star_cycle_loss}) is modified to produce warp fields that are inverse consistent, i.e. the composition of the forward and backward transformations yields an identity transformation:
\begin{equation}
L_{c} = \E_{(\bx, \bc), \bct} \left[ \| T(T(\bA,\bw), \bar{\bw})) - \bA \|_2^2 \right],
\end{equation}
where $\bar{\bw} = W(G(\bx, \bct), \bc)$, and $\bA$ is a two channel image where each pixel takes the value of its coordinates.
This loss is more informative than eq.~\ref{eq:star_cycle_loss}, as a pixel loss provides no information for warps inside constant color regions.

\paragraph{Smoothness loss}
The warping network estimates an independent deformation per pixel.
As such, there are no guarantees that the learned warps will be smooth.
Therefore, an $L2$ penalty on the warp gradients is added to encourage smoothness.
In practice a finite-difference approximation is used as 
\begin{equation}
\resizebox{\hsize}{!}{$ 
L_{s} = \E_{\bx, \bct} \left[ \frac{1}{n} \sum_{(i,j)} \| \bw_{i+1,j} - \bw_{i, j} \|_2^2 + \| \bw_{i, j+1} - \bw_{i, j} \|_2^2 \right],
$}
\end{equation}
where $n$ is the number of pixels in the warp field, and $\bw_{i,j}$ is the displacement vector at pixel coordinates $(i,j)$.

\paragraph{Binary label transformation}
As mentioned in section~\ref{sec:introduction}, a StarGAN type model can make unexpected edits when modifying attributes.
At inference time, the attribute classifier is used to infer the original labels.
Depending on the desired edits, these labels are either changed or copied to the target vector.
This means that the model cannot distinguish between the edited attributes and the copied ones.
Thus, the model tends to accentuate the copied attributes.



\replaced{To}{In order to} address this issue, we propose to explicitly instruct the generator on which attributes should be edited.
The \deleted{target} labels \deleted{used as input} for the generator are transformed to contain three values, $[-1, 0, 1]$, where $-1$ indicates that the attribute should be reversed, $0$ that it should remain unaffected, and $1$ that it should be added.
\deleted{
------------\\
--
}This approach has two distinct benefits.
First, it leads to more localized edits. 
Second, it removes the need for a classifier \deleted{network }during inference, as the unedited entries in the transformed target labels can be set to zero.

The classifier loss for the generator (eq.~\ref{eq:C_g}) is modified to only penalize the attributes that should be edited:
\begin{equation}
L_{cls}^{g} = \E_{\bx, \hat{\bc}} \left[ - h \sum^{r-1}_{i=0} |\hat{c}_i| \, \log \left(C(G(\bx, \hat{\bc}), \bar{c}_i) \right) \right], \label{eq:C_g_tern}
\end{equation}
where $\hat{\mathbf{c}}$ are the transformed target labels, $r$ is the number of attributes, and $h = r / \| \hat{\bc} \|_1$ is a normalization factor, which ensures that there is no bias with respect to the number of edited attributes.
During training, the transformed target label for each attribute, $\hat{c}_i$, is sampled independently from a Categorical distribution with probabilities $\left[ 0.25, 0.5, 0.25 \right]$.
\added{As both types of labels are needed for the classification loss, a} label operator, $\bar{c}_i = B(\hat{c}_i)$, is used to reverse the transformation, which is defined as $B(-1) = 0$ and $B(1) = 1$.
$B(0)$ is undefined as its loss is zero by construction.

\paragraph{Complete objective}
The joint losses for the discriminator and the generator are defined as
\begin{align}
L_{D} &= -L_{adv} + \lambda_{gp} L_{gp}  + \lambda_{cls} L^d_{cls}, \label{eq:D_warp_obj} \\
L_{G} &= L_{adv} + \lambda_{cls} L^g_{cls} + \lambda_{c} L_{c} + \lambda_s L_s, \label{eq:G_warp_obj}
\end{align}
where $\lambda_{cls}$, $\lambda_{gp}$, $\lambda_{c}$ and $\lambda_{s}$ are hyper-parameters that control the relative strength of each loss.
The classification loss in eq.~\ref{eq:C_g_tern} is used for images with several not mutually exclusive binary attributes, and eq.~\ref{eq:C_g} is used otherwise.

\subsection{Inference}

Once the model parameters have been optimized, an input image of arbitrary size can be edited in a single forward pass of the network.

The input image is resized to match the resolution of the training data, and the transformed target labels, $\hat{\bc}$, are set according to the desired edit.
Then, the resized image and target labels are fed into the warping network, which produces a suitable warp field, $\bw$.
The warp field displacement vectors are rescaled and resampled to the original image resolution. 
Lastly, the original image is warped using the resampled warp field to produce the final edited image.



\def\plotw{0.18\linewidth}
\def\imagea{9} 
\def\imageattr{1}

\begin{figure}[t]
	\centering
	\setlength{\tabcolsep}{2pt} 
	\footnotesize{
	\begin{tabular}{ccc|cc}
	Input & WarpGAN+ & StarGAN~\cite{Choi_StarGAN2018} & SGFlow \tiny{$0.05$} & SGFlow \tiny{$0.15$} \\
	\includegraphics[align=c,width=\plotw]{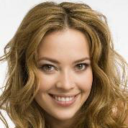} &
	\includegraphics[align=c,width=\plotw]{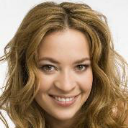} &
	\includegraphics[align=c,width=\plotw]{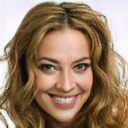} &
	\includegraphics[align=c,width=\plotw]{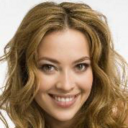} &
	\includegraphics[align=c,width=\plotw]{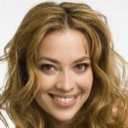}
	\end{tabular}
	}
	\caption{Employing a dense flow method~\cite{Zach2007OpenCVFlow} to transfer a ``big nose'' edit from StarGAN~\cite{Choi_StarGAN2018}.
		Results for the flow method with $\lambda = 0.05$ and $\lambda=0.15$ are shown.
		StarGAN has edited the input to such lengths that good correspondences between the input and output cannot be found by the flow method.}
	\label{fig:celeba_sg_plus_flow}
\end{figure}

\def\plotw{0.08\linewidth}
\def\imagea{7}
\def\imageb{8}
\def\imagec{9}
\def\imaged{12}

\def\inputa{\includegraphics[align=c,width=\plotw]{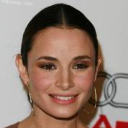}}
\def\inputb{\includegraphics[align=c,width=\plotw]{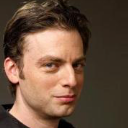}}

\begin{figure*}[t]
	\centering
	\setlength{\tabcolsep}{0pt} 
	\tiny{
	\begin{tabular}{ccccccccccccccccccc}
	& \small{Input} & \small{\makecell{ No \\ smile}} & \small{\makecell{Big \\ nose}} & \small{\makecell{Arched \\ eyebrows}} & \small{\makecell{ Narrowed \\ eyes}}  & \small{\makecell{ No pointy \\ nose}} &
	\small{Input} & \small{Smile} & \small{\makecell{Big \\ nose}} & \small{\makecell{Arched \\ eyebrows}} & \small{\makecell{ Narrowed \\ eyes}}  & \small{\makecell{ No pointy \\ nose}}\\[5pt]
	& & $0.59$ / $0.99$ & $0.36$ / $\mathbf{1.00}$ & $0.79$ / $0.98$ & $0.69$ / $\mathbf{1.00}$ & $0.76$ / $\mathbf{1.00}$ & & $0.58$ / $\mathbf{1.00}$ & $\mathbf{0.76}$ / $0.99$ & $0.29$ / $\mathbf{1.00}$ & $0.61$ / $\mathbf{1.00}$ & $\mathbf{0.86}$ / $0.01$ \\
	\parbox[t]{4mm}{\rotatebox[origin=c]{90}{\scriptsize{StarGAN~\cite{Choi_StarGAN2018}}}} &
	\animatetwo{\inputa}{\inputa} &
	\animatetwo{\includegraphics[align=c,width=\plotw]{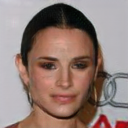}}{\inputa} &
	\animatetwo{\includegraphics[align=c,width=\plotw]{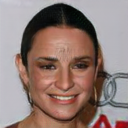}}{\inputa} &
	\animatetwo{\includegraphics[align=c,width=\plotw]{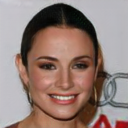}}{\inputa} &
	\animatetwo{\includegraphics[align=c,width=\plotw]{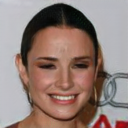}}{\inputa} &
	\animatetwo{\includegraphics[align=c,width=\plotw]{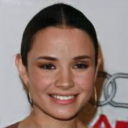}}{\inputa} &
	\hspace{0.1pt} \animatetwo{\inputb}{\inputb} &
	\animatetwo{\includegraphics[align=c,width=\plotw]{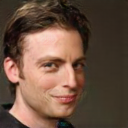}}{\inputb} &
	\animatetwo{\includegraphics[align=c,width=\plotw]{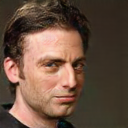}}{\inputb} &
	\animatetwo{\includegraphics[align=c,width=\plotw]{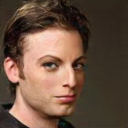}}{\inputb} &
	\animatetwo{\includegraphics[align=c,width=\plotw]{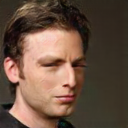}}{\inputb} &
	\animatetwo{\includegraphics[align=c,width=\plotw]{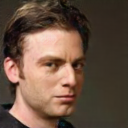}}{\inputb}  \\[19pt]
	& & $0.76$ / $\mathbf{1.00}$ & $0.76$ / $0.99$ & $0.79$ / $\mathbf{0.99}$ & $0.81$ / $0.87$ & $0.84$ / $\mathbf{1.00}$ & & $0.79$ / $\mathbf{1.00}$ & $\mathbf{0.76}$ / $\mathbf{1.00}$ & $0.39$ / $\mathbf{1.00}$ & $\mathbf{0.82}$ / $\mathbf{1.00}$ & $0.85$ / $0.73$ \\
	\parbox[t]{4mm}{\rotatebox[origin=c]{90}{\scriptsize{StarGAN+}}} &
	\animatetwo{\inputa}{\inputa} &
	\animatetwo{\includegraphics[align=c,width=\plotw]{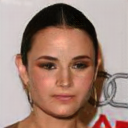}}{\inputa} &
	\animatetwo{\includegraphics[align=c,width=\plotw]{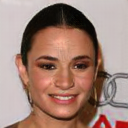}}{\inputa} &
	\animatetwo{\includegraphics[align=c,width=\plotw]{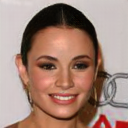}}{\inputa} &
	\animatetwo{\includegraphics[align=c,width=\plotw]{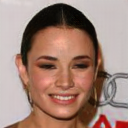}}{\inputa} &
	\animatetwo{\includegraphics[align=c,width=\plotw]{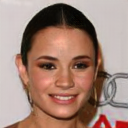}}{\inputa} &
	\hspace{0.1pt} \animatetwo{\inputb}{\inputb} &
	\animatetwo{\includegraphics[align=c,width=\plotw]{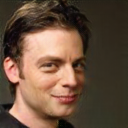}}{\inputb} &
	\animatetwo{\includegraphics[align=c,width=\plotw]{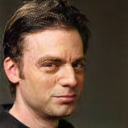}}{\inputb} &
	\animatetwo{\includegraphics[align=c,width=\plotw]{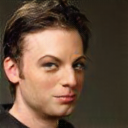}}{\inputb} &
	\animatetwo{\includegraphics[align=c,width=\plotw]{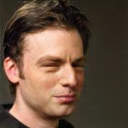}}{\inputb} &
	\animatetwo{\includegraphics[align=c,width=\plotw]{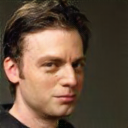}}{\inputb}  \\[19pt]
	& & $\mathbf{0.83}$ / $0.97$ & $\mathbf{0.77}$ / $\mathbf{1.00}$ & $\mathbf{0.89}$ / $\mathbf{0.99}$ & $\mathbf{0.90}$ / $\mathbf{1.00}$ & $\mathbf{0.87}$ / $\mathbf{1.00}$ & & $\mathbf{0.86}$ / $0.94$ & $0.72$ / $\mathbf{1.00}$ & $\mathbf{0.65}$ / $0.99$ & $0.81$ / $0.88$ & $0.85$ / $\mathbf{0.75}$ \\
	\parbox[t]{4mm}{\rotatebox[origin=c]{90}{\scriptsize{\textbf{WarpGAN+}}}} &
	\animatetwo{\inputa}{\inputa} &
	\animatetwo{\includegraphics[align=c,width=\plotw]{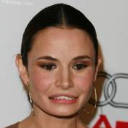}}{\inputa} &
	\animatetwo{\includegraphics[align=c,width=\plotw]{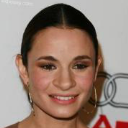}}{\inputa} &
	\animatetwo{\includegraphics[align=c,width=\plotw]{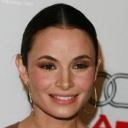}}{\inputa} &
	\animatetwo{\includegraphics[align=c,width=\plotw]{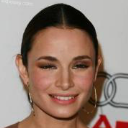}}{\inputa} &
	\animatetwo{\includegraphics[align=c,width=\plotw]{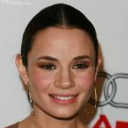}}{\inputa} &
	\hspace{0.1pt} \animatetwo{\inputb}{\inputb} &
	\animatetwo{\includegraphics[align=c,width=\plotw]{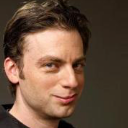}}{\inputb} &
	\animatetwo{\includegraphics[align=c,width=\plotw]{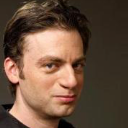}}{\inputb} &
	\animatetwo{\includegraphics[align=c,width=\plotw]{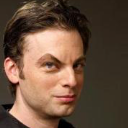}}{\inputb} &
	\animatetwo{\includegraphics[align=c,width=\plotw]{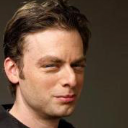}}{\inputb} &
	\animatetwo{\includegraphics[align=c,width=\plotw]{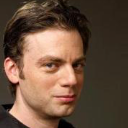}}{\inputb}
	\end{tabular}
	}
	\caption{Comparison to previous work on the CelebA dataset.
	For a given input image, first column, each method attempts to transfer the semantic attribute in its corresponding column.
	A re-identification score and attribute probability are shown as (id / cls) on top of each image (higher is better).
	Our approach edits the attributes of the input images while better preserving the identity of the subject.\animatedfootnote}
	\label{fig:celeba_comparison}
\end{figure*}
\def\plotw{0.095\linewidth}
\def\imga{0}
\def\imgb{16}
\def\imgc{24}
\def\imgd{21}

\def\imgloc{rebuttal/img/cub_200}

\begin{figure*}[t!]
	\centering
	\setlength{\tabcolsep}{1pt} 
	\scriptsize{
		\begin{tabular}{ccccc | ccccc } 
			& \multicolumn{2}{c}{StarGAN~\cite{Choi_StarGAN2018}} & \multicolumn{2}{c}{\textbf{WarpGAN}} & & \multicolumn{2}{c}{StarGAN~\cite{Choi_StarGAN2018}} & \multicolumn{2}{c}{\textbf{WarpGAN}} \\ \cmidrule(lr){2-3} \cmidrule(lr){4-5} \cmidrule(lr){7-8} \cmidrule(lr){9-10}
			Input & \makecell{Beak smaller \\ than head} & \makecell{Beak larger \\ than head} & \makecell{Beak smaller \\ than head} & \makecell{Beak larger \\ than head} & Input & \makecell{Beak smaller \\ than head} & \makecell{Beak larger \\ than head} & \makecell{Beak smaller \\ than head} & \makecell{Beak larger \\ than head} \\
			\includegraphics[width=\plotw]{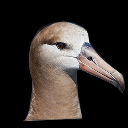} &
			\includegraphics[width=\plotw]{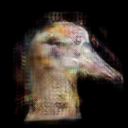} &
			\includegraphics[width=\plotw]{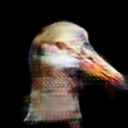} &
			\includegraphics[width=\plotw]{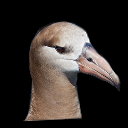} &
			\includegraphics[width=\plotw]{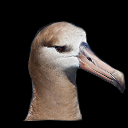} &
			\scalebox{-1}[1]{\includegraphics[width=\plotw]{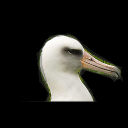}} &
			\scalebox{-1}[1]{\includegraphics[width=\plotw]{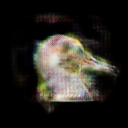}} &
			\scalebox{-1}[1]{\includegraphics[width=\plotw]{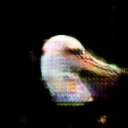}} &
			\scalebox{-1}[1]{\includegraphics[width=\plotw]{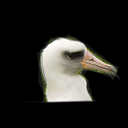}} &
			\scalebox{-1}[1]{\includegraphics[width=\plotw]{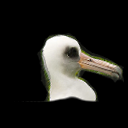}} \\
			\includegraphics[width=\plotw]{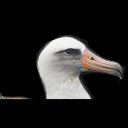} &
			\includegraphics[width=\plotw]{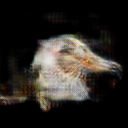} &
			\includegraphics[width=\plotw]{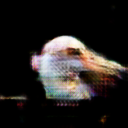} &
			\includegraphics[width=\plotw]{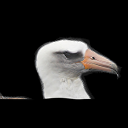} &
			\includegraphics[width=\plotw]{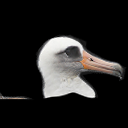} &
			\includegraphics[width=\plotw]{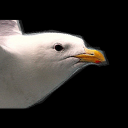} &
			\includegraphics[width=\plotw]{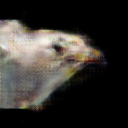} &
			\includegraphics[width=\plotw]{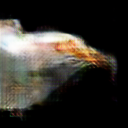} &
			\includegraphics[width=\plotw]{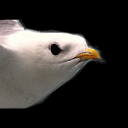} &
			\includegraphics[width=\plotw]{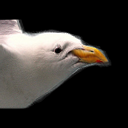} 
		\end{tabular}
	}
	\caption{Comparison to previous work on the Cub200 dataset.
	Each model attempts to transfer the attribute (relative beak size) in each column to the input image.
	StarGAN is unable to produce good quality edits on this dataset, while our model edits the input images in a more plausible manner.
	Due to the more complex nature of this dataset, our model still struggles to produce artifact free transformations.
	Results from this model at the original image resolution, and without masking the background, can be found in the supplemental material.}
	\label{fig:warp:cub200}
\end{figure*}


\section{Results}

\subsection{Datasets}
We evaluate our method and baselines 
on a face dataset, CelebA~\cite{Liu2015CelebA} and a birds dataset, Cub200~\cite{Wah2011CUB_200}.
\paragraph{CelebA}
The CelebA~\cite{Liu2015CelebA} dataset contains 202,599 images of faces and we use the train/test split recommended by the authors.
Importantly, from the 40 binary attributes provided, we choose the ones more amenable to be characterized by warping, namely: smiling, big nose, arched eyebrows, narrow eyes and pointy nose.

\paragraph{Cub200}
The Cub200~\cite{Wah2011CUB_200} dataset contains 11,788 images of 200 bird species. 
The images are annotated with 312 binary attributes and a semantic mask of the bird body. 
We choose the three binary attributes that correspond to the beak size relative to the head and remove the background using the semantic masks.
The train/test split recommended by the authors is employed. 
Due to the alignment step discussed below, only 2,325 images are used for training.


\paragraph{Face alignment}
For both datasets, we make use of face landmark locations to align and resize the images to 128$\times$128 using a global affine transformation, at both training and test time.
At test time, the inverse of the affine transformation is used to transform the warp fields.
The warp is then applied directly to the original image.
This is in contrast with previous methods, that would edit the aligned image and then warp the edited image to the original space.
For images outside of the test set, off-the-shelf methods~\cite{King2009Dlib} can be used to align them to the dataset.

\subsection{Models}
Our main baseline is StarGAN~\cite{Choi_StarGAN2018}, a state of the art model for image-to-image translation. 
We define three novel models to evaluate our contributions.
WarpGAN denotes models that output a warp field.
A ``+'' suffix indicates models that employ our binary labels transformation scheme.
Thus, \mbox{StarGAN+} evaluates the effect of label transformation, and WarpGAN+ is our final proposed model.


An obvious alternative to our model consists of fitting a dense flow field to the results generated by StarGAN. 
We tested it using the dense optical flow matching technique described in~\cite{Zach2007OpenCVFlow}, and we denote this method by SGFlow.

An example of SGFlow is shown in Fig.~\ref{fig:celeba_sg_plus_flow}, using optical flow~\cite{Zach2007OpenCVFlow}.
Warping based on optical flow may lead to artefacts when good correspondences are not found.
Constraining StarGAN to generate images that are amenable to optical flow estimation is not trivial.
Hence, this experiment shows that a na\"ive approach for applying the result of a StarGAN model to a higher resolution image is suboptimal.
Thus, we drop SGFlow for the remaining experiments.

We also experimented with the GANimation~\cite{Pumarola_Ganimation2018} approach using the code provided by the authors.
However we were unable to generate meaningful results when training the method with binary attributes.
We suspect that this is due to the method's reliance on soft action unit labels.

\paragraph{Hyper-parameters}
All models were trained on a single Titan X GPU using \deleted{the }TensorFlow~\cite{Abadi2015Tensorflow}\deleted{ framework}.
\added{The models hyper-parameters are: $\lambda_{\text{cls}}=2$, $\lambda_{\text{gp}}=10$, $\lambda_{\text{c}}=10$ and $\lambda_{\text{s}}=125$.}%
\deleted{The Adam optimizer~\cite{Kingma2015Adam} is used with a learning rate of $0.0001$, with $\beta_1 = 0.5$ and $\beta_2 = 0.999$.
The model hyper-parameters are shared for all datasets: $\lambda_{\text{cls}}=2$, $\lambda_{\text{gp}}=10$, $\lambda_{\text{c}}=10$ and $\lambda_{\text{s}}=125$.}

For the StarGAN baseline we employ the implementation provided by the authors, where we keep all their recommended hyper-parameters except for $\lambda_{\text{cls}}=0.25$.
The choice of $\lambda_{\text{cls}}$, for StarGAN and our models, is informed by the results shown in Fig.~\ref{fig:attr_vs_identity}.
Additional implementation details are provided in the supplementary material. 

%
%
%
%

\subsection{Qualitative results}

We show qualitative results on the CelebA dataset in Fig.~\ref{fig:celeba_comparison}.
For each input image, we show the edited images corresponding to changing a single attribute.
StarGAN~\cite{Choi_StarGAN2018} often changes characteristics that are not related to the perturbed attributes, such as the skin tone or the background color.
StarGAN+ produces more localized edits than StarGAN.
\added{The WarpGAN+ edit for the ``no smile'' attribute is not particularly realistic.
However, for most edits, our technique generates changes that are less exaggerated and better preserve the identity of the subject.}


Qualitative results for the masked and aligned Cub200 dataset are shown in Fig.~\ref{fig:warp:cub200}.
Our approach is able to transfer the corresponding attribute, albeit sometimes producing unrealistic additional deformations.
The poor quality results of StarGAN may be attributed to the increased complexity of this dataset and the reduced number of images, compared to CelebA.
This is a generous comparison, as the predicted warp fields can be applied to the original images with complex backgrounds and at higher resolutions, as shown in the supplemental material.
Fig.~\ref{fig:teaser} demonstrates the power of the warping representation by operating at a far higher resolution (2480$\times$1850) than can be achieved by direct methods.

Please see the supplemental material for animated edits and additional results\added{, which also include more examples of failure cases}. 

\def\plotw{0.2\linewidth}
\def\imagea{8}
\def\imageb{15}

\begin{figure}[tbp!]
	\centering
	\setlength{\tabcolsep}{2pt} 
	\setlength{\fboxsep}{0pt} 
	\setlength{\fboxrule}{1.5pt} 
	\scriptsize{
	\begin{tabular}{cccc}
	$\alpha = -0.25$ & $\alpha = 0$ & $\alpha = 0.2$ & $\alpha = 0.4$ \\[-1pt] 
	\includegraphics[width=\plotw]{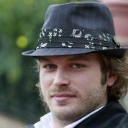} & 
	\fcolorbox{red}{white}{\includegraphics[width=\plotw]{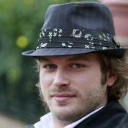}} & 
	\includegraphics[width=\plotw]{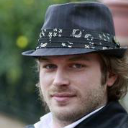} &
	\includegraphics[width=\plotw]{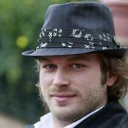} \\[2pt]
	$\alpha = 0.6$ & $\alpha = 0.8$ & $\alpha = 1.0$ & $\alpha = 1.25$ \\[-1pt] 
	\includegraphics[width=\plotw]{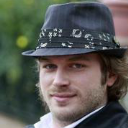} &
	\includegraphics[width=\plotw]{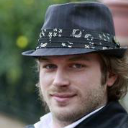} &
	\fcolorbox{green}{white}{\includegraphics[width=\plotw]{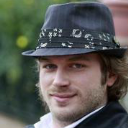}} & 
	\includegraphics[width=\plotw]{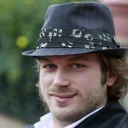} \\[-1pt] 
	\end{tabular}
	} 
	\caption{Partial editing with our model, for the ``smile'' attribute.
	A single warp is generated by our model, which is interpolated and extrapolated by scaling the magnitude of its values by $\alpha$.
	The input image, $\alpha \hspace{-1.5pt} = \hspace{-1.5pt} 0$, is progressively edited in both directions.\animatedfootnote}
	\label{fig:interpolation}
\end{figure}

\def\ploth{1.8cm}
\def\imagea{3}
\def\imageaattra{2}

\begin{figure}[tp]
	\centering
	\setlength{\tabcolsep}{1pt} 
	\small{
	\begin{tabular}{cccccc}
	& Input & \makecell{Arched \\ eyebrows} & \makecell{Stretch \\ map} & Overlay & \\
	\parbox[t]{3mm}{\rotatebox[origin=c]{90}{\scriptsize{WarpGAN}}} &
	\includegraphics[align=c,height=\ploth]{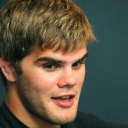} &
	\includegraphics[align=c,height=\ploth]{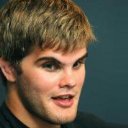} &
	\includegraphics[align=c,height=\ploth]{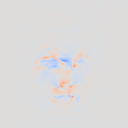} &
	\includegraphics[align=c,height=\ploth]{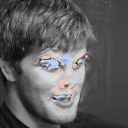} &
	\includegraphics[align=c,height=\ploth]{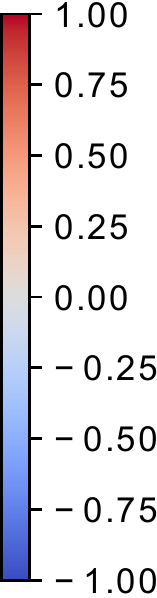} \\[22pt]
	\parbox[t]{3mm}{\rotatebox[origin=c]{90}{\scriptsize{WarpGAN+}}} &
	\includegraphics[align=c,height=\ploth]{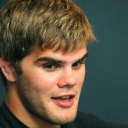} &
	\includegraphics[align=c,height=\ploth]{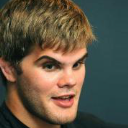} &
	\includegraphics[align=c,height=\ploth]{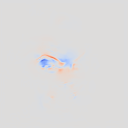} &
	\includegraphics[align=c,height=\ploth]{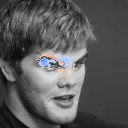} &
	\includegraphics[align=c,height=\ploth]{img/rafd_stretch/lr/colorbar} \\
	\end{tabular}
	}
	\caption{Stretch maps computed from the warp fields, for both WarpGAN and WarpGAN+.
	The log determinant of the Jacobian of the  warp is shown, where blue indicates stretching and red corresponds to squashing.
	Our binary labels transformation scheme, used by WarpGAN+, leads to correctly localized edits. 
	}
	\label{fig:celeba_stretch_example}
\end{figure}


%

\paragraph{Partial edits} Another advantage of our model is that once a warp field has been computed for an input image, partial edits can be applied by simply scaling the predicted displacement vectors by a scalar, $\alpha$. 
Results of interpolation and extrapolation of warp fields generated by our model are shown in Fig.~\ref{fig:interpolation}.
This is a cheap operation as it does not require to run the network for each new value of $\alpha$, in contrast with previous methods that allow for partial edits~\cite{Pumarola_Ganimation2018}; this allows for edits to be performed at interactive speeds. 

\paragraph{Visualizing warp fields} A further advantage of our model is the interpretability of its edits.
This is demonstrated in Fig.~\ref{fig:celeba_stretch_example}, where we show the log determinant of the Jacobian of the warp field, which illustrates image squashing and stretching.
It can be seen how employing our binary label transformation scheme leads to more localized edits.
Moreover, the values from the stretch maps can potentially be used to automatically determine which areas have been stretched or compressed excessively by the network.
Thus they provide an intuitive measure to detect unrealistic edits. 




\subsection{Quantitative results}

Quantitative evaluation is challenging for our setting.
We provide two methodologies: the first measures the model performance based on separately trained networks, and the second is a user study to estimate perceptual quality.
\paragraph{Accuracy vs identity preservation} We train a separate classifier on the training data, to estimate quantitatively 
if the edited images have the requested attributes.
The classifier has the same architecture as the discriminator and is trained with the cross entropy loss of eq.~\ref{eq:C_d}.
We also use a pretrained face re-identification model~\cite{Schroff2015FaceNet} to evaluate whether the edits preserve the identity.\footnote{Additional details can be found in the supplemental material.\label{ft:supplement_details}}

Results of both experiments are shown in Fig.~\ref{fig:attr_vs_identity}, where an ideal editing model would be located on the top-right.
On the $x$-axis we show the rate of images classified as having the target attribute (attribute accuracy), defined as $\frac{1}{m}\sum [C(\bx, \bct) \geq 0.5]$, where $m$ is the total number of images.
On the $y$-axis, an identity preservation score is shown, which is evaluated as $1 - \frac{1}{m}\sum d(\bx, \bxt)$, where $d(\cdot)$ is the L2 distance between the input and the edited image in the feature space of the face re-identification network.
A distance larger than $1.2$ (score lower than $-0.2$) has been used to indicate that two faces belong to different people~\cite{Schroff2015FaceNet}.

There is a trade-off between attribute transfer and identity score.
On one extreme, a new face that has the target attribute and does not match the original face would achieve maximal attribute accuracy with negative identity score.
On the other, not modifying the image has maximal identity score, yet it does not achieve the target edit.
With respect to StarGAN, our binary labels transformation scheme (StarGAN+) moves the curve towards higher attribute transfer with comparable identity score.
Our warping approach (WarpGAN+) allows for stronger identity preservation than StarGAN+.
Overall, our approach better preserves identity than previous work, for similar levels of attribute transfer.
Moreover, we picked $\lambda_{\text{cls}}$ based on these results: choosing the value that leads to both high accuracy and identity score.




\def\plotw{0.65\linewidth}

\begin{figure}[tbp!]
	\centering
	\includegraphics[width=\plotw]{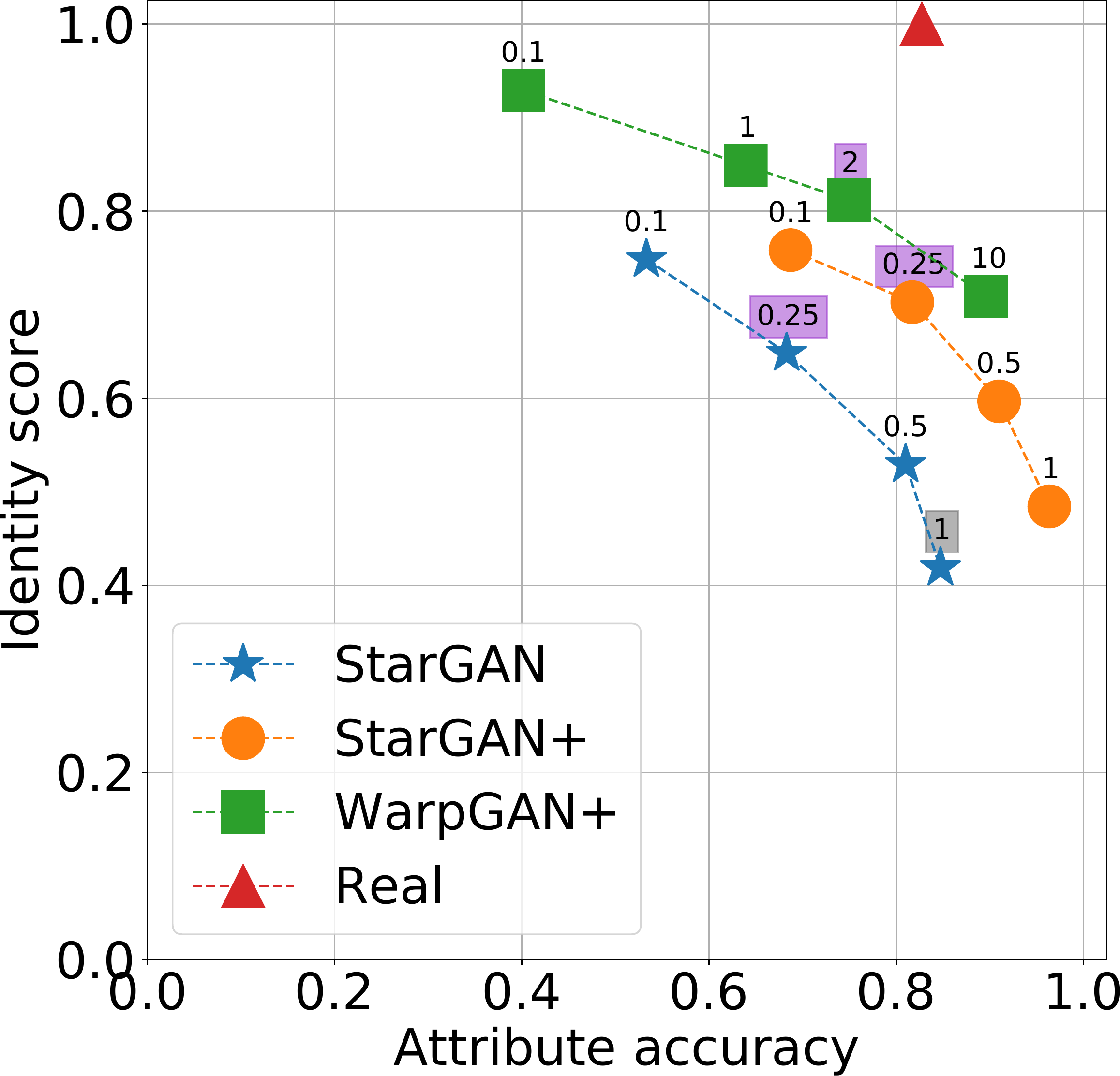}
	\caption{Presence of the edited attribute ($x$-axis) vs face re-identification score ($y$-axis), higher is better.
	The classification loss weight, $\lambda_{\text{cls}}$, is shown on top of each marker.
	Highlighted in gray is the value used by the StarGAN authors for this dataset, and in purple the ones used in this paper.
	Compared to previous work, our model produces edits that better preserve identity.}
	\label{fig:attr_vs_identity}
\end{figure}

\paragraph{Accuracy vs realism} We perform a user study on Amazon Mechanical Turk (MTurk) to evaluate the quality of the generated images, for StarGAN, StarGAN+ and our model.
For each method, we use the same 250 test images from CelebA  and edit the same attribute per image.

We conducted two experiments, one to evaluate the realism of the images, where the workers had to answer whether the image presented was real or fake, and another to evaluate attribute editing, where we asked the workers whether the image contains the target attribute.
In both experiments, the workers were randomly shown a single image at a time: either an edited image or an unaltered original image.\textsuperscript{\ref{ft:supplement_details}}


\def\plotw{0.65\linewidth}

\begin{figure}[tbp!]
	\centering
	\includegraphics[width=\plotw]{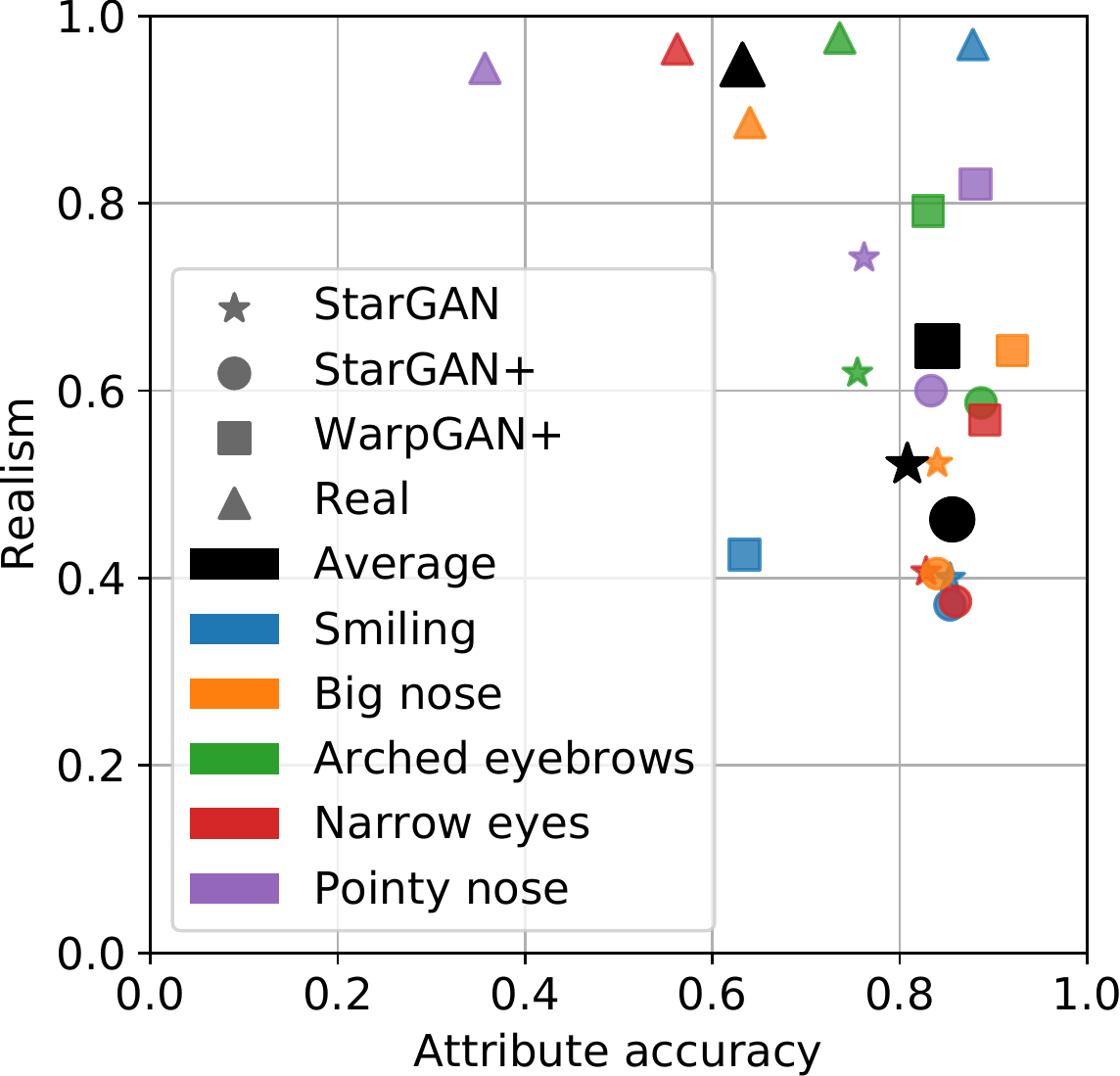}
	\caption{Human perception of the presence of the desired attribute (attribute accuracy) vs realism of the image, as indicated by the user study (higher is better).
	Images generated by our model are more realistic than those generated by previous work. 
	}
	\label{fig:user_study}
\end{figure}

Results of this user study are shown Fig.~\ref{fig:user_study}.
A useful editing model has a high-level of realism and can produce the target edit.
For the real data, the workers reliably evaluated image realism, however they were often inconsistent with the attribute labels.
Nonetheless, the workers performance on real data should not be taken as an upper bound, as all methods tend to generate exaggerated edits to maximize correct classification.
For the editing models, the attribute accuracy is consistent to that reported by the classifier network in Fig.~\ref{fig:attr_vs_identity}.
However, identity score and realism do not align, as they measure different notions.
An image might contain only small edits, which the identity network is invariant to, yet those edits could include unrealistic artefacts that can be easily detected by humans.
All editing models achieve good attribute transfer accuracy, with room for improvement mostly on the realism axis.
Our model (WarpGAN+) achieves this for most attributes, and it is able to generate images that are  more realistic than previous work.

%

\section{Conclusions}
This paper has introduced a novel way to learn how to perform semantic image edits from unpaired data using warp fields.
We have demonstrated that, despite limitations on the set of edits that can be described using warping alone, there are clear advantages to modelling edits in this way: they better preserve the identity of the subject, they allow for partial edits, they are more interpretable, and they are applicable to arbitrary resolutions.
Moreover, our binary label transformation scheme leads to increased performance, and removes the need to use a classifier during inference.

There are several avenues for future work, including different parametrizations for the warps, \eg in the form of velocity fields~\cite{Ceritoglu2013RegistrationBrain}.
Additional intermediate representations that upsample well could be added to increase the model flexibility, such as local color transformations~\cite{Gharbi_Bilateral2017}.
Also, an inpainting method~\cite{Pathak2016ContextEncoders} could be locally applied in areas that have been warped or stretched excessively, which could be automatically detected using the log determinant of the Jacobian of the warp fields.

%
%
%

\FloatBarrier

\flushcolsend

\newpage

\section*{\added{Acknowledgements}}
\added{This work has been generously supported by Anthropics Technology Ltd., as well as the EPSRC CDE (EP/L016540/1) and CAMERA (EP/M023281/1) grants.}

{\small
\bibliographystyle{ieee_fullname}
\bibliography{bibliography}
}

\onecolumn
\section*{Appendix}
\appendix
\begin{minipage}{\textwidth}

\section{High-resolution Flickr faces}

\def\inputimga{\includegraphics[trim={55cm 5cm 47cm 5cm}, clip, width=0.32\linewidth]{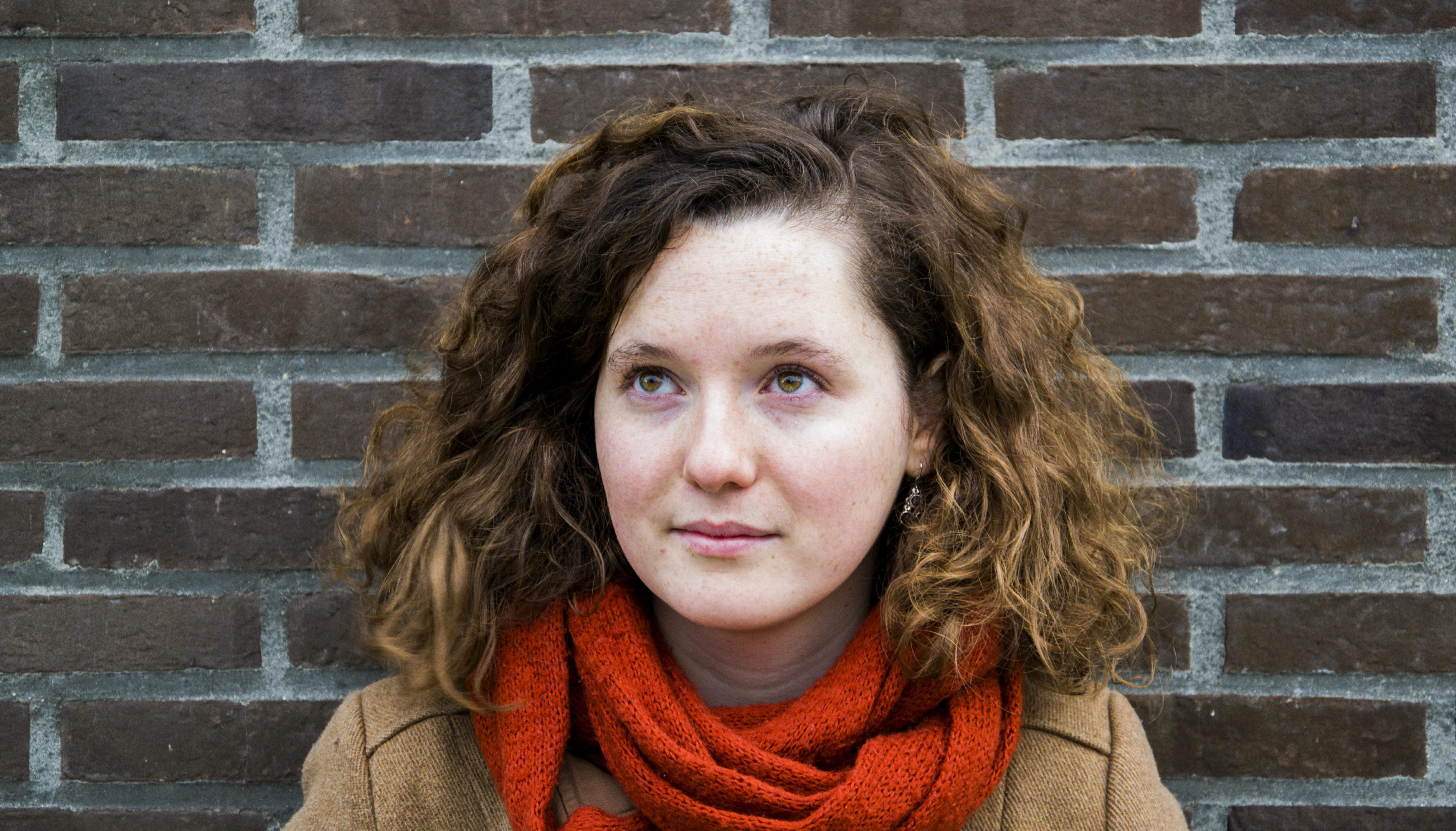}}
\def\inputimgb{\includegraphics[trim={40cm 0cm 45cm 0cm}, clip, width=0.32\linewidth]{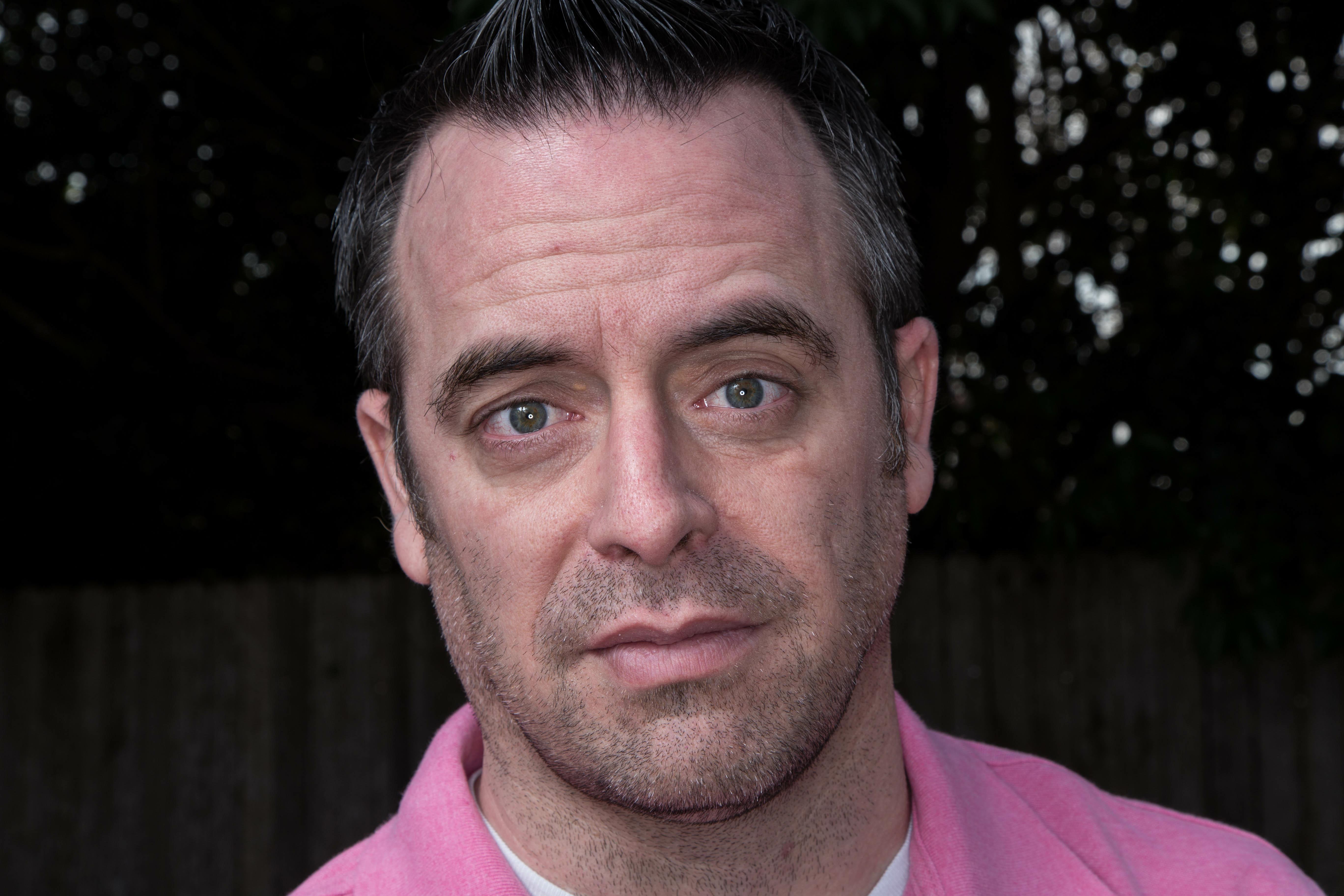}}

\begin{minipage}{\linewidth}
	\centering
	\setlength{\tabcolsep}{1pt} 
	\small{
	\begin{tabular}{ccc}
	Input & Big nose & Smiling \\
	\includegraphics[trim={55cm 5cm 47cm 5cm}, clip, width=0.32\linewidth]{sup_imgs/flickr/1/input} &
	\animatetwo{\includegraphics[trim={55cm 5cm 47cm 5cm}, clip, width=0.32\linewidth]{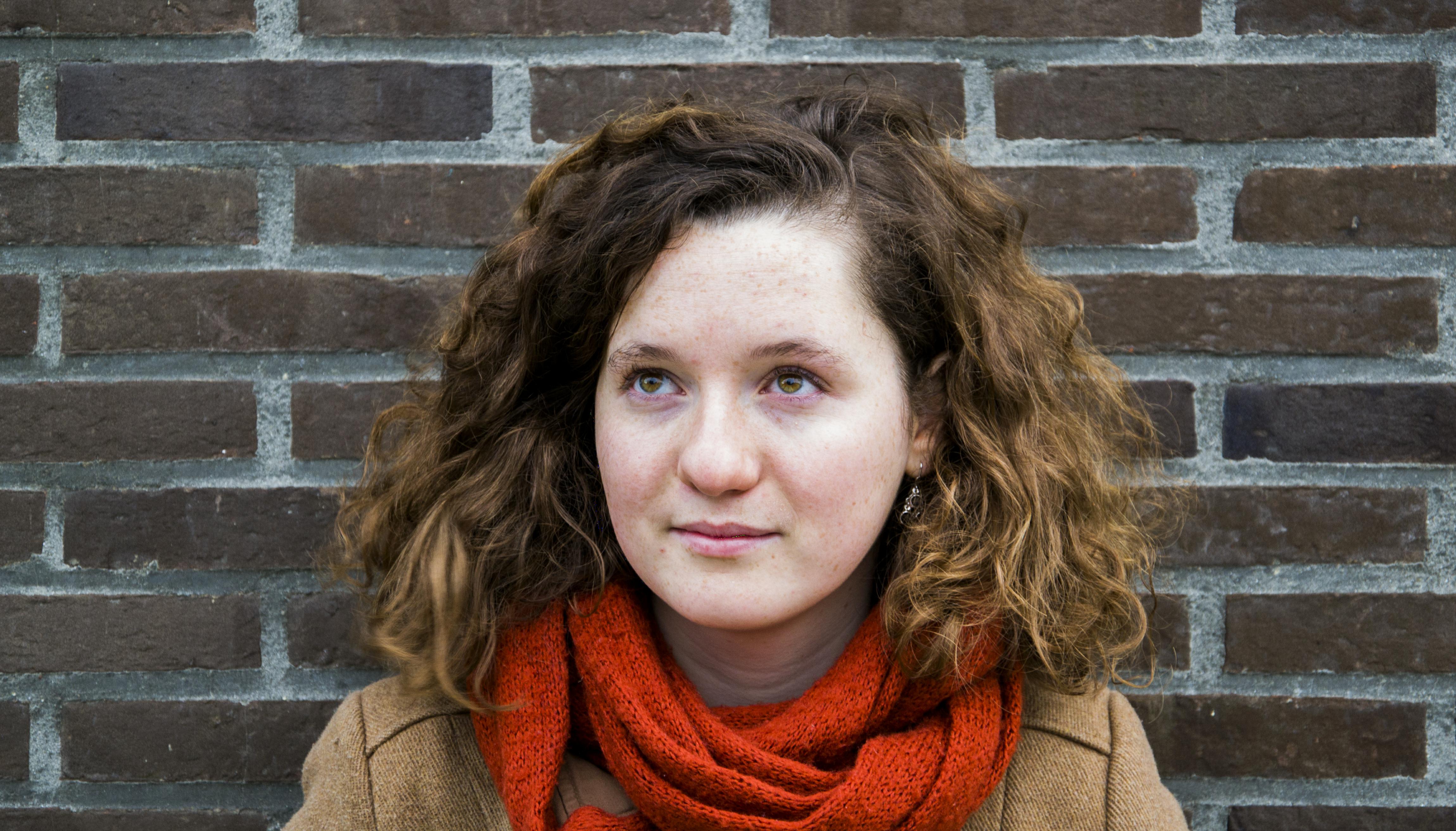}}{\inputimga} &
	\animatetwo{\includegraphics[trim={55cm 5cm 47cm 5cm}, clip, width=0.32\linewidth]{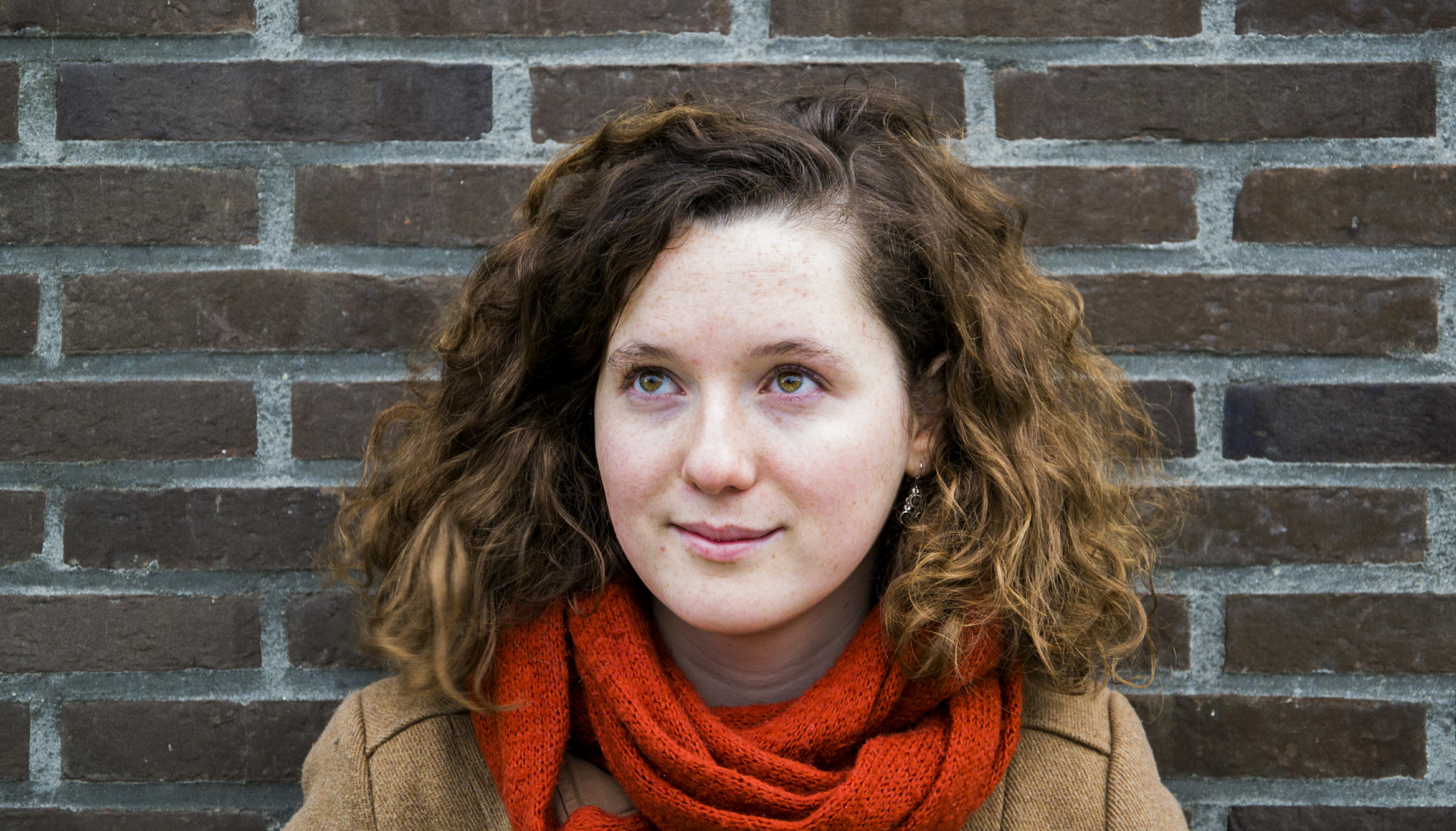}}{\inputimga} \\[20pt]
	Input & Big nose & Arched Eyebrows \\
	\includegraphics[trim={40cm 0cm 45cm 0cm}, clip, width=0.32\linewidth]{sup_imgs/flickr/5/input} &
	\animatetwo{\includegraphics[trim={40cm 0cm 45cm 0cm}, clip, width=0.32\linewidth]{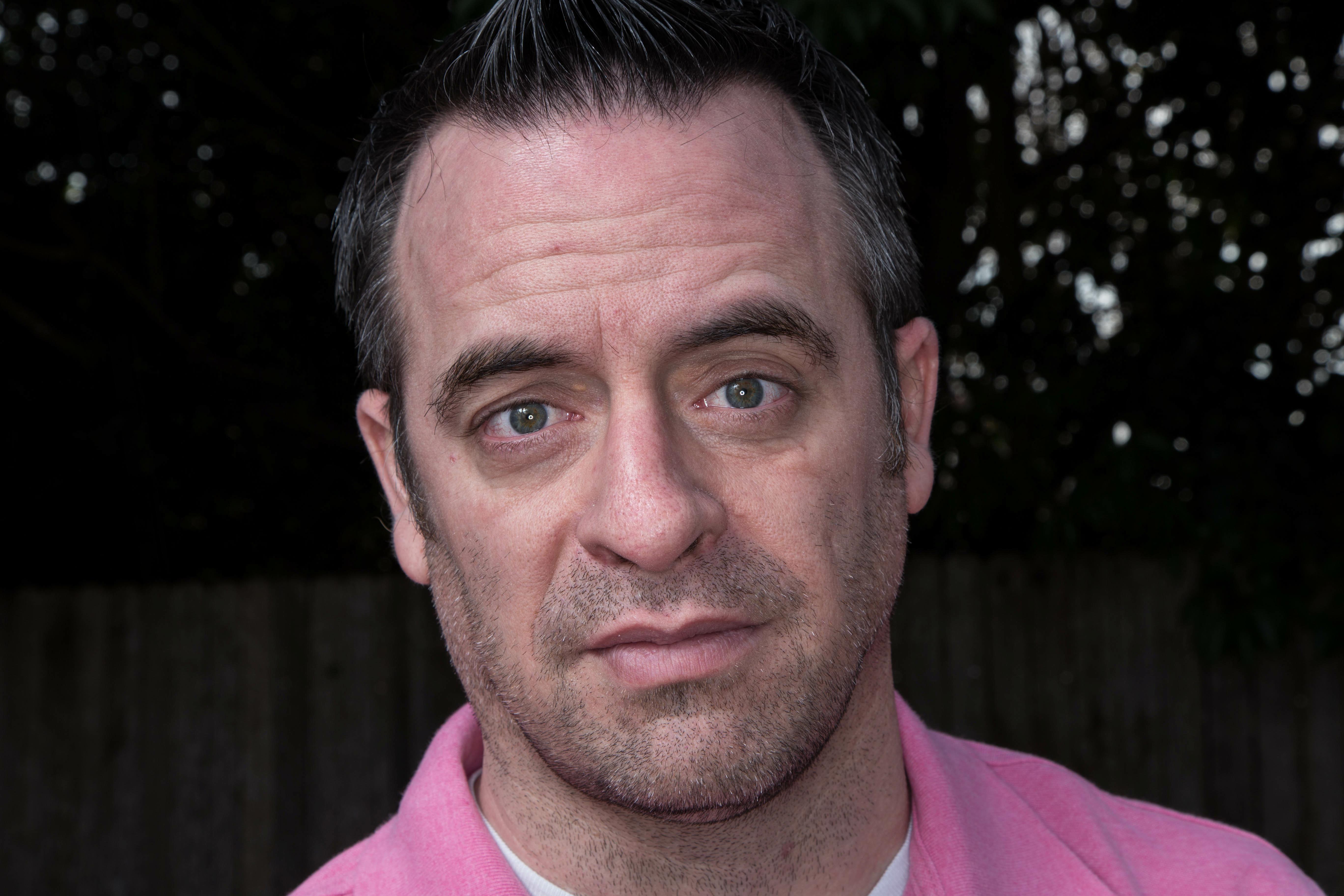}}{\inputimgb} &
	\animatetwo{\includegraphics[trim={40cm 0cm 45cm 0cm}, clip, width=0.32\linewidth]{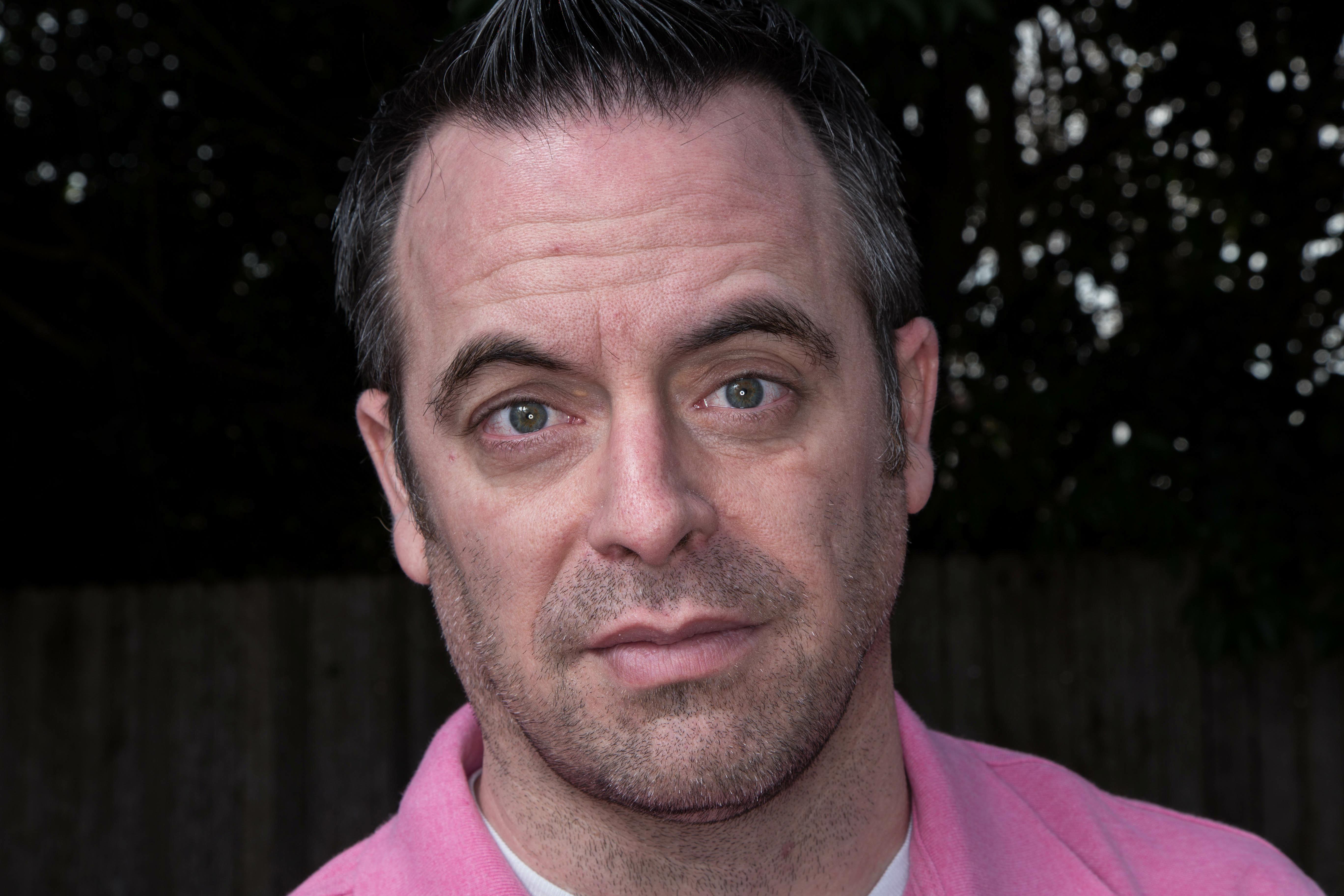}}{\inputimgb}
	\end{tabular}
	}
	\captionof{figure}{Additional results of our model on high-resolution images.
	Our model predicts warps at low resolution that can then be resized and applied to high resolution images.
	The model is able to keep the content and identity at high resolution.
	Please see supplemental videos demonstrating animated edits.
	Input images courtesy of Flickr users Kenneth DM and Randall Pugh.
	(Zoom in for details)}
	\label{fig:hr_edits}
\end{minipage}%

\end{minipage}

\FloatBarrier
\newpage
\section{High-resolution Flickr birds}
\def\plotw{0.32\linewidth}

\begin{figure}[H]
	\centering
	\setlength{\tabcolsep}{1pt} 
		\begin{tabular}{ccc}
			\multicolumn{3}{c}{HR results} \\
			Input & Beak longer than head  & Beak shorter than head \\
			\includegraphics[trim={1260px 350px 700px 0px}, clip, width=\plotw]{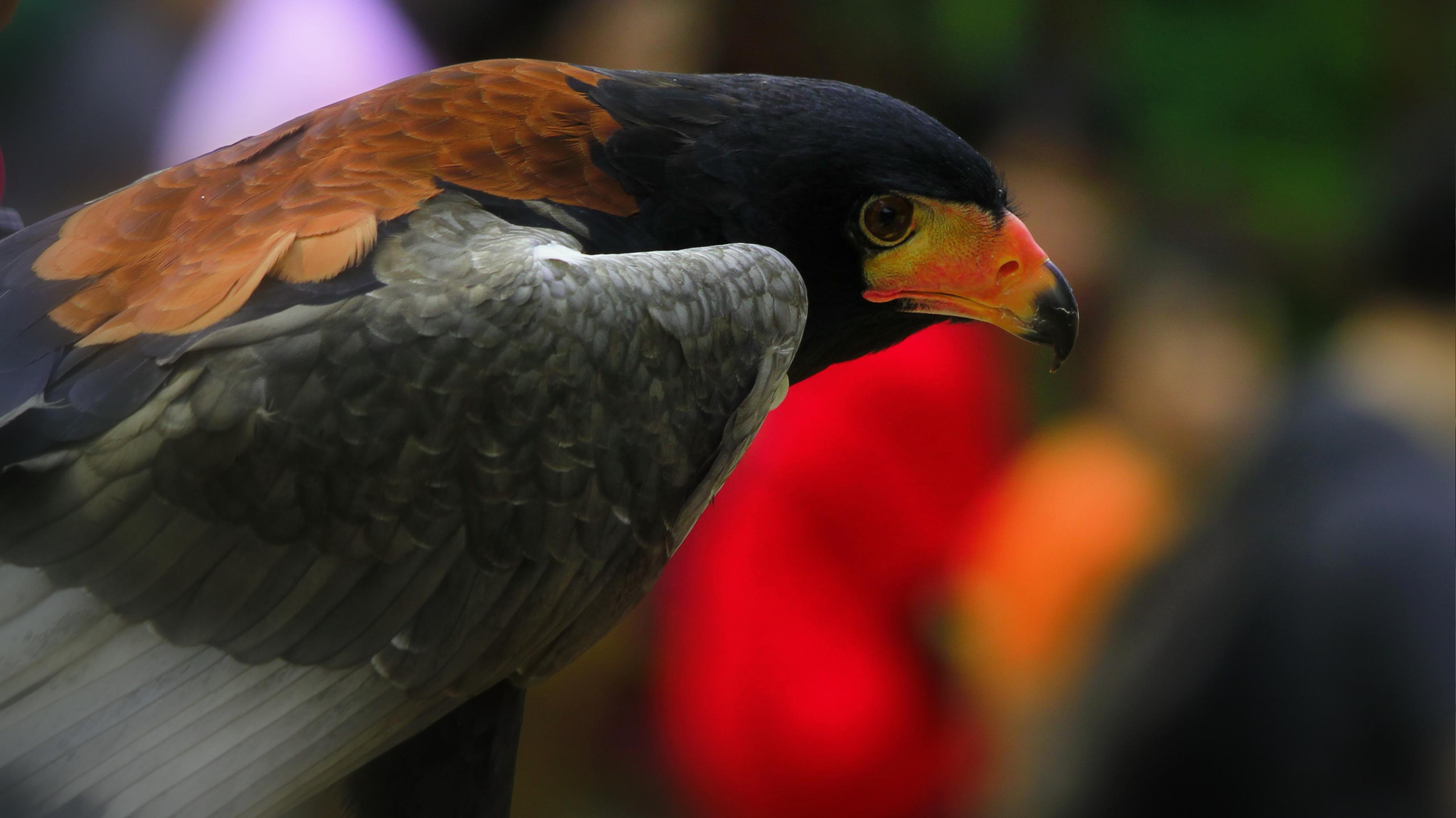} &
			\includegraphics[trim={1260px 350px 700px 0px}, clip,width=\plotw]{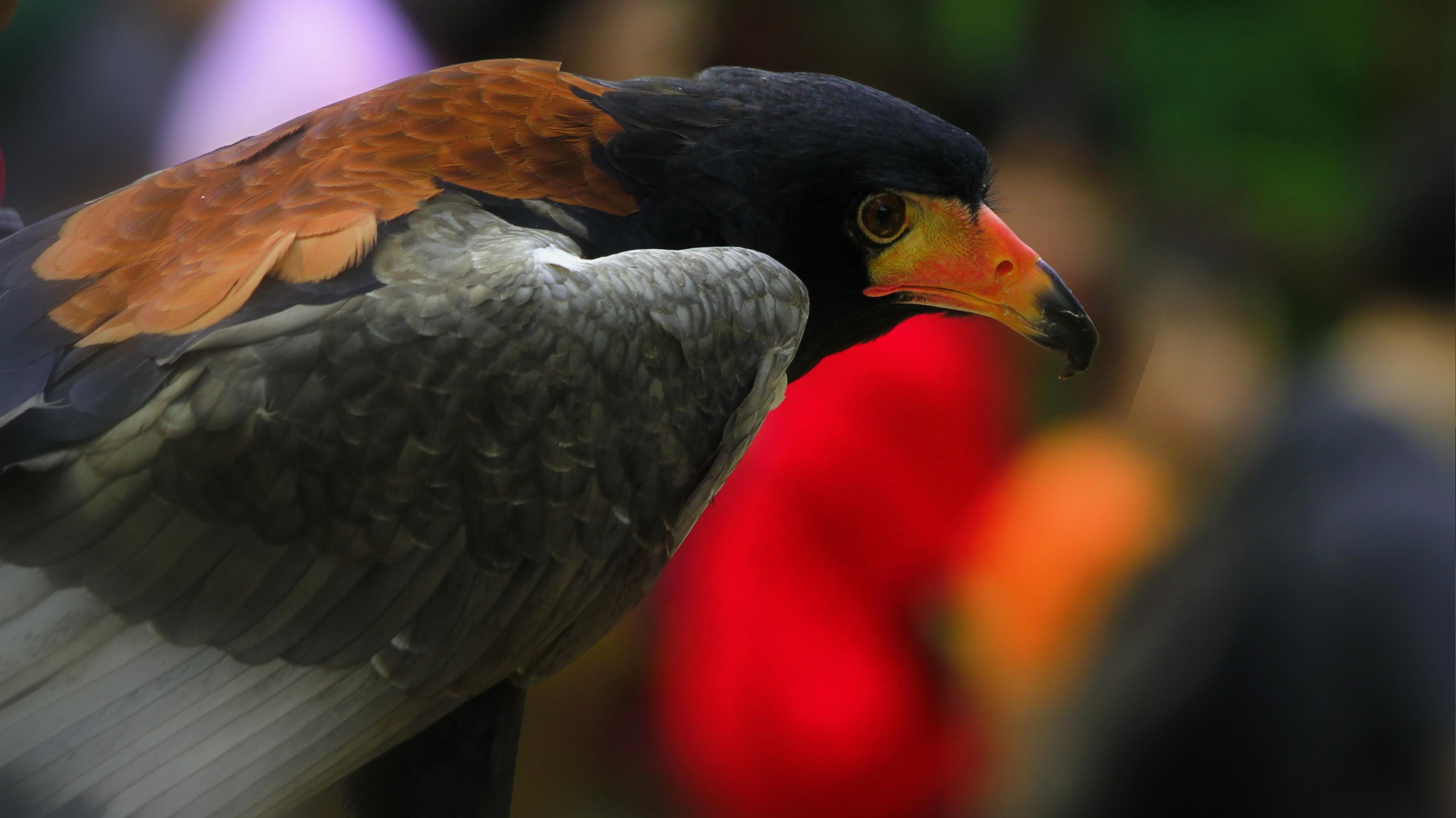} &
			\includegraphics[trim={1260px 350px 700px 0px}, clip,width=\plotw]{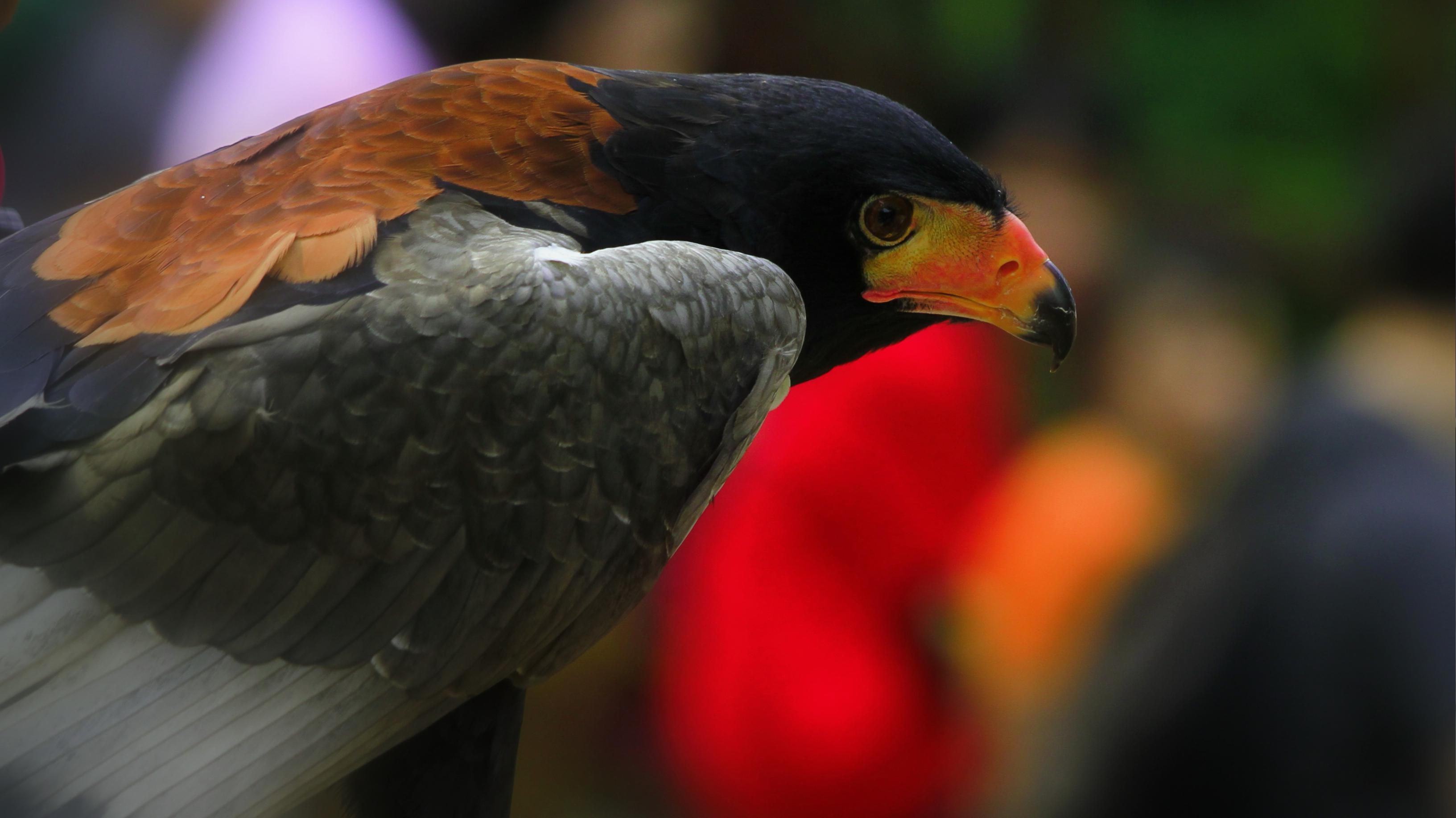} \\
			\scalebox{-1}[1]{\includegraphics[width=\plotw]{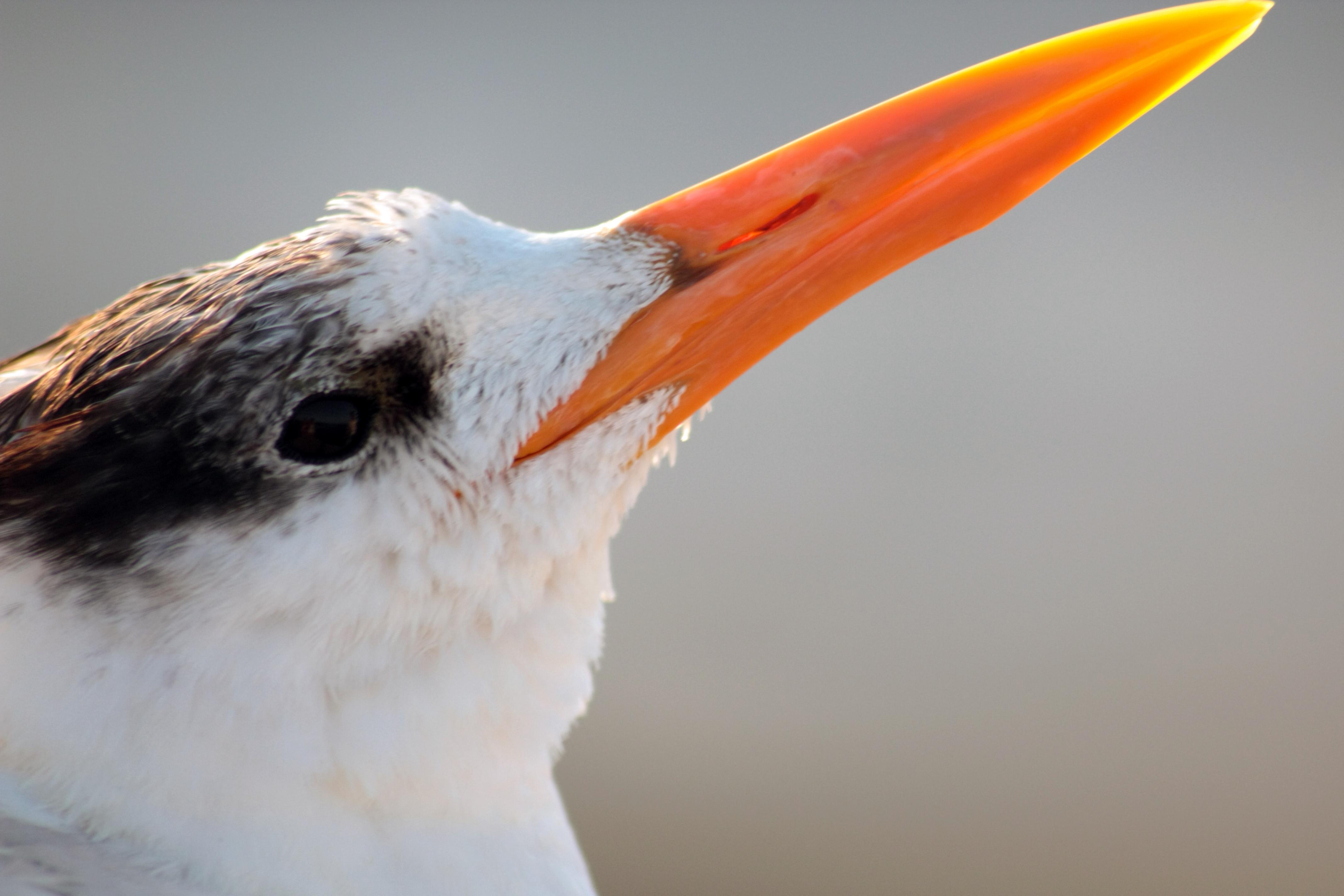}} &
			\scalebox{-1}[1]{\includegraphics[width=\plotw]{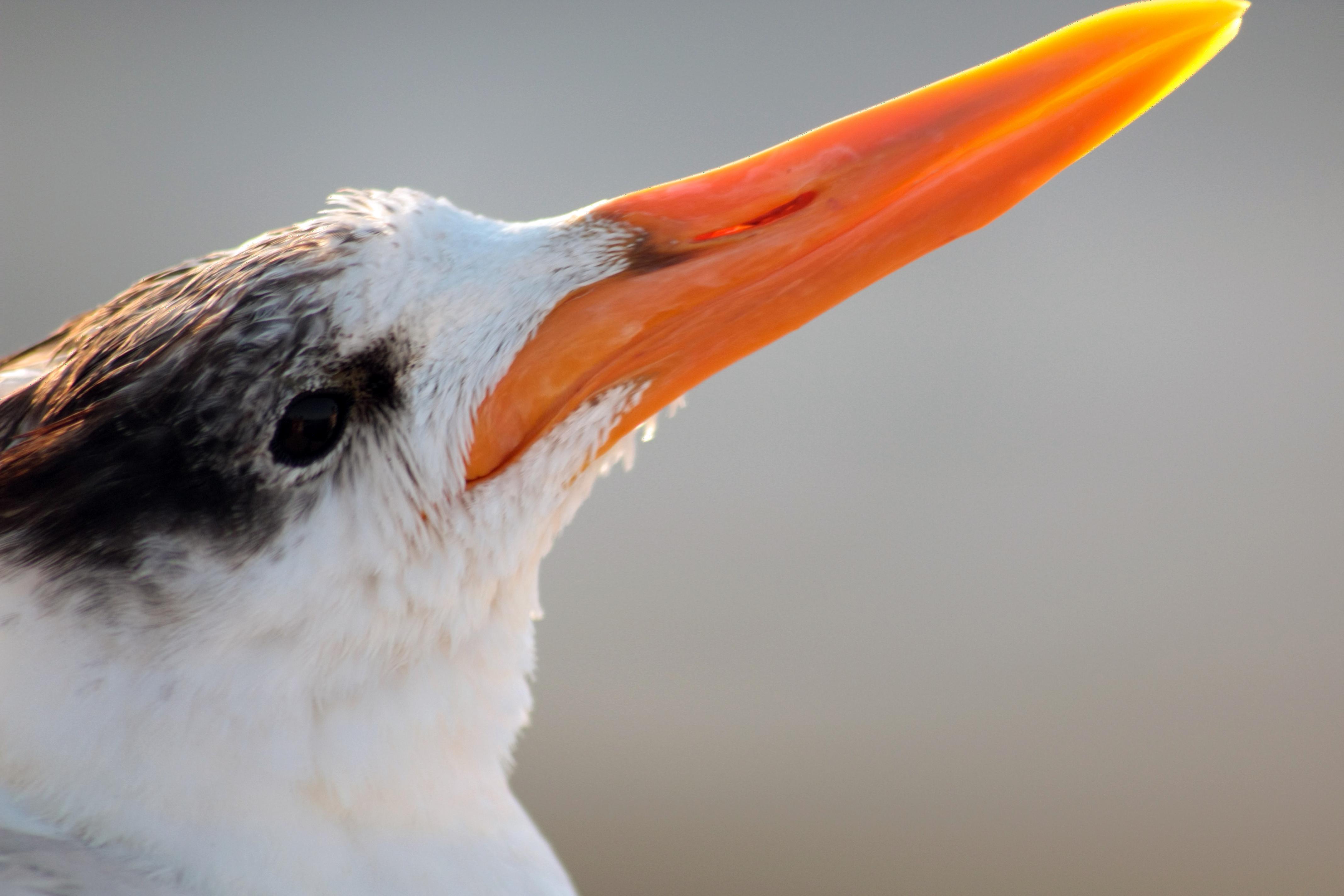}} &
			\scalebox{-1}[1]{\includegraphics[width=\plotw]{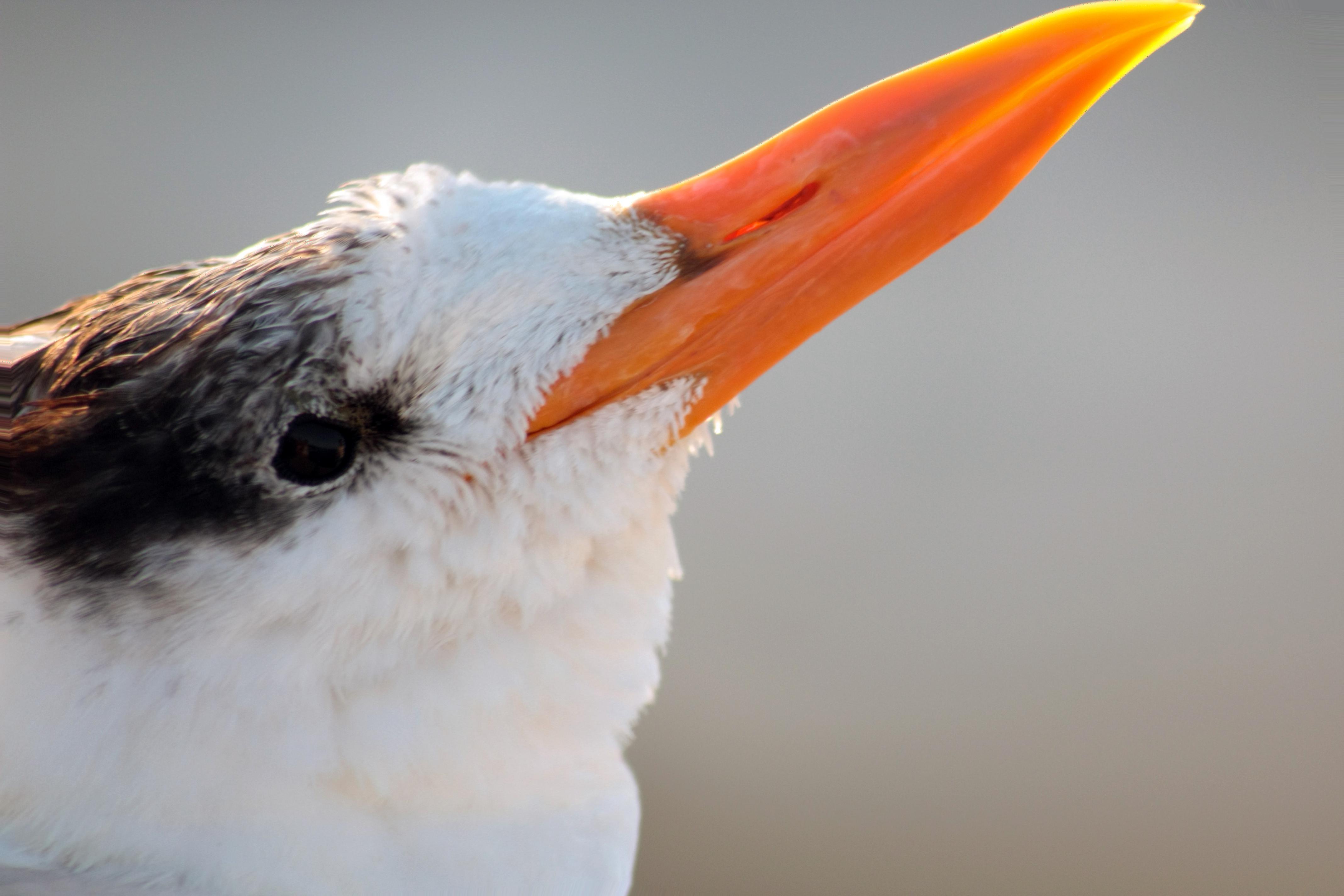}} \\
			\includegraphics[trim={500px 0cm 500px 0cm}, clip,width=\plotw]{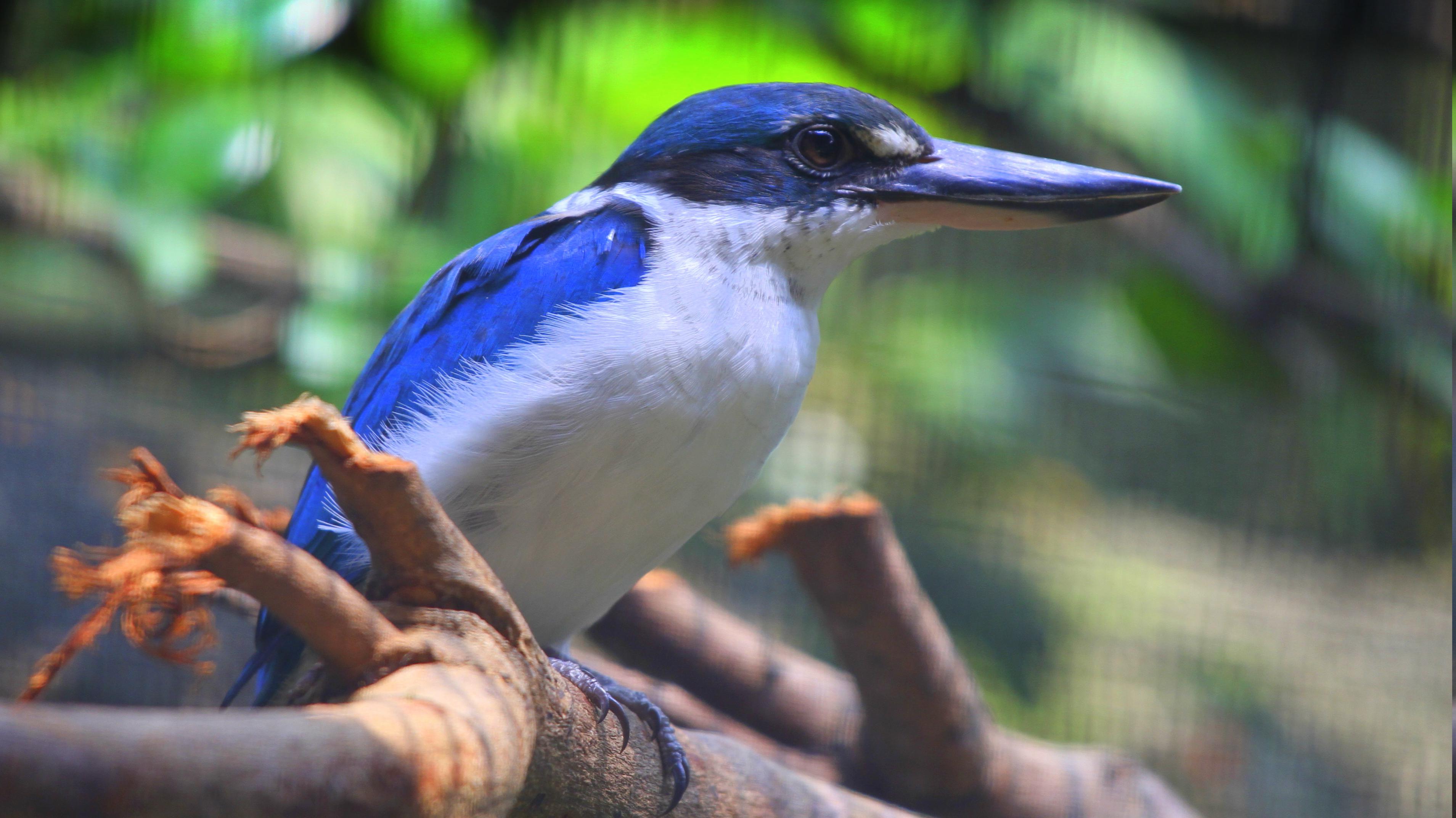} &
			\includegraphics[trim={500px 0cm 500px 0cm}, clip,width=\plotw]{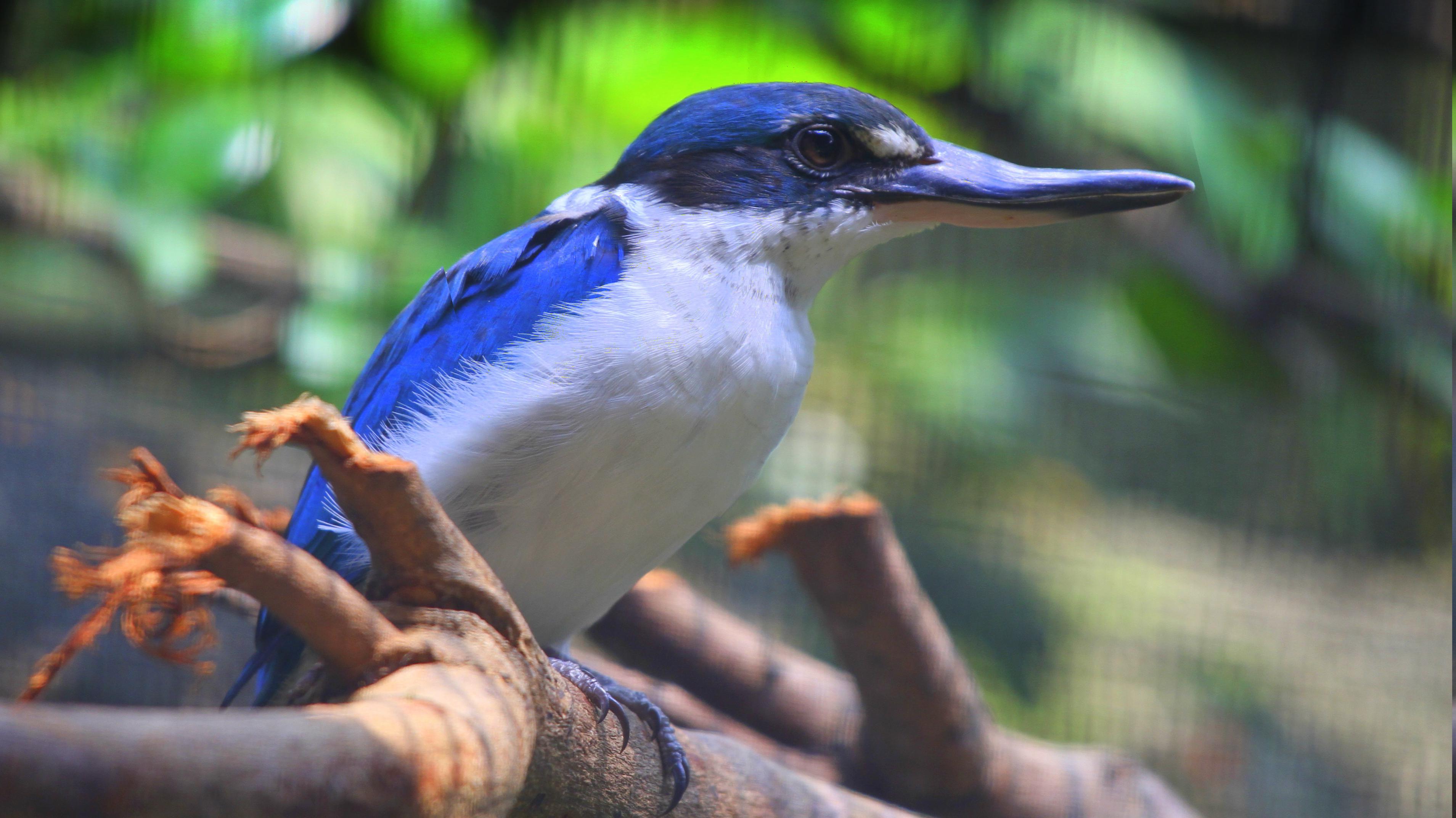} &
			\includegraphics[trim={500px 0cm 500px 0cm}, clip,width=\plotw]{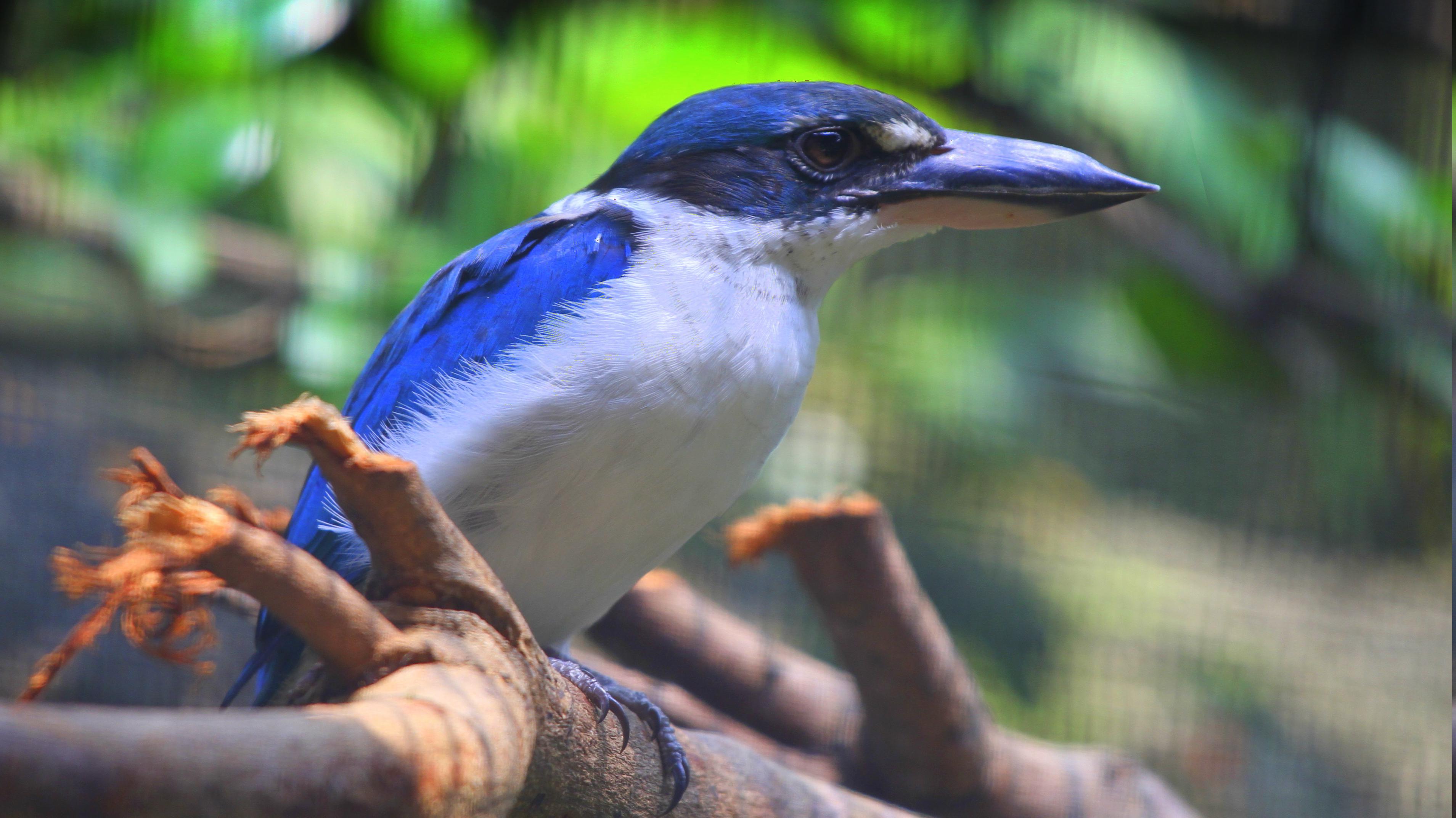} \\
			\scalebox{-1}[1]{\includegraphics[width=\plotw]{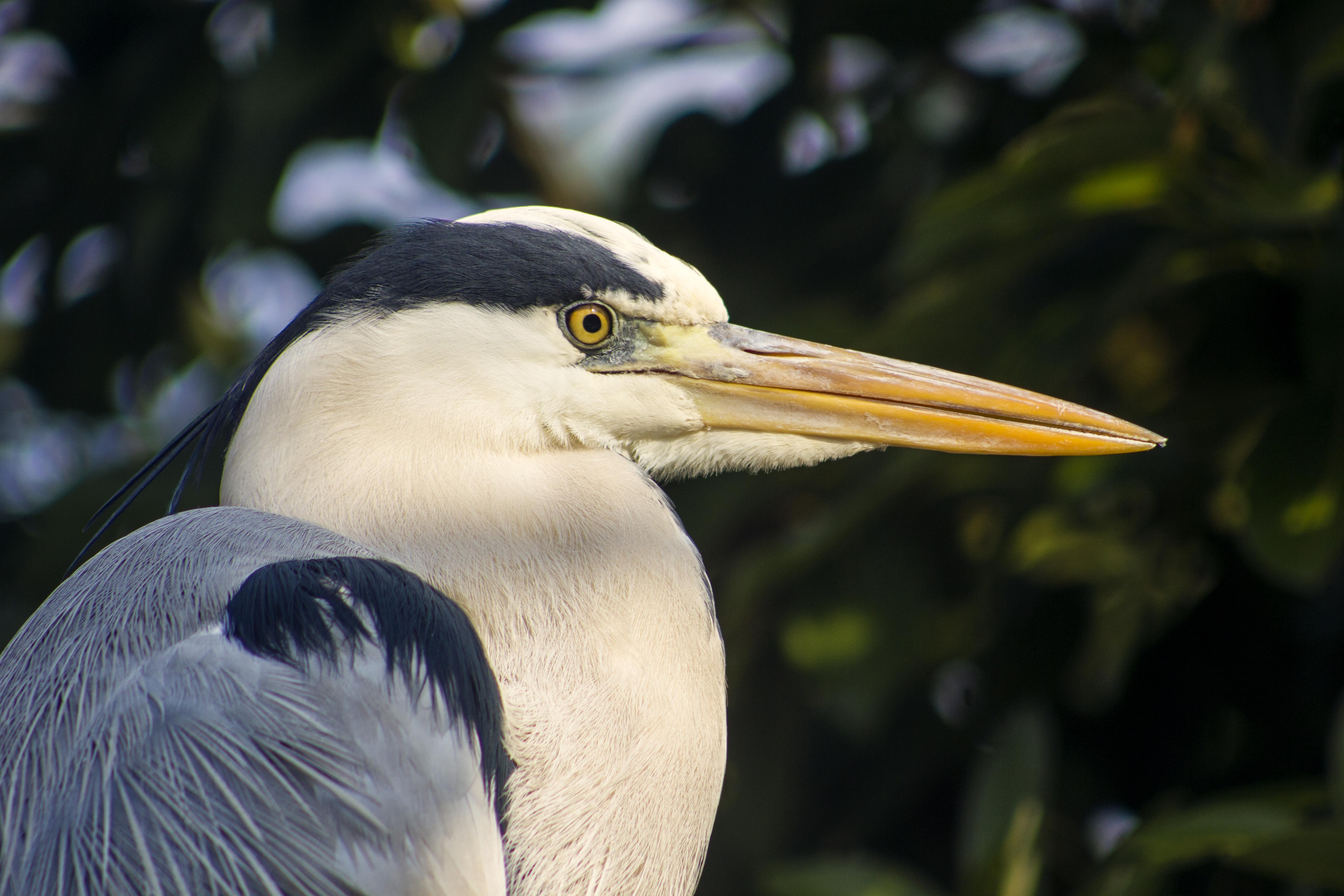}} &
			\scalebox{-1}[1]{\includegraphics[width=\plotw]{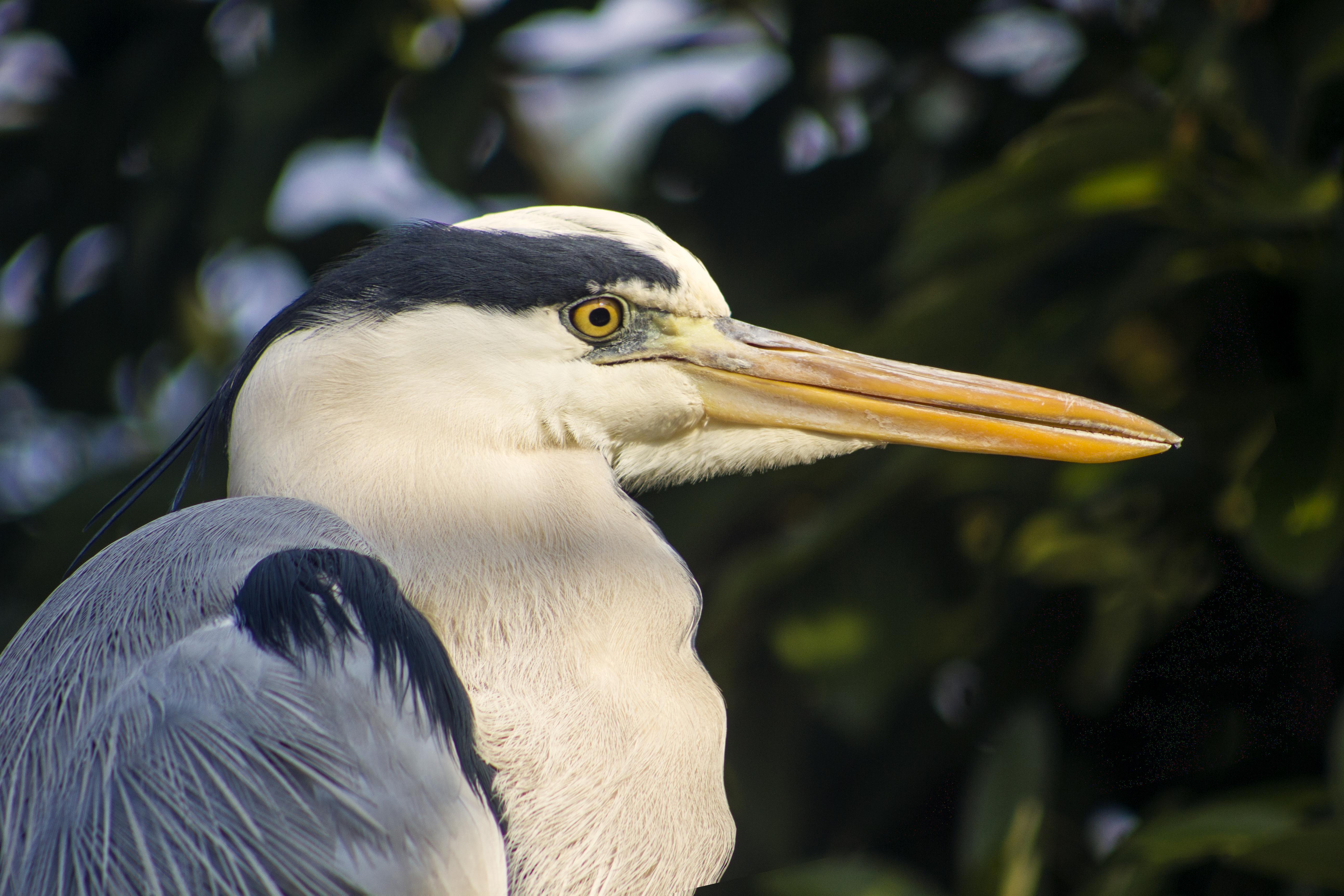}} &
			\scalebox{-1}[1]{\includegraphics[width=\plotw]{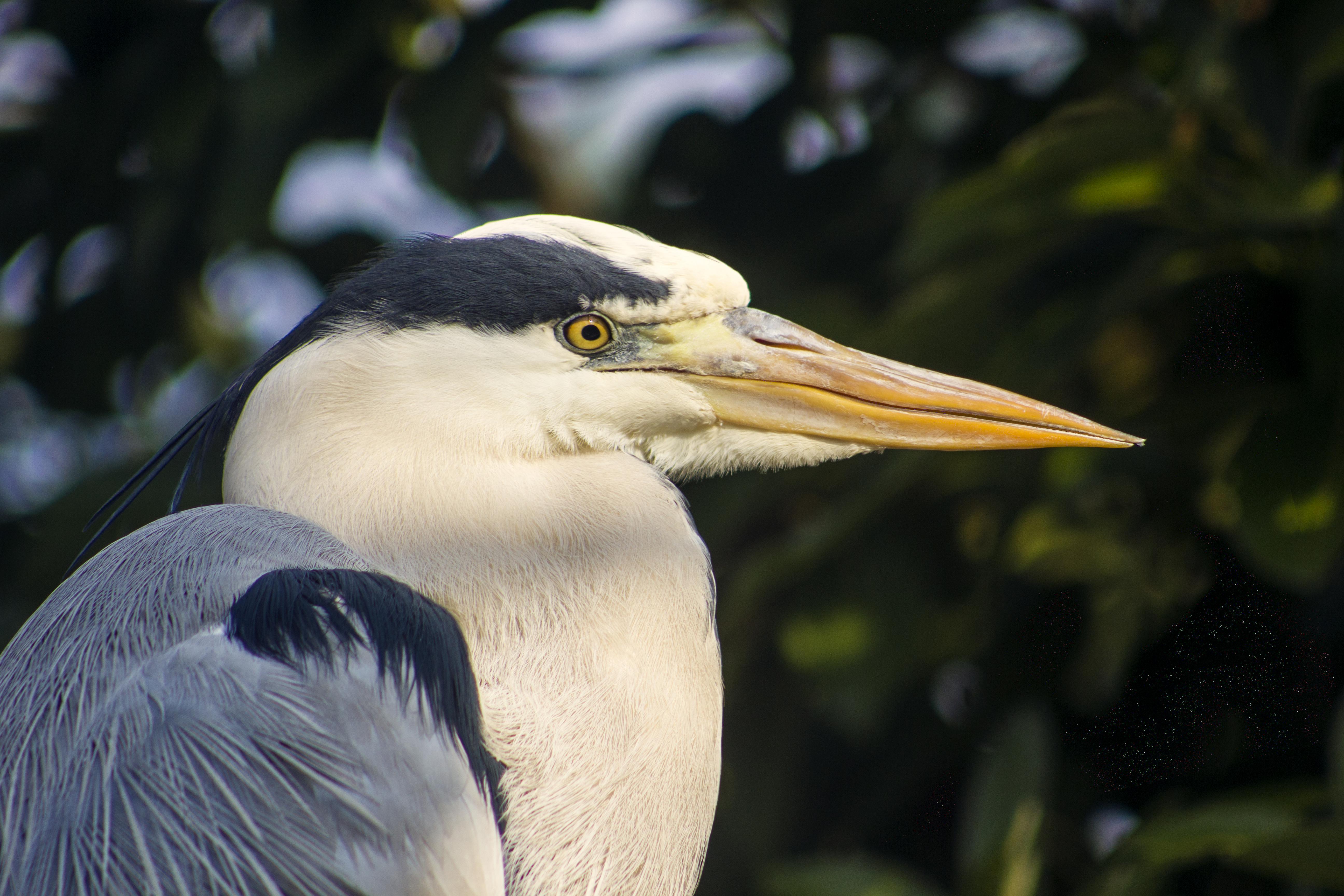}}
		\end{tabular}
		\caption{Additional results of our model on high-resolution images.
			Our model predicts warps at low resolution that can then be resized and applied to high resolution images.
			The model is able to keep the content and identity at high resolution.
			Please see supplemental videos demonstrating animated edits.
			Input images courtesy of Flickr users mickey, Lisa Leonardelli, and Andrey Grushnikov.}
\end{figure}

\FloatBarrier
\newpage
\section{Qualitative results on CelebA}
\def\plotw{0.08\linewidth}
\def\imagea{0}
\def\imageb{1}
\def\imagec{2}
\def\imaged{3}
\def\imagee{4}
\def\imagef{5}
\def\imageg{10}
\def\imageh{11}

\begin{figure}[H]
	\centering
	\setlength{\tabcolsep}{0pt} 
	\tiny{
	\begin{tabular}{ccccccccccccc}
	& \small{Input} & \small{\makecell{ Smile }} & \small{\makecell{Big \\ nose}} & \small{\makecell{Arched \\ eyebrows}} & \small{\makecell{ Narrowed \\ eyes}}  & \small{\makecell{ Pointy \\ nose}} &
	\small{Input} & \small{\makecell{No \\ smile}} & \small{\makecell{No big \\ nose}} & \small{\makecell{Arched \\ eyebrows}} & \small{\makecell{ Narrowed \\ eyes}}  & \small{\makecell{ Pointy \\ nose}}\\[9pt]
	&  & $0.76$ / $\mathbf{1.00}$ & $0.62$ / $\mathbf{1.00}$ & $0.84$ / $0.92$ & $0.86$ / $0.35$ & $0.75$ / $0.01$ & & $0.78$ / $\mathbf{1.00}$ & $0.77$ / $0.99$ & $0.71$ / $\mathbf{1.00}$ & $0.76$ / $0.89$ & $0.50$ / $0.98$ \\
	 \parbox[t]{4mm}{\rotatebox[origin=c]{90}{\scriptsize{StarGAN}}} & 
	\includegraphics[align=c,width=\plotw]{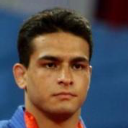} &
	\includegraphics[align=c,width=\plotw]{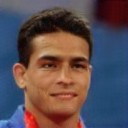} &
	\includegraphics[align=c,width=\plotw]{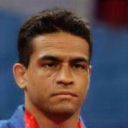} &
	\includegraphics[align=c,width=\plotw]{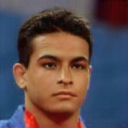} &
	\includegraphics[align=c,width=\plotw]{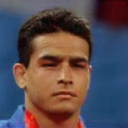} &
	\includegraphics[align=c,width=\plotw]{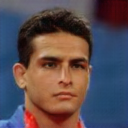}&
	~\includegraphics[align=c,width=\plotw]{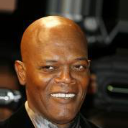} &
	\includegraphics[align=c,width=\plotw]{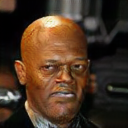} &
	\includegraphics[align=c,width=\plotw]{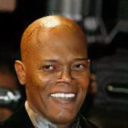} &
	\includegraphics[align=c,width=\plotw]{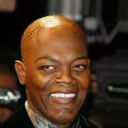} &
	\includegraphics[align=c,width=\plotw]{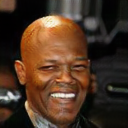} &
	\includegraphics[align=c,width=\plotw]{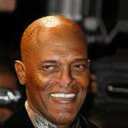} \\[24pt]
	& & $\mathbf{0.87}$ / $0.98$ & $0.69$ / $\mathbf{1.00}$ & $0.73$ / $\mathbf{1.00}$ & $0.87$ / $0.04$ & $0.73$ / $0.93$ & & $0.83$ / $\mathbf{1.00}$ & $0.85$ / $\mathbf{1.00}$ & $0.74$ / $\mathbf{1.00}$ & $0.87$ / $\mathbf{0.98}$ & $0.59$ / $\mathbf{1.00}$ \\
	\parbox[t]{4mm}{\rotatebox[origin=c]{90}{\scriptsize{StarGAN+}}}  & 
	\includegraphics[align=c,width=\plotw]{img/celeba_lr_warp/input/\imagea} &
	\includegraphics[align=c,width=\plotw]{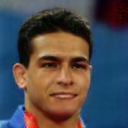} &
	\includegraphics[align=c,width=\plotw]{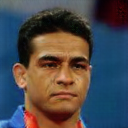} &
	\includegraphics[align=c,width=\plotw]{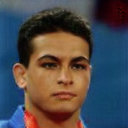} &
	\includegraphics[align=c,width=\plotw]{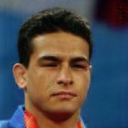} &
	\includegraphics[align=c,width=\plotw]{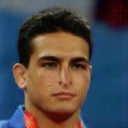} &
	~\includegraphics[align=c,width=\plotw]{img/celeba_lr_warp/input/\imageb} &
	\includegraphics[align=c,width=\plotw]{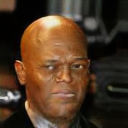} &
	\includegraphics[align=c,width=\plotw]{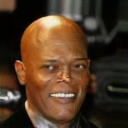} &
	\includegraphics[align=c,width=\plotw]{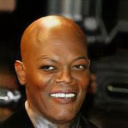} &
	\includegraphics[align=c,width=\plotw]{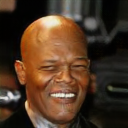} &
	\includegraphics[align=c,width=\plotw]{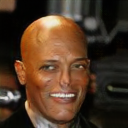} \\[24pt]
	& &  $0.80$ / $0.00$ & $\mathbf{0.78}$ / $\mathbf{1.00}$ & $\mathbf{0.86}$ / $\mathbf{1.00}$ & $\mathbf{0.91}$ / $\mathbf{0.46}$ & $\mathbf{0.83}$ / $\mathbf{0.97}$ & & $\mathbf{0.90}$ / $\mathbf{1.00}$ & $\mathbf{0.90}$ / $0.03$ & $\mathbf{0.94}$ / $0.97$ & $\mathbf{0.97}$ / $0.81$ & $\mathbf{0.86}$ / $\mathbf{1.00}$ \\
	\parbox[t]{4mm}{\rotatebox[origin=c]{90}{\scriptsize{\textbf{WarpGAN+}}}}  & 
	\includegraphics[align=c,width=\plotw]{img/celeba_lr_warp/input/\imagea} &
	\includegraphics[align=c,width=\plotw]{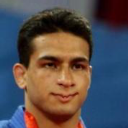} &
	\includegraphics[align=c,width=\plotw]{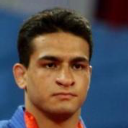} &
	\includegraphics[align=c,width=\plotw]{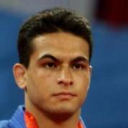} &
	\includegraphics[align=c,width=\plotw]{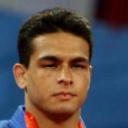} &
	\includegraphics[align=c,width=\plotw]{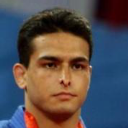} &
	~\includegraphics[align=c,width=\plotw]{img/celeba_lr_warp/input/\imageb} &
	\includegraphics[align=c,width=\plotw]{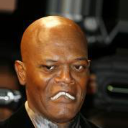} &
	\includegraphics[align=c,width=\plotw]{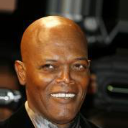} &
	\includegraphics[align=c,width=\plotw]{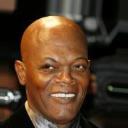} &
	\includegraphics[align=c,width=\plotw]{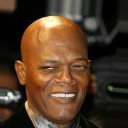} &
	\includegraphics[align=c,width=\plotw]{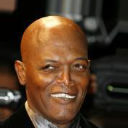} \\[20pt]\midrule
	& \small{Input} & \small{\makecell{ No \\ smile}} & \small{\makecell{Big \\ nose}} & \small{\makecell{No arched \\ eyebrows}} & \small{\makecell{ Narrowed \\ eyes}}  & \small{\makecell{ No pointy \\ nose}} &
	\small{Input} & \small{Smile} & \small{\makecell{Big \\ nose}} & \small{\makecell{Arched \\ eyebrows}} & \small{\makecell{ Narrowed \\ eyes}}  & \small{\makecell{ Pointy \\ nose}}\\[9pt]
	& & $0.47$ / $\mathbf{1.00}$ & $0.60$ / $0.94$ & $0.82$ / $\mathbf{1.00}$ & $0.77$ / $\mathbf{1.00}$ & $0.83$ / $\mathbf{1.00}$ & &  $0.64$ / $\mathbf{1.00}$ & $0.51$ / $\mathbf{0.91}$ & $0.60$ / $0.99$ & $0.63$ / $0.82$ & $0.54$ / $0.39$ \\
	 \parbox[t]{4mm}{\rotatebox[origin=c]{90}{\scriptsize{StarGAN}}} & 
	\includegraphics[align=c,width=\plotw]{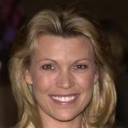} &
	\includegraphics[align=c,width=\plotw]{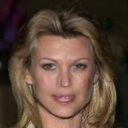} &
	\includegraphics[align=c,width=\plotw]{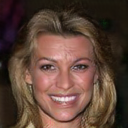} &
	\includegraphics[align=c,width=\plotw]{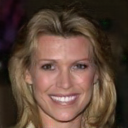} &
	\includegraphics[align=c,width=\plotw]{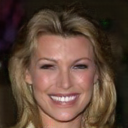} &
	\includegraphics[align=c,width=\plotw]{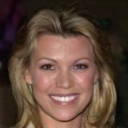}&
	~\includegraphics[align=c,width=\plotw]{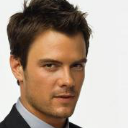} &
	\includegraphics[align=c,width=\plotw]{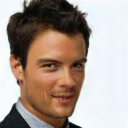} &
	\includegraphics[align=c,width=\plotw]{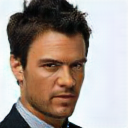} &
	\includegraphics[align=c,width=\plotw]{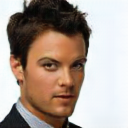} &
	\includegraphics[align=c,width=\plotw]{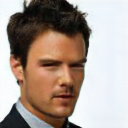} &
	\includegraphics[align=c,width=\plotw]{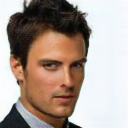} \\[24pt]
	& &  $0.60$ / $\mathbf{1.00}$ & $0.65$ / $\mathbf{1.00}$ & $0.74$ / $\mathbf{1.00}$ & $0.75$ / $\mathbf{1.00}$ & $0.76$ / $\mathbf{1.00}$ & & $0.73$ / $0.99$ & $\mathbf{0.81}$ / $0.26$ & $0.62$ / $\mathbf{1.00}$ & $0.55$ / $0.97$ & $0.62$ / $\mathbf{0.66}$ \\
	\parbox[t]{4mm}{\rotatebox[origin=c]{90}{\scriptsize{StarGAN+}}}  & 
	\includegraphics[align=c,width=\plotw]{img/celeba_lr_warp/input/\imagec} &
	\includegraphics[align=c,width=\plotw]{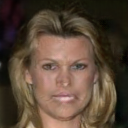} &
	\includegraphics[align=c,width=\plotw]{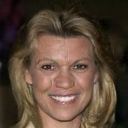} &
	\includegraphics[align=c,width=\plotw]{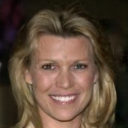} &
	\includegraphics[align=c,width=\plotw]{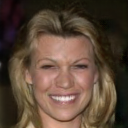} &
	\includegraphics[align=c,width=\plotw]{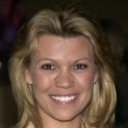} &
	~\includegraphics[align=c,width=\plotw]{img/celeba_lr_warp/input/\imaged} &
	\includegraphics[align=c,width=\plotw]{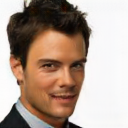} &
	\includegraphics[align=c,width=\plotw]{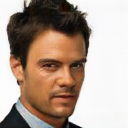} &
	\includegraphics[align=c,width=\plotw]{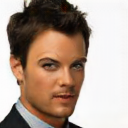} &
	\includegraphics[align=c,width=\plotw]{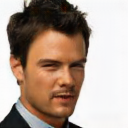} &
	\includegraphics[align=c,width=\plotw]{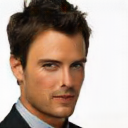} \\[24pt]
	& &  $\mathbf{0.81}$ / $\mathbf{1.00}$ & $\mathbf{0.83}$ / $\mathbf{1.00}$ & $\mathbf{0.84}$ / $\mathbf{1.00}$ & $\mathbf{0.85}$ / $0.99$ & $\mathbf{0.93}$ / $\mathbf{1.00}$ & & $\mathbf{0.77}$ / $0.76$ & $0.69$ / $0.31$ & $\mathbf{0.72}$ / $0.99$ & $\mathbf{0.83}$ / $\mathbf{0.99}$ & $\mathbf{0.66}$ / $0.02$ \\
	\parbox[t]{4mm}{\rotatebox[origin=c]{90}{\scriptsize{\textbf{WarpGAN+}}}}  & 
	\includegraphics[align=c,width=\plotw]{img/celeba_lr_warp/input/\imagec} &
	\includegraphics[align=c,width=\plotw]{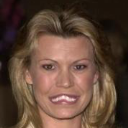} &
	\includegraphics[align=c,width=\plotw]{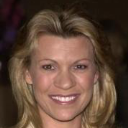} &
	\includegraphics[align=c,width=\plotw]{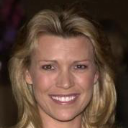} &
	\includegraphics[align=c,width=\plotw]{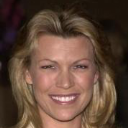} &
	\includegraphics[align=c,width=\plotw]{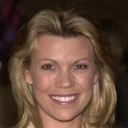} &
	~\includegraphics[align=c,width=\plotw]{img/celeba_lr_warp/input/\imaged} &
	\includegraphics[align=c,width=\plotw]{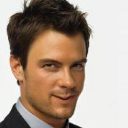} &
	\includegraphics[align=c,width=\plotw]{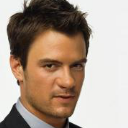} &
	\includegraphics[align=c,width=\plotw]{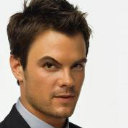} &
	\includegraphics[align=c,width=\plotw]{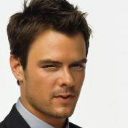} &
	\includegraphics[align=c,width=\plotw]{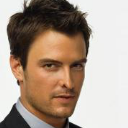} \\[20pt]\midrule
	& \small{Input} & \small{\makecell{ No \\ smile}} & \small{\makecell{Big \\ nose}} & \small{\makecell{No arched \\ eyebrows}} & \small{\makecell{ Narrowed \\ eyes}}  & \small{\makecell{ Pointy \\ nose}} &
	\small{Input} & \small{Smile} & \small{\makecell{Big \\ nose}} & \small{\makecell{Arched \\ eyebrows}} & \small{\makecell{ Narrowed \\ eyes}}  & \small{\makecell{ Pointy \\ nose}}\\[9pt]
	&  &  $0.59$ / $\mathbf{1.00}$ & $0.57$ / $0.99$ & $0.81$ / $\mathbf{1.00}$ & $0.50$ / $\mathbf{1.00}$ & $0.71$ / $\mathbf{1.00}$ & & $0.59$ / $\mathbf{1.00}$ & $0.30$ / $\mathbf{1.00}$ & $0.32$ / $\mathbf{0.99}$ & $0.66$ / $\mathbf{1.00}$ & $0.58$ / $0.98$ \\
	 \parbox[t]{4mm}{\rotatebox[origin=c]{90}{\scriptsize{StarGAN}}} & 
	\includegraphics[align=c,width=\plotw]{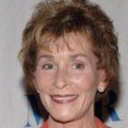} &
	\includegraphics[align=c,width=\plotw]{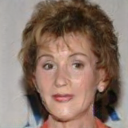} &
	\includegraphics[align=c,width=\plotw]{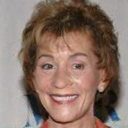} &
	\includegraphics[align=c,width=\plotw]{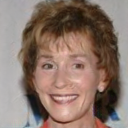} &
	\includegraphics[align=c,width=\plotw]{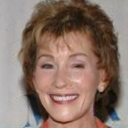} &
	\includegraphics[align=c,width=\plotw]{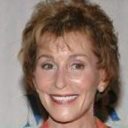}&
	~\includegraphics[align=c,width=\plotw]{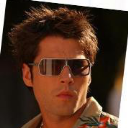} &
	\includegraphics[align=c,width=\plotw]{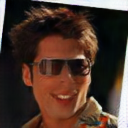} &
	\includegraphics[align=c,width=\plotw]{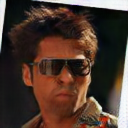} &
	\includegraphics[align=c,width=\plotw]{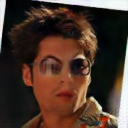} &
	\includegraphics[align=c,width=\plotw]{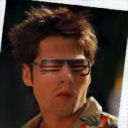} &
	\includegraphics[align=c,width=\plotw]{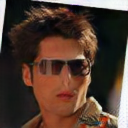} \\[24pt]
	& & $\mathbf{0.81}$ / $\mathbf{1.00}$ & $0.56$ / $\mathbf{1.00}$ & $0.73$ / $\mathbf{1.00}$ & $0.65$ / $\mathbf{1.00}$ & $0.75$ / $\mathbf{1.00}$ & & $0.66$ / $\mathbf{1.00}$ & $0.18$ / $\mathbf{1.00}$ & $0.12$ / $0.54$ & $0.42$ / $\mathbf{1.00}$ & $0.48$ / $\mathbf{1.00}$ \\
	\parbox[t]{4mm}{\rotatebox[origin=c]{90}{\scriptsize{StarGAN+}}}  & 
	\includegraphics[align=c,width=\plotw]{img/celeba_lr_warp/input/\imagee} &
	\includegraphics[align=c,width=\plotw]{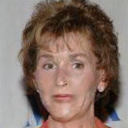} &
	\includegraphics[align=c,width=\plotw]{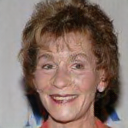} &
	\includegraphics[align=c,width=\plotw]{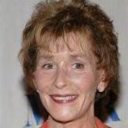} &
	\includegraphics[align=c,width=\plotw]{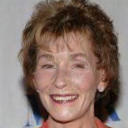} &
	\includegraphics[align=c,width=\plotw]{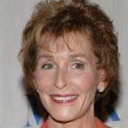} &
	~\includegraphics[align=c,width=\plotw]{img/celeba_lr_warp/input/\imagef} &
	\includegraphics[align=c,width=\plotw]{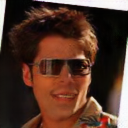} &
	\includegraphics[align=c,width=\plotw]{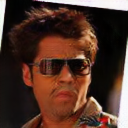} &
	\includegraphics[align=c,width=\plotw]{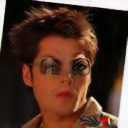} &
	\includegraphics[align=c,width=\plotw]{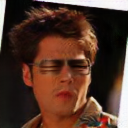} &
	\includegraphics[align=c,width=\plotw]{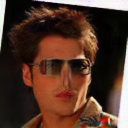} \\[24pt]
	& & $0.79$ / $0.99$ & $\mathbf{0.74}$ / $\mathbf{1.00}$ & $\mathbf{0.89}$ / $\mathbf{1.00}$ & $\mathbf{0.86}$ / $\mathbf{1.00}$ & $\mathbf{0.92}$ / $\mathbf{1.00}$ & & $\mathbf{0.77}$ / $0.01$ & $\mathbf{0.57}$ / $\mathbf{1.00}$ & $\mathbf{0.80}$ / $0.09$ & $\mathbf{0.79}$ / $0.03$ & $\mathbf{0.78}$ / $\mathbf{1.00}$ \\
	\parbox[t]{4mm}{\rotatebox[origin=c]{90}{\scriptsize{\textbf{WarpGAN+}}}}  & 
	\includegraphics[align=c,width=\plotw]{img/celeba_lr_warp/input/\imagee} &
	\includegraphics[align=c,width=\plotw]{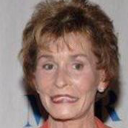} &
	\includegraphics[align=c,width=\plotw]{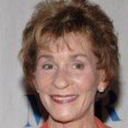} &
	\includegraphics[align=c,width=\plotw]{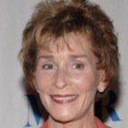} &
	\includegraphics[align=c,width=\plotw]{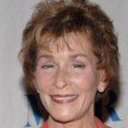} &
	\includegraphics[align=c,width=\plotw]{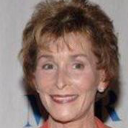} &
	~\includegraphics[align=c,width=\plotw]{img/celeba_lr_warp/input/\imagef} &
	\includegraphics[align=c,width=\plotw]{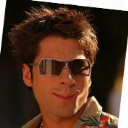} &
	\includegraphics[align=c,width=\plotw]{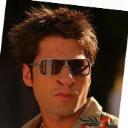} &
	\includegraphics[align=c,width=\plotw]{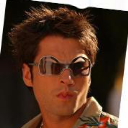} &
	\includegraphics[align=c,width=\plotw]{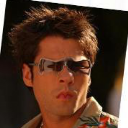} &
	\includegraphics[align=c,width=\plotw]{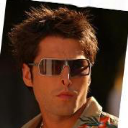} \\[20pt]
	\end{tabular}
	}
	\caption{Comparison to previous work on the CelebA dataset.
	From a given input image, first column, each method attempts to transfer the semantic attribute in its corresponding column.
	On top of each image the re-identification score  and the classification accuracy are shown as (id $/$ cls) (higher is better). (Zoom in for details)}
	\label{fig:sup_celeba_comparison}
\end{figure}

\FloatBarrier
\newpage
\section{Qualitative results on Cub200}
\def\plotw{0.16\linewidth}
\def\imga{0}
\def\imgaa{16}
\def\imgb{24}
\def\imgbb{21}
\def\imgc{242}
\def\imgd{221}
\def\imge{190}
\def\imgf{67}

\begin{figure}[H]
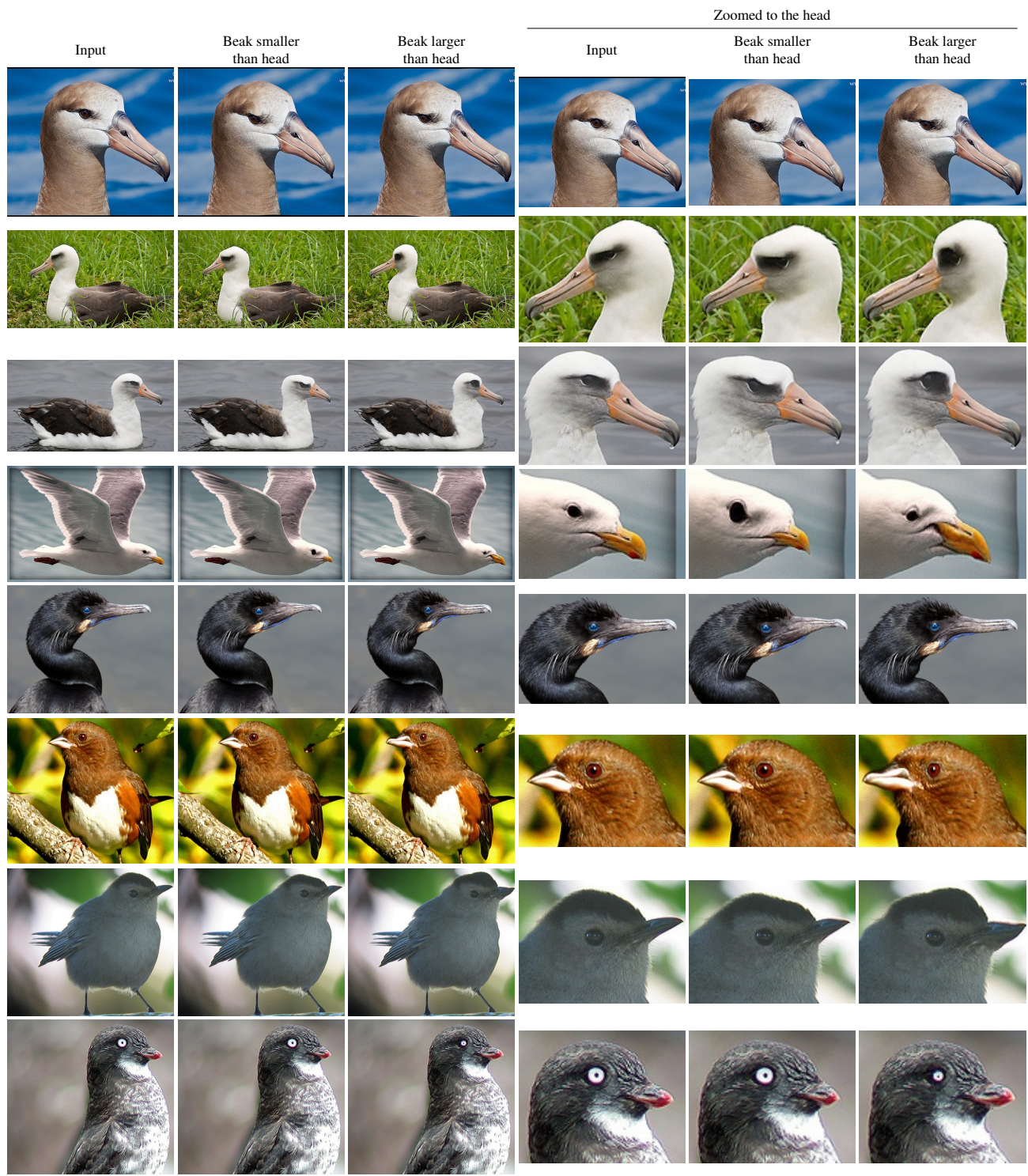

	\centering
	\setlength{\tabcolsep}{1pt} 
	\scriptsize{
		\begin{tabular}{cccccc} 
			& & & \multicolumn{3}{c}{Zoomed to the head} \\ \cmidrule(lr){4-6}
			Input & \makecell{Beak smaller \\ than head} & \makecell{Beak larger \\ than head} & Input & \makecell{Beak smaller \\ than head} & \makecell{Beak larger \\ than head} \\
			\includegraphics[align=c,width=\plotw]{../rebuttal/img/cub_200/input_hr/\imga} &
			\includegraphics[align=c,width=\plotw]{../rebuttal/img/cub_200/warpgan/\imga/generated_hr/2} &
			\includegraphics[align=c,width=\plotw]{../rebuttal/img/cub_200/warpgan/\imga/generated_hr/1} &
			\includegraphics[align=c,trim={0cm 1.5cm 0cm 0cm},clip,width=\plotw]{../rebuttal/img/cub_200/input_hr/\imga} &
			\includegraphics[align=c,trim={1.2cm 2cm 0cm 0.7cm},clip,width=\plotw]{../rebuttal/img/cub_200/warpgan/\imga/generated_hr/2} &
			\includegraphics[align=c,trim={1.2cm 2cm 0cm 0.7cm},clip,width=\plotw]{../rebuttal/img/cub_200/warpgan/\imga/generated_hr/1} \\[25pt]
			\scalebox{-1}[1]{\includegraphics[align=c,width=\plotw]{../rebuttal/img/cub_200/input_hr/\imgaa}} &
			\scalebox{-1}[1]{\includegraphics[align=c,width=\plotw]{../rebuttal/img/cub_200/warpgan/\imgaa/generated_hr/2}} &
			\scalebox{-1}[1]{\includegraphics[align=c,width=\plotw]{../rebuttal/img/cub_200/warpgan/\imgaa/generated_hr/1}} &
			\scalebox{-1}[1]{\includegraphics[align=c,trim={8.5cm 4cm 1.9cm 1.2cm},clip,width=\plotw]{../rebuttal/img/cub_200/input_hr/\imgaa}} &
			\scalebox{-1}[1]{\includegraphics[align=c,trim={8.5cm 4cm 1.9cm 1.2cm},clip,width=\plotw]{../rebuttal/img/cub_200/warpgan/\imgaa/generated_hr/2}} &
			\scalebox{-1}[1]{\includegraphics[align=c,trim={8.5cm 4cm 1.9cm 1.2cm},clip,width=\plotw]{../rebuttal/img/cub_200/warpgan/\imgaa/generated_hr/1}} \\[28pt]
			\includegraphics[align=c,width=\plotw]{../rebuttal/img/cub_200/input_hr/\imgb} &
			\includegraphics[align=c,width=\plotw]{../rebuttal/img/cub_200/warpgan/\imgb/generated_hr/2} &
			\includegraphics[align=c,width=\plotw]{../rebuttal/img/cub_200/warpgan/\imgb/generated_hr/1} &
			\includegraphics[align=c,trim={10.8cm 4.31cm 1cm 1.25cm},clip,width=\plotw]{../rebuttal/img/cub_200/input_hr/\imgb} &
			\includegraphics[align=c,trim={10.8cm 4.31cm 1cm 1.25cm},clip,width=\plotw]{../rebuttal/img/cub_200/warpgan/\imgb/generated_hr/2} &
			\includegraphics[align=c,trim={10.8cm 4.31cm 1cm 1.25cm},clip,width=\plotw]{../rebuttal/img/cub_200/warpgan/\imgb/generated_hr/1} \\[25pt]
			\includegraphics[align=c,width=\plotw]{../rebuttal/img/cub_200/input_hr/\imgbb} &
			\includegraphics[align=c,width=\plotw]{../rebuttal/img/cub_200/warpgan/\imgbb/generated_hr/2} &
			\includegraphics[align=c,width=\plotw]{../rebuttal/img/cub_200/warpgan/\imgbb/generated_hr/1} &
			\includegraphics[align=c,trim={13cm 1.3cm 0cm 7.9cm},clip,width=\plotw]{../rebuttal/img/cub_200/input_hr/\imgbb} &
			\includegraphics[align=c,trim={13cm 1.3cm 0cm 7.9cm},clip,width=\plotw]{../rebuttal/img/cub_200/warpgan/\imgbb/generated_hr/2} &
			\includegraphics[align=c,trim={13cm 1.3cm 0cm 7.9cm},clip,width=\plotw]{../rebuttal/img/cub_200/warpgan/\imgbb/generated_hr/1} \\[25pt]
			\includegraphics[align=c,trim={0cm 0cm 0cm 1cm},clip,width=\plotw]{../rebuttal/img/cub_200/input_hr/\imgc} &
			\includegraphics[align=c,trim={0cm 0cm 0cm 1cm},clip,width=\plotw]{../rebuttal/img/cub_200/warpgan/\imgc/generated_hr/2} &
			\includegraphics[align=c,trim={0cm 0cm 0cm 1cm},clip,width=\plotw]{../rebuttal/img/cub_200/warpgan/\imgc/generated_hr/1} &
			\includegraphics[align=c,trim={2cm 4cm 1.5cm 1cm},clip,width=\plotw]{../rebuttal/img/cub_200/input_hr/\imgc} &
			\includegraphics[align=c,trim={2cm 4cm 1.5cm 1cm},clip,width=\plotw]{../rebuttal/img/cub_200/warpgan/\imgc/generated_hr/2} &
			\includegraphics[align=c,trim={2cm 4cm 1.5cm 1cm},clip,width=\plotw]{../rebuttal/img/cub_200/warpgan/\imgc/generated_hr/1} \\[29pt]
			\scalebox{-1}[1]{\includegraphics[align=c,trim={0cm 3cm 0cm 2cm},clip,width=\plotw]{../rebuttal/img/cub_200/input_hr/\imgd}} &
			\scalebox{-1}[1]{\includegraphics[align=c,trim={0cm 3cm 0cm 2cm},clip,width=\plotw]{../rebuttal/img/cub_200/warpgan/\imgd/generated_hr/2}} &
			\scalebox{-1}[1]{\includegraphics[align=c,trim={0cm 3cm 0cm 2cm},clip,width=\plotw]{../rebuttal/img/cub_200/warpgan/\imgd/generated_hr/1}} &
			\scalebox{-1}[1]{\includegraphics[align=c,trim={3cm 9cm 2.5cm 2cm},clip,width=\plotw]{../rebuttal/img/cub_200/input_hr/\imgd}} &
			\scalebox{-1}[1]{\includegraphics[align=c,trim={3cm 9cm 2.5cm 2cm},clip,width=\plotw]{../rebuttal/img/cub_200/warpgan/\imgd/generated_hr/2}} &
			\scalebox{-1}[1]{\includegraphics[align=c,trim={3cm 9cm 2.5cm 2cm},clip,width=\plotw]{../rebuttal/img/cub_200/warpgan/\imgd/generated_hr/1}} \\[33pt]
			\includegraphics[align=c,trim={0cm 0cm 0cm 1.5cm},clip,width=\plotw]{../rebuttal/img/cub_200/input_hr/\imge} &
			\includegraphics[align=c,trim={0cm 0cm 0cm 1.5cm},clip,width=\plotw]{../rebuttal/img/cub_200/warpgan/\imge/generated_hr/2} &
			\includegraphics[align=c,trim={0cm 0cm 0cm 1.5cm},clip,width=\plotw]{../rebuttal/img/cub_200/warpgan/\imge/generated_hr/1} &
			\includegraphics[align=c,trim={7cm 7cm 0cm 1.5cm},clip,width=\plotw]{../rebuttal/img/cub_200/input_hr/\imge} &
			\includegraphics[align=c,trim={7cm 7cm 0cm 1.5cm},clip,width=\plotw]{../rebuttal/img/cub_200/warpgan/\imge/generated_hr/2} &
			\includegraphics[align=c,trim={7cm 7cm 0cm 1.5cm},clip,width=\plotw]{../rebuttal/img/cub_200/warpgan/\imge/generated_hr/1} \\[33pt]
			\includegraphics[align=c,trim={0cm 5cm 0cm 1cm},clip,width=\plotw]{../rebuttal/img/cub_200/input_hr/\imgf} &
			\includegraphics[align=c,trim={0cm 5cm 0cm 1cm},clip,width=\plotw]{../rebuttal/img/cub_200/warpgan/\imgf/generated_hr/2} &
			\includegraphics[align=c,trim={0cm 5cm 0cm 1cm},clip,width=\plotw]{../rebuttal/img/cub_200/warpgan/\imgf/generated_hr/1} &
			\includegraphics[align=c,trim={3.5cm 9cm 0.25cm 1cm},clip,width=\plotw]{../rebuttal/img/cub_200/input_hr/\imgf} &
			\includegraphics[align=c,trim={3.5cm 9cm 0.25cm 1cm},clip,width=\plotw]{../rebuttal/img/cub_200/warpgan/\imgf/generated_hr/2} &
			\includegraphics[align=c,trim={3.5cm 9cm 0.25cm 1cm},clip,width=\plotw]{../rebuttal/img/cub_200/warpgan/\imgf/generated_hr/1}
		\end{tabular}
	}
	\caption{
	Additional results from our model on test images from the Cub200 dataset.
	The model attempts to transfer the attribute (relative beak size) in each column to the input image.
	For easiness of comparison, a crop of the head area is shown in the last three columns.}
	\label{fig:appendix:cub200}
\end{figure}

\FloatBarrier
\newpage
\section{Partial edits on CelebA}
\def\plotw{0.1\linewidth}

\def\imagea{3}
\def\attra{0}

\def\imageb{15}
\def\attrb{0_not}

\def\imagec{2}
\def\attrc{1}

\def\imaged{1}
\def\attrd{1_not}

\def\imagee{5}
\def\attre{2}

\def\imagef{4}
\def\attrf{2_not}

\def\imageg{7}
\def\attrg{3}

\def\imageh{6}
\def\attrh{3_not}

\def\imagei{11}
\def\attri{4}

\def\imagej{9}
\def\attrj{4_not}

\begin{figure}[H]
	\centering
	\setlength{\tabcolsep}{2pt} 
	\setlength{\fboxsep}{0pt} 
	\setlength{\fboxrule}{1.5pt} 
	\small{	
	\begin{tabular}{cccccccc}
	\textbf{Attribute} & $\alpha = -0.25$ & $\alpha = 0.0$ & $\alpha = 0.25$ & $\alpha = 0.50$ & $\alpha = 0.75$ & $\alpha = 1.00$ & $\alpha = 1.25$ \\[1pt]
	\makecell{ Smile } &
	\includegraphics[align=c,width=\plotw]{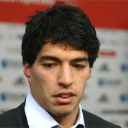} &
	\fcolorbox{red}{white}{\includegraphics[align=c,width=\plotw]{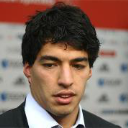}} &
	\includegraphics[align=c,width=\plotw]{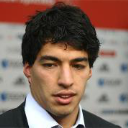} &
	\includegraphics[align=c,width=\plotw]{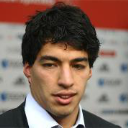} &
	\includegraphics[align=c,width=\plotw]{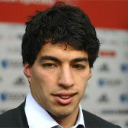} &
	\fcolorbox{green}{white}{\includegraphics[align=c,width=\plotw]{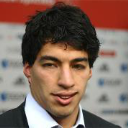}} &
	\includegraphics[align=c,width=\plotw]{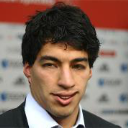} \\[22pt]
	\makecell{ Not \\ Smile } &
	\includegraphics[align=c,width=\plotw]{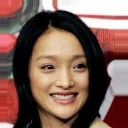} &
	\fcolorbox{red}{white}{\includegraphics[align=c,width=\plotw]{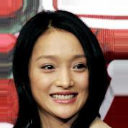}} &
	\includegraphics[align=c,width=\plotw]{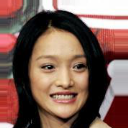} &
	\includegraphics[align=c,width=\plotw]{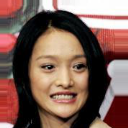} &
	\includegraphics[align=c,width=\plotw]{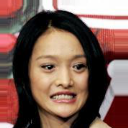} &
	\fcolorbox{green}{white}{\includegraphics[align=c,width=\plotw]{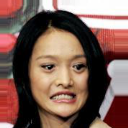}} &
	\includegraphics[align=c,width=\plotw]{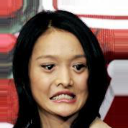} \\[22pt]
	\makecell{ Big \\ nose } &
	\includegraphics[align=c,width=\plotw]{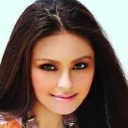} &
	\fcolorbox{red}{white}{\includegraphics[align=c,width=\plotw]{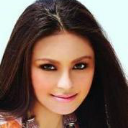}} &
	\includegraphics[align=c,width=\plotw]{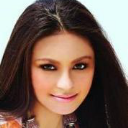} &
	\includegraphics[align=c,width=\plotw]{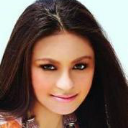} &
	\includegraphics[align=c,width=\plotw]{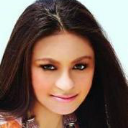} &
	\fcolorbox{green}{white}{\includegraphics[align=c,width=\plotw]{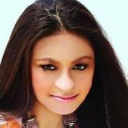}} &
	\includegraphics[align=c,width=\plotw]{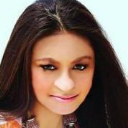} \\[22pt]
	\makecell{ Not big \\ nose } &
	\includegraphics[align=c,width=\plotw]{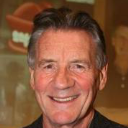} &
	\fcolorbox{red}{white}{\includegraphics[align=c,width=\plotw]{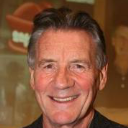}} &
	\includegraphics[align=c,width=\plotw]{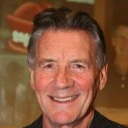} &
	\includegraphics[align=c,width=\plotw]{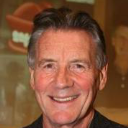} &
	\includegraphics[align=c,width=\plotw]{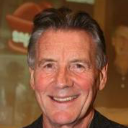} &
	\fcolorbox{green}{white}{\includegraphics[align=c,width=\plotw]{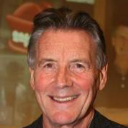}} &
	\includegraphics[align=c,width=\plotw]{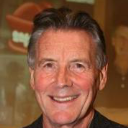} \\[22pt]
	\makecell{ Arched \\ eyebrows } &
	\includegraphics[align=c,width=\plotw]{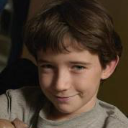} &
	\fcolorbox{red}{white}{\includegraphics[align=c,width=\plotw]{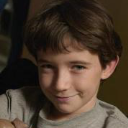}} &
	\includegraphics[align=c,width=\plotw]{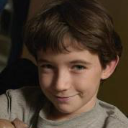} &
	\includegraphics[align=c,width=\plotw]{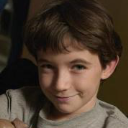} &
	\includegraphics[align=c,width=\plotw]{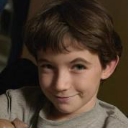} &
	\fcolorbox{green}{white}{\includegraphics[align=c,width=\plotw]{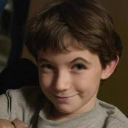}} &
	\includegraphics[align=c,width=\plotw]{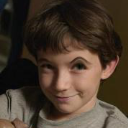} \\[22pt]
	\makecell{ No arched \\ eyebrows } &
	\includegraphics[align=c,width=\plotw]{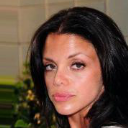} &
	\fcolorbox{red}{white}{\includegraphics[align=c,width=\plotw]{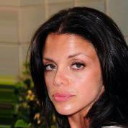}} &
	\includegraphics[align=c,width=\plotw]{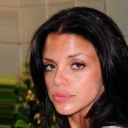} &
	\includegraphics[align=c,width=\plotw]{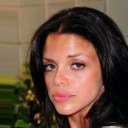} &
	\includegraphics[align=c,width=\plotw]{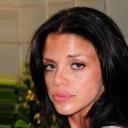} &
	\fcolorbox{green}{white}{\includegraphics[align=c,width=\plotw]{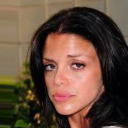}} &
	\includegraphics[align=c,width=\plotw]{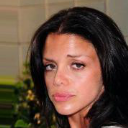} \\[22pt]
	\makecell{ Narrowed \\ eyes } &
	\includegraphics[align=c,width=\plotw]{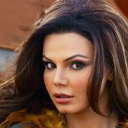} &
	\fcolorbox{red}{white}{\includegraphics[align=c,width=\plotw]{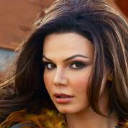}} &
	\includegraphics[align=c,width=\plotw]{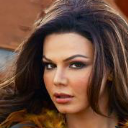} &
	\includegraphics[align=c,width=\plotw]{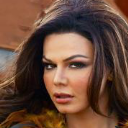} &
	\includegraphics[align=c,width=\plotw]{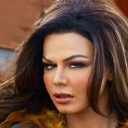} &
	\fcolorbox{green}{white}{\includegraphics[align=c,width=\plotw]{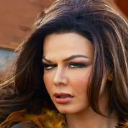}} &
	\includegraphics[align=c,width=\plotw]{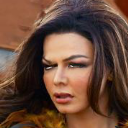} \\[22pt]
	\makecell{ No narrowed \\ eyes } &
	\includegraphics[align=c,width=\plotw]{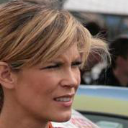} &
	\fcolorbox{red}{white}{\includegraphics[align=c,width=\plotw]{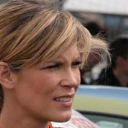}} &
	\includegraphics[align=c,width=\plotw]{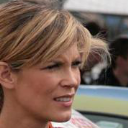} &
	\includegraphics[align=c,width=\plotw]{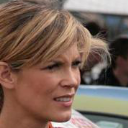} &
	\includegraphics[align=c,width=\plotw]{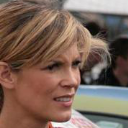} &
	\fcolorbox{green}{white}{\includegraphics[align=c,width=\plotw]{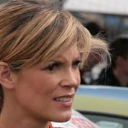}} &
	\includegraphics[align=c,width=\plotw]{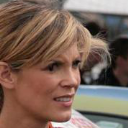} \\[22pt]
	\makecell{ Pointy \\ nose } &
	\includegraphics[align=c,width=\plotw]{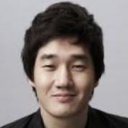} &
	\fcolorbox{red}{white}{\includegraphics[align=c,width=\plotw]{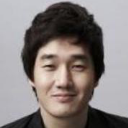}} &
	\includegraphics[align=c,width=\plotw]{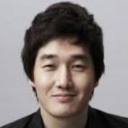} &
	\includegraphics[align=c,width=\plotw]{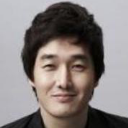} &
	\includegraphics[align=c,width=\plotw]{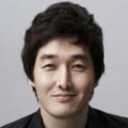} &
	\fcolorbox{green}{white}{\includegraphics[align=c,width=\plotw]{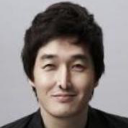}} &
	\includegraphics[align=c,width=\plotw]{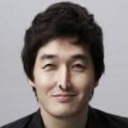} \\[22pt]
	\makecell{ No pointy \\ nose } &
	\includegraphics[align=c,width=\plotw]{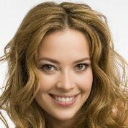} &
	\fcolorbox{red}{white}{\includegraphics[align=c,width=\plotw]{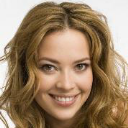}} &
	\includegraphics[align=c,width=\plotw]{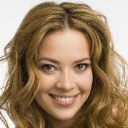} &
	\includegraphics[align=c,width=\plotw]{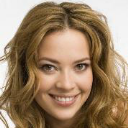} &
	\includegraphics[align=c,width=\plotw]{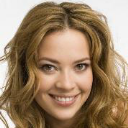} &
	\fcolorbox{green}{white}{\includegraphics[align=c,width=\plotw]{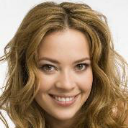}} &
	\includegraphics[align=c,width=\plotw]{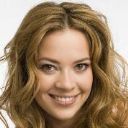}
	\end{tabular}
	}
	\caption{Partial editing with our model, for the the attribute indicated in the first column.
		A single warp is generated by our model, which is interpolated and extrapolated by scaling the magnitude of its values by $\alpha$.
		The input image, $\alpha \hspace{-1.5pt} = \hspace{-1.5pt} 0$, is progressively edited in both directions.
		A red box denotes the input image, and a green one the output of the generator without $\alpha$ scaling.
		Please see supplemental videos demonstrating animated edits.}
	\label{fig:supp_celeba_partial}
\end{figure}

\FloatBarrier
\newpage
\section{Stretch maps on CelebA}

\def\ploth{1.32cm}
\def\imagea{0}
\def\imageb{1}
\def\imagec{2}
\def\imaged{3}
\def\imagee{4}
\def\imagef{5}

\begin{figure}[H]
	\centering
	\setlength{\tabcolsep}{1pt} 
	\small{
	\begin{tabular}{ccccccccccccccc}
	& & \small{Input} & \small{\makecell{No \\ smile }} & \small{\makecell{Big \\ nose}} & \small{\makecell{Arched \\ eyebrows}} & \small{\makecell{ Narrowed \\ eyes}}  & \small{\makecell{ Pointy \\ nose}} &
	\small{Input} & \small{\makecell{No \\ smile}} & \small{\makecell{Big \\ nose}} & \small{\makecell{Arched \\ eyebrows}} & \small{\makecell{ Narrowed \\ eyes}}  & \small{\makecell{ Pointy \\ nose}} & \\[9pt]
	\multirow{3}{*}{\parbox[t]{4mm}{\rotatebox[origin=c]{90}{\hspace{-2.2cm} WarpGAN}}} &
	\parbox[t]{3mm}{\rotatebox[origin=c]{90}{\scriptsize{Input}}} &
	\includegraphics[align=c,height=\ploth]{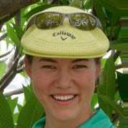} &
	\includegraphics[align=c,height=\ploth]{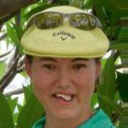} &
	\includegraphics[align=c,height=\ploth]{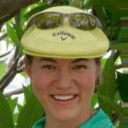} &
	\includegraphics[align=c,height=\ploth]{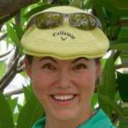} &
	\includegraphics[align=c,height=\ploth]{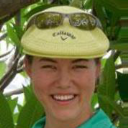} &
	\includegraphics[align=c,height=\ploth]{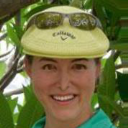} &
	\includegraphics[align=c,height=\ploth]{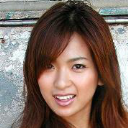} &
	\includegraphics[align=c,height=\ploth]{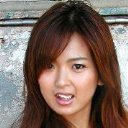} &
	\includegraphics[align=c,height=\ploth]{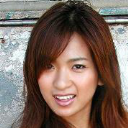} &
	\includegraphics[align=c,height=\ploth]{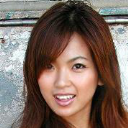} &
	\includegraphics[align=c,height=\ploth]{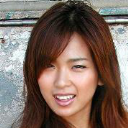} &
	\includegraphics[align=c,height=\ploth]{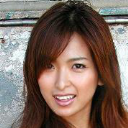} & 
	\\[16pt]
	& \parbox[t]{3mm}{\rotatebox[origin=c]{90}{\scriptsize{Stretch map}}} &
	\includegraphics[align=c,height=\ploth]{sup_imgs/celeba_stretch/binary/\imagea/input} &
	\includegraphics[align=c,height=\ploth]{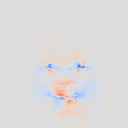} &
	\includegraphics[align=c,height=\ploth]{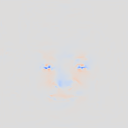} &
	\includegraphics[align=c,height=\ploth]{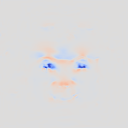} &
	\includegraphics[align=c,height=\ploth]{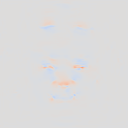} &
	\includegraphics[align=c,height=\ploth]{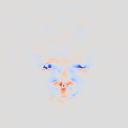} &
	\includegraphics[align=c,height=\ploth]{sup_imgs/celeba_stretch/binary/\imageb/input} &
	\includegraphics[align=c,height=\ploth]{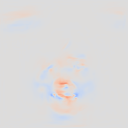} &
	\includegraphics[align=c,height=\ploth]{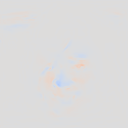} &
	\includegraphics[align=c,height=\ploth]{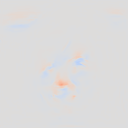} &
	\includegraphics[align=c,height=\ploth]{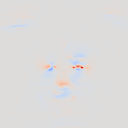} &
	\includegraphics[align=c,height=\ploth]{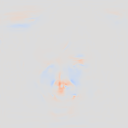} &
	\includegraphics[align=c,height=\ploth]{img/rafd_stretch/lr/colorbar} 
	\\[16pt]
	& \parbox[t]{3mm}{\rotatebox[origin=c]{90}{\scriptsize{Overlay}}} &
	\includegraphics[align=c,height=\ploth]{sup_imgs/celeba_stretch/binary/\imagea/input} &
	\includegraphics[align=c,height=\ploth]{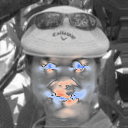} &
	\includegraphics[align=c,height=\ploth]{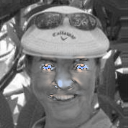} &
	\includegraphics[align=c,height=\ploth]{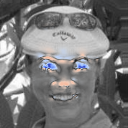} &
	\includegraphics[align=c,height=\ploth]{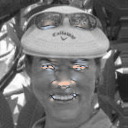} &
	\includegraphics[align=c,height=\ploth]{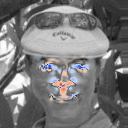} &
	\includegraphics[align=c,height=\ploth]{sup_imgs/celeba_stretch/binary/\imageb/input} &
	\includegraphics[align=c,height=\ploth]{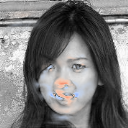} &
	\includegraphics[align=c,height=\ploth]{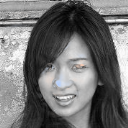} &
	\includegraphics[align=c,height=\ploth]{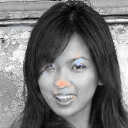} &
	\includegraphics[align=c,height=\ploth]{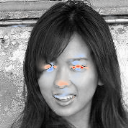} &
	\includegraphics[align=c,height=\ploth]{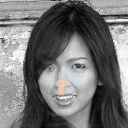} &
	\includegraphics[align=c,height=\ploth]{img/rafd_stretch/lr/colorbar} 
	\\[14pt]\midrule
	\multirow{3}{*}{\parbox[t]{4mm}{\rotatebox[origin=c]{90}{\hspace{-2.2cm} WarpGAN+}}} &
	\parbox[t]{3mm}{\rotatebox[origin=c]{90}{\scriptsize{Input}}} &
	\includegraphics[align=c,height=\ploth]{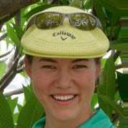} &
	\includegraphics[align=c,height=\ploth]{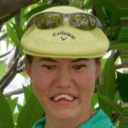} &
	\includegraphics[align=c,height=\ploth]{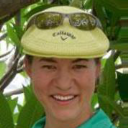} &
	\includegraphics[align=c,height=\ploth]{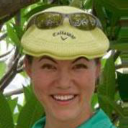} &
	\includegraphics[align=c,height=\ploth]{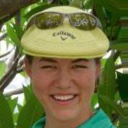} &
	\includegraphics[align=c,height=\ploth]{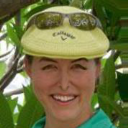} &
	\includegraphics[align=c,height=\ploth]{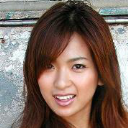} &
	\includegraphics[align=c,height=\ploth]{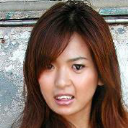} &
	\includegraphics[align=c,height=\ploth]{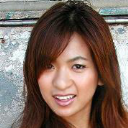} &
	\includegraphics[align=c,height=\ploth]{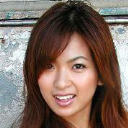} &
	\includegraphics[align=c,height=\ploth]{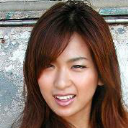} &
	\includegraphics[align=c,height=\ploth]{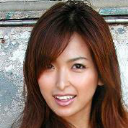} & 
	\\[16pt]
	& \parbox[t]{3mm}{\rotatebox[origin=c]{90}{\scriptsize{Stretch map}}} &
	\includegraphics[align=c,height=\ploth]{sup_imgs/celeba_stretch/ternary/\imagea/input} &
	\includegraphics[align=c,height=\ploth]{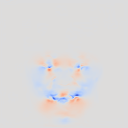} &
	\includegraphics[align=c,height=\ploth]{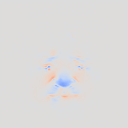} &
	\includegraphics[align=c,height=\ploth]{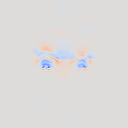} &
	\includegraphics[align=c,height=\ploth]{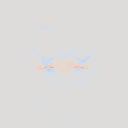} &
	\includegraphics[align=c,height=\ploth]{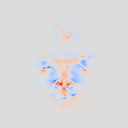} &
	\includegraphics[align=c,height=\ploth]{sup_imgs/celeba_stretch/ternary/\imageb/input} &
	\includegraphics[align=c,height=\ploth]{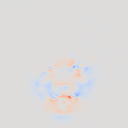} &
	\includegraphics[align=c,height=\ploth]{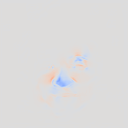} &
	\includegraphics[align=c,height=\ploth]{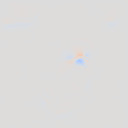} &
	\includegraphics[align=c,height=\ploth]{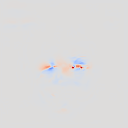} &
	\includegraphics[align=c,height=\ploth]{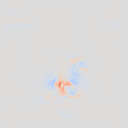} &
	\includegraphics[align=c,height=\ploth]{img/rafd_stretch/lr/colorbar} 
	\\[16pt]
	& \parbox[t]{3mm}{\rotatebox[origin=c]{90}{\scriptsize{Overlay}}} &
	\includegraphics[align=c,height=\ploth]{sup_imgs/celeba_stretch/ternary/\imagea/input} &
	\includegraphics[align=c,height=\ploth]{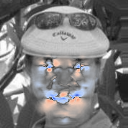} &
	\includegraphics[align=c,height=\ploth]{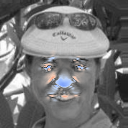} &
	\includegraphics[align=c,height=\ploth]{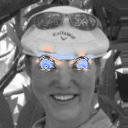} &
	\includegraphics[align=c,height=\ploth]{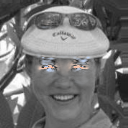} &
	\includegraphics[align=c,height=\ploth]{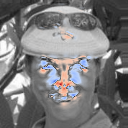} &
	\includegraphics[align=c,height=\ploth]{sup_imgs/celeba_stretch/ternary/\imageb/input} &
	\includegraphics[align=c,height=\ploth]{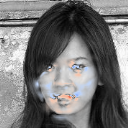} &
	\includegraphics[align=c,height=\ploth]{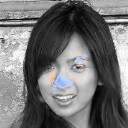} &
	\includegraphics[align=c,height=\ploth]{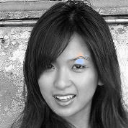} &
	\includegraphics[align=c,height=\ploth]{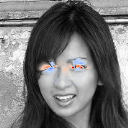} &
	\includegraphics[align=c,height=\ploth]{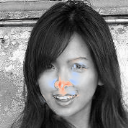} &
	\includegraphics[align=c,height=\ploth]{img/rafd_stretch/lr/colorbar} 
	\\[15pt]\midrule
	\\[-13pt]\midrule
	& & \small{Input} & \small{\makecell{No \\ smile }} & \small{\makecell{Big \\ nose}} & \small{\makecell{Arched \\ eyebrows}} & \small{\makecell{ Narrowed \\ eyes}}  & \small{\makecell{ Pointy \\ nose}} &
	\small{Input} & \small{\makecell{Smile}} & \small{\makecell{Big \\ nose}} & \small{\makecell{Arched \\ eyebrows}} & \small{\makecell{ Narrowed \\ eyes}}  & \small{\makecell{ Pointy \\ nose}} & \\[9pt]
		\multirow{3}{*}{\parbox[t]{4mm}{\rotatebox[origin=c]{90}{\hspace{-2.2cm} WarpGAN}}} &
	\parbox[t]{3mm}{\rotatebox[origin=c]{90}{\scriptsize{Input}}} &
	\includegraphics[align=c,height=\ploth]{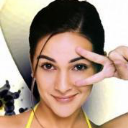} &
	\includegraphics[align=c,height=\ploth]{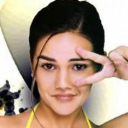} &
	\includegraphics[align=c,height=\ploth]{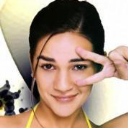} &
	\includegraphics[align=c,height=\ploth]{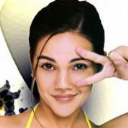} &
	\includegraphics[align=c,height=\ploth]{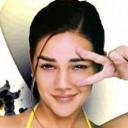} &
	\includegraphics[align=c,height=\ploth]{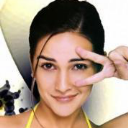} &
	\includegraphics[align=c,height=\ploth]{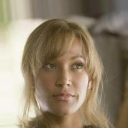} &
	\includegraphics[align=c,height=\ploth]{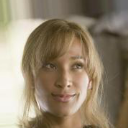} &
	\includegraphics[align=c,height=\ploth]{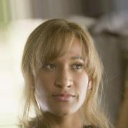} &
	\includegraphics[align=c,height=\ploth]{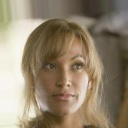} &
	\includegraphics[align=c,height=\ploth]{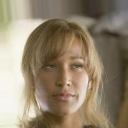} &
	\includegraphics[align=c,height=\ploth]{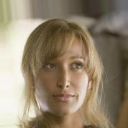} & 
	\\[16pt]
	& \parbox[t]{3mm}{\rotatebox[origin=c]{90}{\scriptsize{Stretch map}}} &
	\includegraphics[align=c,height=\ploth]{sup_imgs/celeba_stretch/binary/\imagec/input} &
	\includegraphics[align=c,height=\ploth]{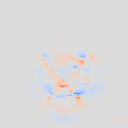} &
	\includegraphics[align=c,height=\ploth]{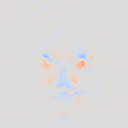} &
	\includegraphics[align=c,height=\ploth]{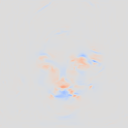} &
	\includegraphics[align=c,height=\ploth]{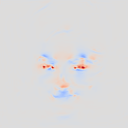} &
	\includegraphics[align=c,height=\ploth]{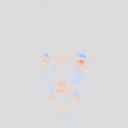} &
	\includegraphics[align=c,height=\ploth]{sup_imgs/celeba_stretch/binary/\imaged/input} &
	\includegraphics[align=c,height=\ploth]{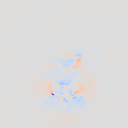} &
	\includegraphics[align=c,height=\ploth]{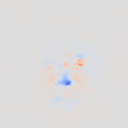} &
	\includegraphics[align=c,height=\ploth]{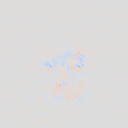} &
	\includegraphics[align=c,height=\ploth]{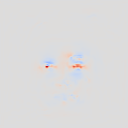} &
	\includegraphics[align=c,height=\ploth]{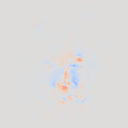} &
	\includegraphics[align=c,height=\ploth]{img/rafd_stretch/lr/colorbar} 
	\\[16pt]
	& \parbox[t]{3mm}{\rotatebox[origin=c]{90}{\scriptsize{Overlay}}} &
	\includegraphics[align=c,height=\ploth]{sup_imgs/celeba_stretch/binary/\imagec/input} &
	\includegraphics[align=c,height=\ploth]{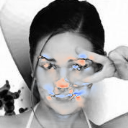} &
	\includegraphics[align=c,height=\ploth]{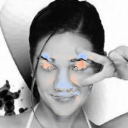} &
	\includegraphics[align=c,height=\ploth]{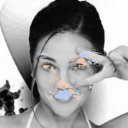} &
	\includegraphics[align=c,height=\ploth]{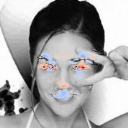} &
	\includegraphics[align=c,height=\ploth]{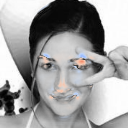} &
	\includegraphics[align=c,height=\ploth]{sup_imgs/celeba_stretch/binary/\imaged/input} &
	\includegraphics[align=c,height=\ploth]{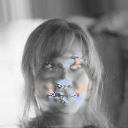} &
	\includegraphics[align=c,height=\ploth]{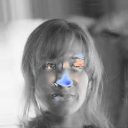} &
	\includegraphics[align=c,height=\ploth]{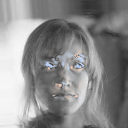} &
	\includegraphics[align=c,height=\ploth]{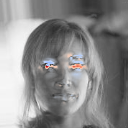} &
	\includegraphics[align=c,height=\ploth]{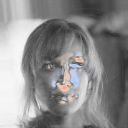} &
	\includegraphics[align=c,height=\ploth]{img/rafd_stretch/lr/colorbar} 
	\\[14pt]\midrule
	\multirow{3}{*}{\parbox[t]{4mm}{\rotatebox[origin=c]{90}{\hspace{-2.2cm} WarpGAN+}}} &
	\parbox[t]{3mm}{\rotatebox[origin=c]{90}{\scriptsize{Input}}} &
	\includegraphics[align=c,height=\ploth]{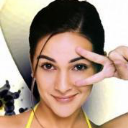} &
	\includegraphics[align=c,height=\ploth]{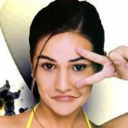} &
	\includegraphics[align=c,height=\ploth]{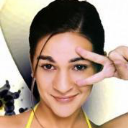} &
	\includegraphics[align=c,height=\ploth]{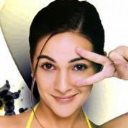} &
	\includegraphics[align=c,height=\ploth]{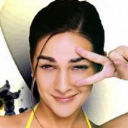} &
	\includegraphics[align=c,height=\ploth]{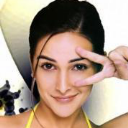} &
	\includegraphics[align=c,height=\ploth]{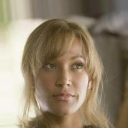} &
	\includegraphics[align=c,height=\ploth]{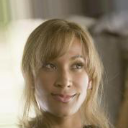} &
	\includegraphics[align=c,height=\ploth]{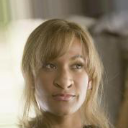} &
	\includegraphics[align=c,height=\ploth]{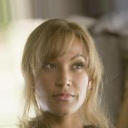} &
	\includegraphics[align=c,height=\ploth]{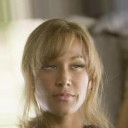} &
	\includegraphics[align=c,height=\ploth]{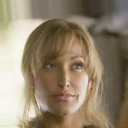} & 
	\\[16pt]
	& \parbox[t]{3mm}{\rotatebox[origin=c]{90}{\scriptsize{Stretch map}}} &
	\includegraphics[align=c,height=\ploth]{sup_imgs/celeba_stretch/ternary/\imagec/input} &
	\includegraphics[align=c,height=\ploth]{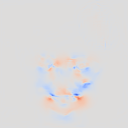} &
	\includegraphics[align=c,height=\ploth]{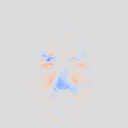} &
	\includegraphics[align=c,height=\ploth]{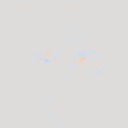} &
	\includegraphics[align=c,height=\ploth]{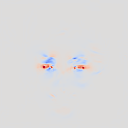} &
	\includegraphics[align=c,height=\ploth]{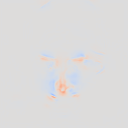} &
	\includegraphics[align=c,height=\ploth]{sup_imgs/celeba_stretch/ternary/\imaged/input} &
	\includegraphics[align=c,height=\ploth]{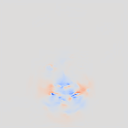} &
	\includegraphics[align=c,height=\ploth]{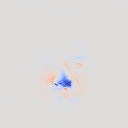} &
	\includegraphics[align=c,height=\ploth]{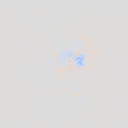} &
	\includegraphics[align=c,height=\ploth]{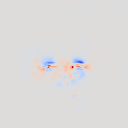} &
	\includegraphics[align=c,height=\ploth]{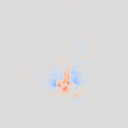} &
	\includegraphics[align=c,height=\ploth]{img/rafd_stretch/lr/colorbar} 
	\\[16pt]
	& \parbox[t]{3mm}{\rotatebox[origin=c]{90}{\scriptsize{Overlay}}} &
	\includegraphics[align=c,height=\ploth]{sup_imgs/celeba_stretch/ternary/\imagec/input} &
	\includegraphics[align=c,height=\ploth]{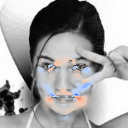} &
	\includegraphics[align=c,height=\ploth]{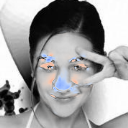} &
	\includegraphics[align=c,height=\ploth]{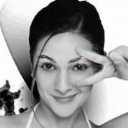} &
	\includegraphics[align=c,height=\ploth]{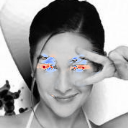} &
	\includegraphics[align=c,height=\ploth]{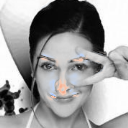} &
	\includegraphics[align=c,height=\ploth]{sup_imgs/celeba_stretch/ternary/\imaged/input} &
	\includegraphics[align=c,height=\ploth]{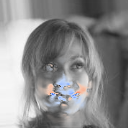} &
	\includegraphics[align=c,height=\ploth]{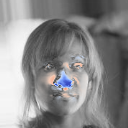} &
	\includegraphics[align=c,height=\ploth]{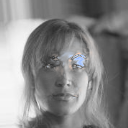} &
	\includegraphics[align=c,height=\ploth]{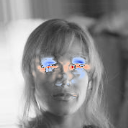} &
	\includegraphics[align=c,height=\ploth]{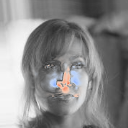} &
	\includegraphics[align=c,height=\ploth]{img/rafd_stretch/lr/colorbar} 
	\end{tabular}
	}
	\caption{Stretch maps computed from the warp fields, for WarpGAN and WarpGAN+.
	The log determinant of the Jacobian of the  warp is shown, where blue indicates stretching and red corresponds to squashing.
	Our binary label transformation scheme (WarpGAN+) leads to correctly localized edits.
	}
	\label{fig:supp_celeba_stretch_example}
\end{figure}

%
%

%
%

\FloatBarrier
\newpage
\section{Ablation study}

\subsection{Effect of each loss}
In this section we evaluate the performance of the model after removing each of the losses, where ``(w/o) Cycle'' corresponds to removing $L_{\text{c}}$, ``(w/o) Smooth'' corresponds to removing $L_{\text{s}}$, ``(w/o) Cls'' corresponds to removing $L_{\text{cls}}$, ``(pixel) Cycle'' corresponds to using eq.~4 instead of eq.~8 in the paper, and ``(w/o) Adv'' corresponds to removing $L_{\text{adv}}$ and $L_{\text{gp}}$.


\def\plotw{0.35\linewidth}

\begin{figure}[H]
	\centering
	\includegraphics[width=\plotw]{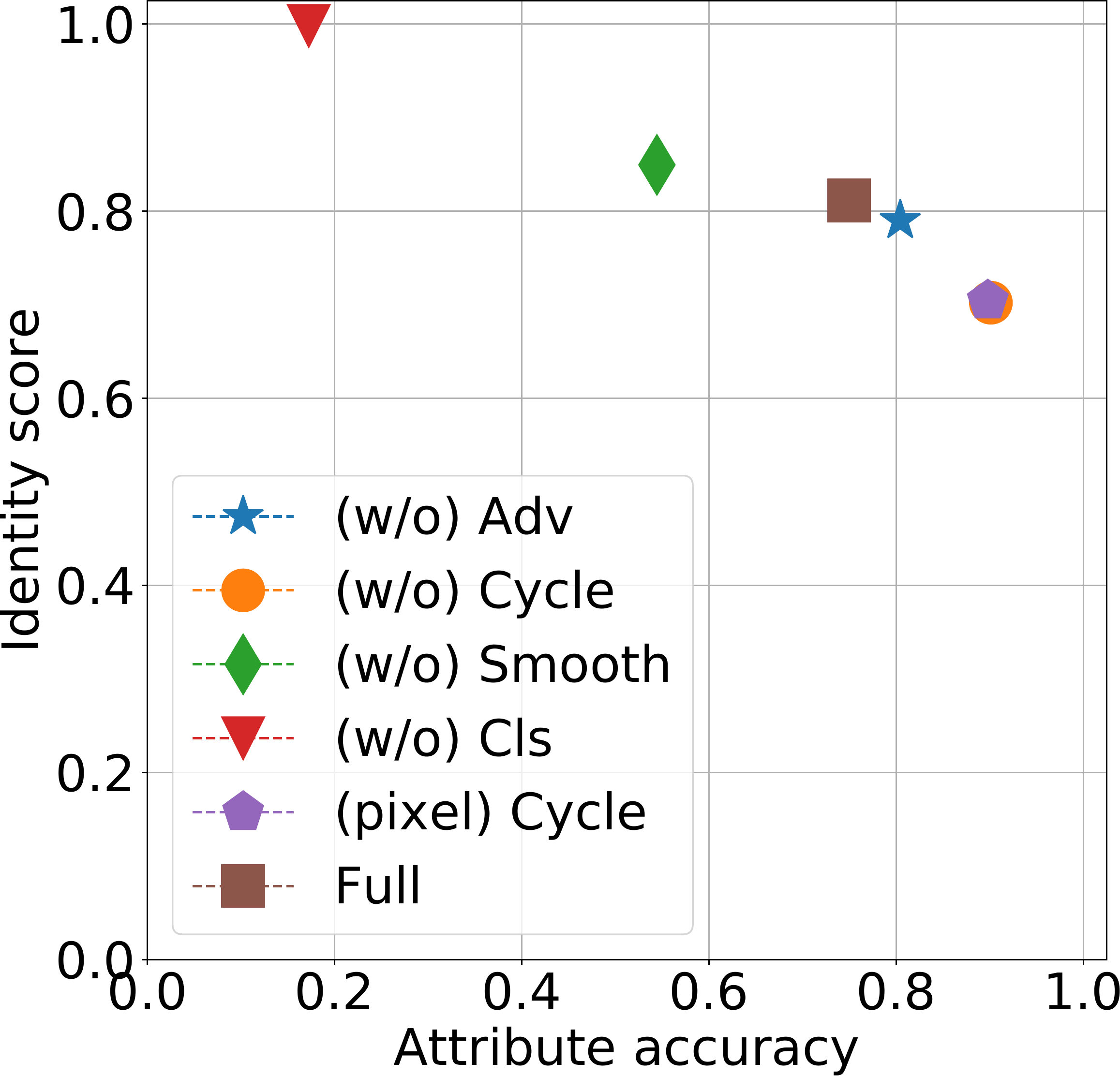}
	\caption{Presence of the edited attribute ($x$-axis) vs face re-identification score ($y$-axis), higher is better.
	Removing each loss in our model has a detrimental effect in either accuracy or identity preservation.
	The adversarial loss seems to have little effect, however, we qualitatively observed that without it, the edited images were less realistic.}
	\label{fig:suppl_attr_vs_identity_abblation}
\end{figure}

\def\plotw{0.1\linewidth}
\def\imagea{1} 
\def\attra{0}
\def\attrb{1}
\def\attrc{2}
\def\attrd{3}
\def\attre{4}

\begin{figure}[H]
	\centering
	\setlength{\tabcolsep}{1pt} 
	\small{	
	\begin{tabular}{cccccccc}
	& Input & (w/o) Cycle & (w/o) Smooth & (w/o) Adv & (w/o) Cls & (pixel) Cycle & Full \\[5pt]
	\makecell{ No \\ smile } &
	\includegraphics[align=c,width=\plotw]{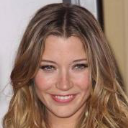} &
	\includegraphics[align=c,width=\plotw]{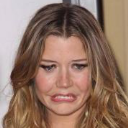} &
	\includegraphics[align=c,width=\plotw]{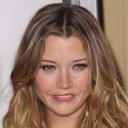} &
	\includegraphics[align=c,width=\plotw]{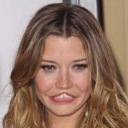} &
	\includegraphics[align=c,width=\plotw]{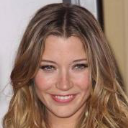} &
	\includegraphics[align=c,width=\plotw]{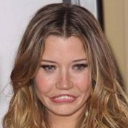} &
	\includegraphics[align=c,width=\plotw]{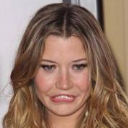} \\[21pt]
	\makecell{ Big \\ nose } &
	\includegraphics[align=c,width=\plotw]{img/ablation/input/\imagea} &
	\includegraphics[align=c,width=\plotw]{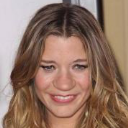} &
	\includegraphics[align=c,width=\plotw]{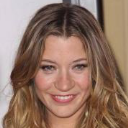} &
	\includegraphics[align=c,width=\plotw]{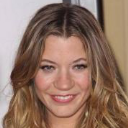} &
	\includegraphics[align=c,width=\plotw]{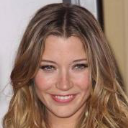} &
	\includegraphics[align=c,width=\plotw]{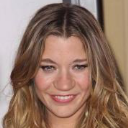} &
	\includegraphics[align=c,width=\plotw]{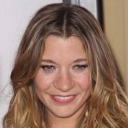} \\[21pt]
	\makecell{ Arched \\ eyebrows } &
	\includegraphics[align=c,width=\plotw]{img/ablation/input/\imagea} &
	\includegraphics[align=c,width=\plotw]{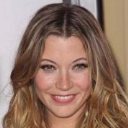} &
	\includegraphics[align=c,width=\plotw]{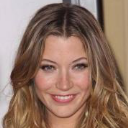} &
	\includegraphics[align=c,width=\plotw]{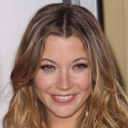} &
	\includegraphics[align=c,width=\plotw]{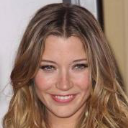} &
	\includegraphics[align=c,width=\plotw]{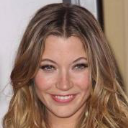} &
	\includegraphics[align=c,width=\plotw]{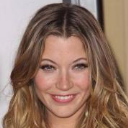} \\[21pt]
	\makecell{ Narrowed \\ eyes } &
	\includegraphics[align=c,width=\plotw]{img/ablation/input/\imagea} &
	\includegraphics[align=c,width=\plotw]{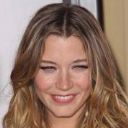} &
	\includegraphics[align=c,width=\plotw]{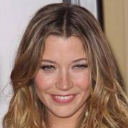} &
	\includegraphics[align=c,width=\plotw]{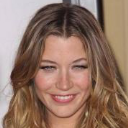} &
	\includegraphics[align=c,width=\plotw]{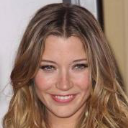} &
	\includegraphics[align=c,width=\plotw]{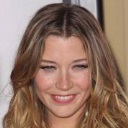} &
	\includegraphics[align=c,width=\plotw]{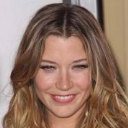} \\[21pt]
	\makecell{ Pointy \\ nose } &
	\includegraphics[align=c,width=\plotw]{img/ablation/input/\imagea} &
	\includegraphics[align=c,width=\plotw]{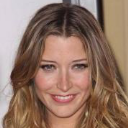} &
	\includegraphics[align=c,width=\plotw]{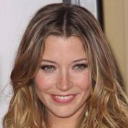} &
	\includegraphics[align=c,width=\plotw]{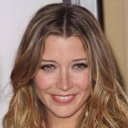} &
	\includegraphics[align=c,width=\plotw]{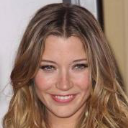} &
	\includegraphics[align=c,width=\plotw]{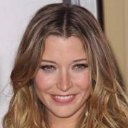} &
	\includegraphics[align=c,width=\plotw]{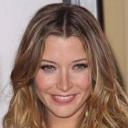} 
	\end{tabular}
	}
	\caption{Ablation study, where we remove different losses in our model.
	For each loss, (w/o) Cycle: significant artifacts are introduced, (w/o) Smooth: leads to poor generalization, (w/o) Adv: unrealistic warps, (pixel) Cycle: exaggerated warps,  and (w/o) Cls: trivial solution on the identity transform.}
	\label{fig:supp_ablation}
\end{figure}

\subsection{Effect of $\alpha$}
In this section we quantitatively evaluate the effect of scaling the displacement fields by a scalar $\alpha$.
For this experiment, we take WarpGAN+ trained with $\lambda_{\text{cls}} = 0.25$ and we evaluate the identity score and the attribute accuracy on the test set for different values of $\alpha$.
Results are shown in Fig.~\ref{fig:suppl_attr_vs_identity_alpha_abblation} for this model, which is denoted as WarpGAN+$\alpha$.
The curve produced by employing different values of $\alpha$ is very similar to the curve in Figure~8 in the paper, which was produced by modifying $\lambda_{\text{cls}}$.
This implies that the model is relatively robust to the choice of $\lambda_{\text{cls}}$, as a similar effect to changing the value of $\lambda_{\text{cls}}$ used during training can be achieved by choosing an alternative value of $\alpha$ at test time.
This is in contrast to previous work, where modifying the strength of the effects requires training a model with the new parameters.


\def\plotw{0.38\linewidth}

\begin{figure}[H]
	\centering
	\includegraphics[width=\plotw]{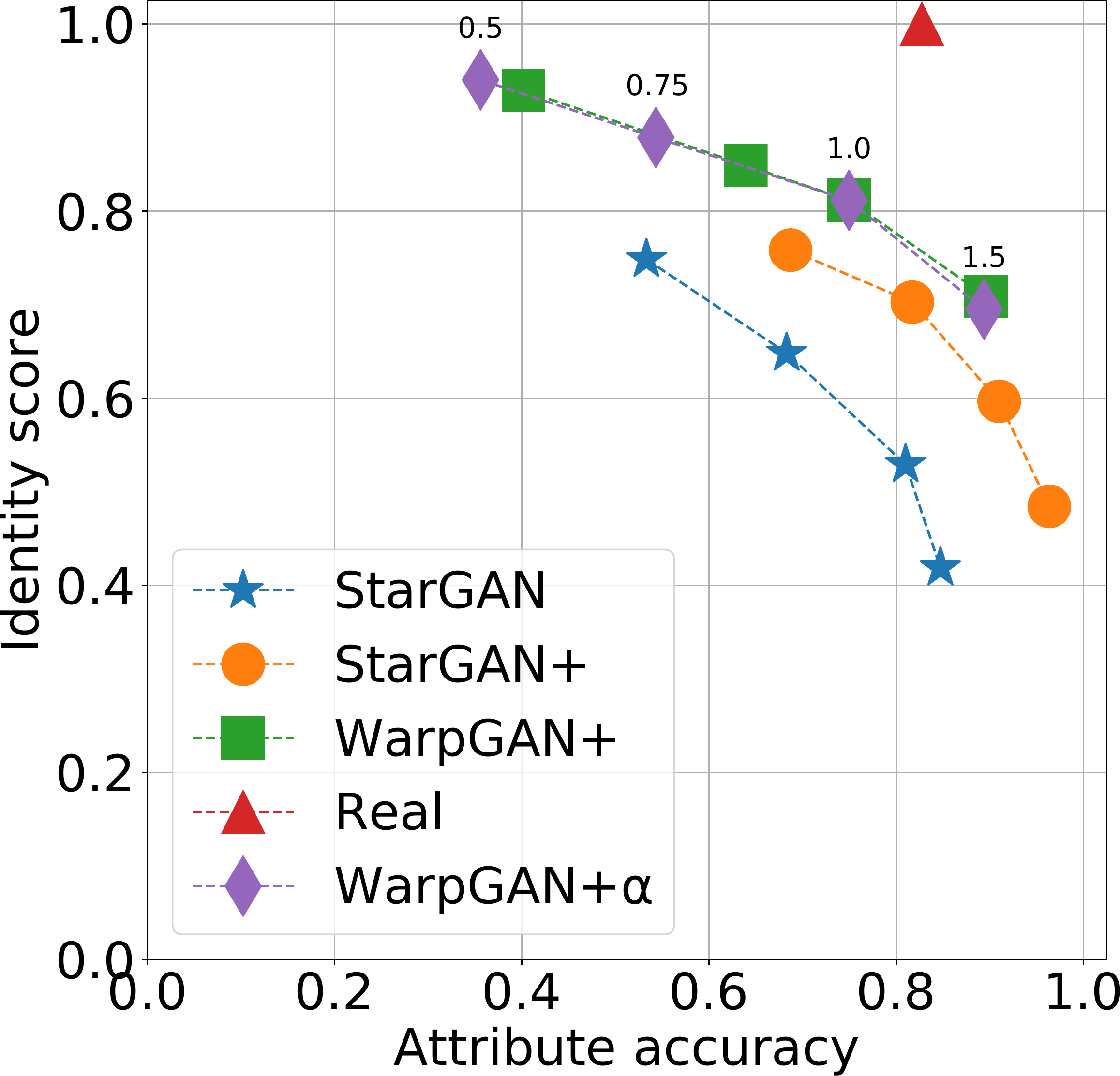}
	\caption{Presence of the edited attribute ($x$-axis) vs face re-identification score ($y$-axis), higher is better.
	For all models except WarpGAN+$\alpha$, this figure is identical to Fig.~8 in the paper.
	For WarpGAN+$\alpha$  the value of $\alpha$ is shown on top of each marker.
	Modifying the $\alpha$ value at test time in our model has a similar effect as training the model with different $\lambda_{\text{cls}}$ values.}
	\label{fig:suppl_attr_vs_identity_alpha_abblation}
\end{figure}

\section{Face alignment}

\begin{figure}[H]
\centering
\includegraphics[width=0.3\linewidth]{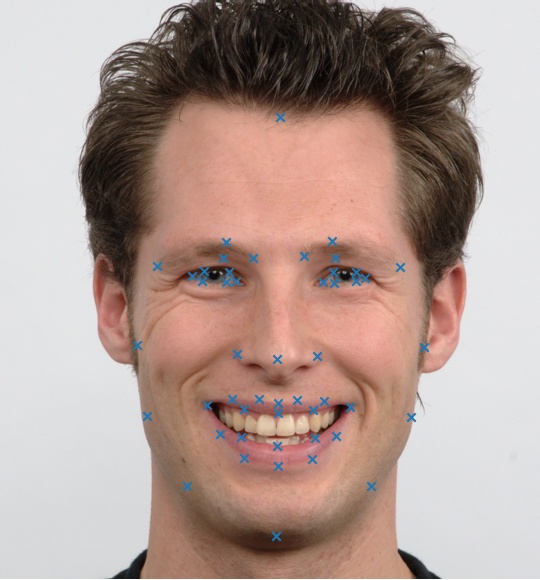}
\caption{An example of the locations of the 49 face landmarks used for the internal face landmark detection network.
}
\label{fig:landmark_locations}
\end{figure}

For the CelebA dataset we use the aligned version provided by the authors, which uses two landmark locations located at the eyes of each subject.
Each image is first center-cropped to 178$\times$178, and then resized to 128$\times$128.
For the in-the-wild high resolution images from Flickr, an internal face landmark detection network is used to automatically align and resize images to the mean CelebA face at 128$\times$128.
The location of the face landmarks used by the network are shown in Fig.~\ref{fig:landmark_locations}.
For the Cub200 dataset the face alignment to 128 $\times$ 128 uses four landmark locations: the beak, the crown, the forehead and the right eye.
If the right eye is not visible, the image is left-right flipped.

\FloatBarrier
\newpage
\section{Network architectures and training details}
\label{sec:network_architecture}

The networks were trained on CelebA for 20 epochs and on Cub200 for 1545 epochs (due to the reduced size of this dataset).
The Adam optimizer~\cite{Kingma2015Adam} is used with a learning rate of $0.0001$, with $\beta_1 = 0.5$ and $\beta_2 = 0.999$.

Our network architectures are based on the StarGAN model.
In the generator all transpose convolutions are replaced with bilinear resizing followed by convolution, and instance normalization is replaced by batch normalization.
For the discriminator the StarGAN architecture is used without any modifications.
In both tables the following notation is used, N is the number of output channels, K is the kernel size, S is the stride size, P is the padding size, and BN is batch normalization.
The warping function, $T$, is implemented with a TensorFlow function during training, and with an OpenCV one for inference:
\\ \footnotesize{\texttt{$T(\bx, \bw) = $ tf.contrib.image.dense\_image\_warp($\bx$, $\bw$)}},
\\ \footnotesize{\texttt{$T(\bx, \bw) = $ cv2.remap($\bx$, $\bw$, interpolation=cv2.INTER\_CUBIC)}}.

\begin{table}[H]
\centering
\setlength{\tabcolsep}{15pt}
\small{
\begin{tabular}{ccc}
Part & Input $\rightarrow$ Output Shape & Layer information \\[5pt]\midrule
\multirow{3}{*}{Down-sampling} & $(h,w, 3 + r) \rightarrow (h,w, 64)$ & CONV-(N64, K7x7, S1, P3), ReLU, BN \\[5pt]
& $(h,w, 64) \rightarrow (\frac{h}{2} , \frac{w}{2}, 128)$ & CONV-(N128, K4x4, S2, P1), ReLU, BN \\[5pt]
& $(\frac{h}{2} , \frac{w}{2}, 128) \rightarrow (\frac{h}{4} , \frac{w}{4}, 256)$ & CONV-(N256, K4x4, S2, P1), ReLU, BN \\[5pt] \midrule
\multirow{6}{*}{Bottleneck} & $(\frac{h}{4} , \frac{w}{4}, 256) \rightarrow (\frac{h}{4} , \frac{w}{4}, 256)$ & Residual Block: CONV-(N256, K3x3, S1, P1), ReLU, BN \\[5pt]
& $(\frac{h}{4} , \frac{w}{4}, 256) \rightarrow (\frac{h}{4} , \frac{w}{4}, 256)$ & Residual Block: CONV-(N256, K3x3, S1, P1), ReLU, BN \\[5pt]
& $(\frac{h}{4} , \frac{w}{4}, 256) \rightarrow (\frac{h}{4} , \frac{w}{4}, 256)$ & Residual Block: CONV-(N256, K3x3, S1, P1), ReLU, BN \\[5pt]
& $(\frac{h}{4} , \frac{w}{4}, 256) \rightarrow (\frac{h}{4} , \frac{w}{4}, 256)$ & Residual Block: CONV-(N256, K3x3, S1, P1), ReLU, BN \\[5pt]
& $(\frac{h}{4} , \frac{w}{4}, 256) \rightarrow (\frac{h}{4} , \frac{w}{4}, 256)$ & Residual Block: CONV-(N256, K3x3, S1, P1), ReLU, BN \\[5pt]
& $(\frac{h}{4} , \frac{w}{4}, 256) \rightarrow (\frac{h}{4} , \frac{w}{4}, 256)$ & Residual Block: CONV-(N256, K3x3, S1, P1), ReLU, BN \\[5pt] \midrule
\multirow{5}{*}{Up-sampling} & $(\frac{h}{4} , \frac{w}{4}, 256) \rightarrow (\frac{h}{2} , \frac{w}{2}, 256)$ & Bilinear resize \\[5pt]
& $(\frac{h}{2} , \frac{w}{2}, 256) \rightarrow (\frac{h}{2} , \frac{w}{2}, 128)$ & CONV-(N128, K4x4, S1, P1), ReLU, BN \\[5pt]
& $(\frac{h}{2} , \frac{w}{2}, 128) \rightarrow (h , w, 128)$ & Bilinear resize \\[5pt]
& $(h , w, 64) \rightarrow (h , w, 64)$ & CONV-(N64, K4x4, S1, P1), ReLU, BN \\[5pt]
& $(h , w, 64) \rightarrow (h , w, 2)$ & CONV-(N2, K7x7, S1, P1) \\[5pt] \bottomrule
\end{tabular}
}
\label{tb:generator_arch}
\caption{Architecture for the warping network, $W$, the last layer is the displacement field $\bw$, $h$ and $w$ denote the dimensionality of the input image, and $r$ the number of attributes.}
\end{table}

\begin{table}[H]
\centering
\setlength{\tabcolsep}{15pt}
\small{
\begin{tabular}{ccc}
Part & Input $\rightarrow$ Output Shape & Layer information \\[5pt]\midrule
\multirow{6}{*}{Down-sampling} & $(h,w, 3) \rightarrow (\frac{h}{2} , \frac{w}{2}, 64)$ & CONV-(N64, K4x4, S2, P1), Leaky ReLU \\[5pt]
& $(\frac{h}{2} , \frac{w}{2}, 64) \rightarrow (\frac{h}{4} , \frac{w}{4}, 128)$ & CONV-(N128, K4x4, S2, P1), Leaky ReLU \\[5pt]
& $(\frac{h}{4} , \frac{w}{4}, 128) \rightarrow (\frac{h}{8} , \frac{w}{8}, 256)$ & CONV-(N256, K4x4, S2, P1), Leaky ReLU \\[5pt]
& $(\frac{h}{8} , \frac{w}{8}, 256) \rightarrow (\frac{h}{16} , \frac{w}{16}, 512)$ & CONV-(N512, K4x4, S2, P1), Leaky ReLU \\[5pt]
& $(\frac{h}{16} , \frac{w}{16}, 512) \rightarrow (\frac{h}{32} , \frac{w}{32}, 1024)$ & CONV-(N1024, K4x4, S2, P1), Leaky ReLU \\[5pt]
& $(\frac{h}{32} , \frac{w}{32}, 1024) \rightarrow (\frac{h}{64} , \frac{w}{64}, 2048)$ & CONV-(N2048, K4x4, S2, P1), Leaky ReLU \\[5pt]\midrule
Output layer $D$ & $(\frac{h}{64} , \frac{w}{64}, 2048) \rightarrow (\frac{h}{64} , \frac{w}{64}, 1)$ & CONV-(N1, K3x3, S1, P1)\\[5pt]
Output layer $C$ & $(\frac{h}{64} , \frac{w}{64}, 2048) \rightarrow (1, 1, r)$ & CONV-(N($r$), K$\frac{h}{64}$x$\frac{w}{64}$, S1, P0)\\[5pt] \bottomrule
\end{tabular}
}
\label{tb:discriminator_arch}
\caption{Architecture for the discriminator and the classifier networks, $D$ and $C$.
The kernel weights in the down-sampling layers are shared by $D$ and $C$.}
\end{table}

\FloatBarrier
\newpage

\section{Quantitative results: details}

\subsection{Accuracy vs identity preservation}

\normalsize
In this section we give additional detail about the face re-identification network.
We also provide attribute accuracy values and identity scores per attribute for the models used in the paper, namely, for StarGAN and StarGAN+ trained with $\lambda_{\text{cls}} = 0.25$ and for WarpGAN+ with $\lambda_{\text{cls}} = 2.00$.

\subsubsection{Re-identification network}
For the face re-identificaiton scores, presented in Fig.~8 in the paper, we use a Facenet model pretrained on the MS-Celeb-1M dataset~\cite{Guo2016Celeb1M}.
This dataset consists of 10 million images and 100k unique identities.
As both CelebA and MS-Celeb-1M were collected from publicly available Internet images, we expect some overlap between both datasets.
This pretrained model is provided by the authors and is publicly available at \url{https://github.com/davidsandberg/facenet}.

\begin{table}[H]
\centering
\setlength{\tabcolsep}{2pt}
\small{
\begin{tabularx}{\linewidth}{YYYYYY|Y}
	\makecell{\textbf{Model} \\ ~} & \makecell{\textbf{Smiling} \\ ~}  & \makecell{\textbf{Big} \\ \textbf{nose}} & \makecell{\textbf{Arched} \\ \textbf{eyebrows}} & \makecell{\textbf{Narrowed} \\ \textbf{eyes}} & \makecell{\textbf{Pointy } \\ \textbf{nose}} & \makecell{\textbf{Mean} \\ ~} \\ \midrule
	StarGAN  & $0.65$ & $0.60$ & $0.64$ & $0.66$ & $0.68$ & $0.65$ \\ \midrule
	StarGAN+  & $0.72$ & $0.66$ & $0.67$ & $0.78$ & $0.69$ & $0.70$ \\ \midrule
	\textbf{WarpGAN+} & $\mathbf{0.83}$ & $\mathbf{0.73}$ & $\mathbf{0.81}$ & $\mathbf{0.87}$ & $\mathbf{0.82}$ & $\mathbf{0.81}$ \\ \midrule \midrule
	Real  & $1.00$ & $1.00$ & $1.00$ & $1.00$ & $1.00$ & $1.00$ \\ \bottomrule
\end{tabularx} 
}
\caption{Quantitative comparison of the re-identification score on real and generated images on the CelebA dataset evaluated with the face re-identification network, higher is better.
}
\label{tb:supp_id_score}
\end{table}





\subsubsection{Attribute classification accuracy}

\begin{table}[H]
\centering
\setlength{\tabcolsep}{2pt}
\small{
\begin{tabularx}{\linewidth}{YYYYYY|Y}
	\makecell{\textbf{Model} \\ ~} & \makecell{\textbf{Smiling} \\ ~}  & \makecell{\textbf{Big} \\ \textbf{nose}} & \makecell{\textbf{Arched} \\ \textbf{eyebrows}} & \makecell{\textbf{Narrowed} \\ \textbf{eyes}} & \makecell{\textbf{Pointy } \\ \textbf{nose}} & \makecell{\textbf{Mean} \\ ~} \\ \midrule
	StarGAN  & $0.84$ & $0.60$ & $0.69$ & $0.65$ 	 & $0.62$ & $0.68$ \\ \midrule
	StarGAN+  & $\mathbf{0.92}$ & $\mathbf{0.73}$ & $\mathbf{0.87}$ & $\mathbf{0.75}$ 	 & $\mathbf{0.82}$ & $\mathbf{0.82}$ \\ \midrule
	\textbf{WarpGAN+}  & $0.72$ 			& $0.72$ 		   & $0.83$ 		  & $0.74$ 		  	 & $0.74$ 			& $0.75$ 		   \\ \midrule \midrule
	Real      						 & $0.92$ 			& $0.81$ 		   & $0.82$ 		  & $0.88$			 	 & $0.72$ 			& $0.83$  		   \\ \bottomrule
\end{tabularx} 
}
\caption{Quantitative comparison of the attribute classification accuracy on real and generated images on the CelebA dataset evaluated with a separate classification network, higher is better.
}
\label{tb:supp_classification_accuracy}
\end{table}

%

%

%

%

\subsection{User study}

In the user study, for both experiments, to evaluate the reliability of the workers, a number of easy to classify images were mixed with the data, and used as a control.
Workers needed to give the right answer to at least $90$\% of the control images for their data to be considered reliable.
Images with fewer than 3 annotations are discarded, as they are considered unreliable data.
Finally, a simple majority voting scheme was used to determine the classification of each image.

For the experiment evaluating realism, typical failure cases for all models were shown to the workers before commencing the task, as examples of fake images.
For the evaluation of the presence of the target attribute, to guide the workers, curated examples from training data edited with our model were shown to highlight the differences between the attributes.

Some images in the CelebA dataset contain border artifacts due to the alignment process that the authors used for the aligned version of the dataset.
In order to get more reliable results, none of these images were included in the pool of 250 images used for the study.

\subsubsection{Attribute classification accuracy}

\begin{table}[H]
\centering
\setlength{\tabcolsep}{2pt}
\small{
\begin{tabularx}{\linewidth}{YYYYYY|Y}
	\makecell{\textbf{Model} \\ ~} & \makecell{\textbf{Smiling} \\ ~}  & \makecell{\textbf{Big} \\ \textbf{nose}} & \makecell{\textbf{Arched} \\ \textbf{eyebrows}} & \makecell{\textbf{Narrowed} \\ \textbf{eyes}} & \makecell{\textbf{Pointy } \\ \textbf{nose}} & \makecell{\textbf{Mean} \\ ~} \\ \midrule
	StarGAN  & $\mathbf{0.85}$ & $0.84$ & $0.75$ & $0.83$ & $0.76$ & $0.81$ \\ \midrule
	StarGAN+  & $\mathbf{0.85}$ & $0.84$ & $\mathbf{0.89}$ & $0.86$ & $0.83$ & $\mathbf{0.86}$ \\ \midrule
	\textbf{WarpGAN+}  & $0.63$ & $\mathbf{0.92}$ & $0.83$ & $\mathbf{0.89}$ & $\mathbf{0.88}$ & $0.84$ \\ \midrule \midrule
	Real  & $0.88$ & $0.64$ & $0.74$ & $0.56$ & $0.36$ & $0.63$ \\ \bottomrule
\end{tabularx} 
}
\caption{Quantitative comparison of the attribute classification accuracy on real and generated images on the CelebA dataset evaluated with a user study, higher is better.
}
\label{tb:supp_users_classification_accuracy}
\end{table}





\subsubsection{Realism accuracy}
\begin{table}[H]
\centering
\setlength{\tabcolsep}{2pt}
\small{
\begin{tabularx}{\linewidth}{YYYYYY|Y}
	\makecell{\textbf{Model} \\ ~} & \makecell{\textbf{Smiling} \\ ~}  & \makecell{\textbf{Big} \\ \textbf{nose}} & \makecell{\textbf{Arched} \\ \textbf{eyebrows}} & \makecell{\textbf{Narrowed} \\ \textbf{eyes}} & \makecell{\textbf{Pointy } \\ \textbf{nose}} & \makecell{\textbf{Mean} \\ ~} \\ \midrule
	StarGAN  & $0.40$ & $0.52$ & $0.62$ & $0.41$ & $0.74$ & $0.52$ \\ \midrule
	StarGAN+  & $0.37$ & $0.40$ & $0.59$ & $0.38$ & $0.60$ & $0.46$ \\ \midrule
	\textbf{WarpGAN+}  & $\mathbf{0.42}$ & $\mathbf{0.64}$ & $\mathbf{0.79}$ & $\mathbf{0.57}$ & $\mathbf{0.82}$ & $\mathbf{0.65}$ \\ \midrule \midrule
	Real  & $0.97$ & $0.89$ & $0.98$ & $0.96$ & $0.94$ & $0.95$ \\ \bottomrule
\end{tabularx} 
}
\caption{Quantitative comparison of image realism both on real and generated images on the CelebA dataset evaluated with a user study, higher is better.}
\label{tb:supp_realism_accuracy}
\end{table}

\end{document}